\documentclass[10pt,twocolumn,letterpaper]{article}
\PassOptionsToPackage{table,dvipsnames}{xcolor}
\usepackage[pagenumbers]{cvpr}
\input{preamble}
\usepackage{booktabs,multirow,makecell}
\usepackage{graphicx}
\usepackage{xspace}
\newcommand{\bench}{ASTERISM\xspace}
\newcommand{\benchlite}{\bench-Lite\xspace}
\newcommand{\cica}{\textsc{CICA}\xspace}
\definecolor{oursgreen}{HTML}{DCEDC8}
\newcommand{\oursrow}{\rowcolor{oursgreen}}
\newcommand{\cicaseven}{\textsc{CICA}-Qwen2.5-VL-7B\xspace}
\newcommand{\cicanine}{\textsc{CICA}-Qwen3.5-9B\xspace}
\newcommand{\cicaglm}{\textsc{CICA}-GLM-4.1V-9B-Thinking\xspace}
\newcommand{\sub}{\hspace{0.6em}$\llcorner$\,+\,}

\definecolor{cvprblue}{rgb}{0.21,0.49,0.74}
\usepackage[pagebackref,breaklinks,colorlinks,allcolors=cvprblue]{hyperref}
\hypersetup{pdftitle={ASTERISM: Do Multimodal LLMs Actually Look at All the Images?},pdfauthor={Chanseul Cho, Dongwoo Kim, Noori Bae, Seungjae Moon, Seongsu Lee, Youngmin Ro}}
\usepackage{etoc}
\AtEndPreamble{\crefname{figure}{Fig.}{Figs.}\Crefname{figure}{Fig.}{Figs.}\crefname{table}{Tab.}{Tabs.}\Crefname{table}{Tab.}{Tabs.}\crefname{section}{Sec.}{Secs.}\Crefname{section}{Sec.}{Secs.}\crefname{subsection}{Sec.}{Secs.}\Crefname{subsection}{Sec.}{Secs.}\crefname{equation}{Eq.}{Eqs.}\Crefname{equation}{Eq.}{Eqs.}\crefname{appendix}{Supp.}{Supp.}\Crefname{appendix}{Supp.}{Supp.}\crefname{subappendix}{Supp.}{Supp.}\Crefname{subappendix}{Supp.}{Supp.}\crefname{subsubappendix}{Supp.}{Supp.}\Crefname{subsubappendix}{Supp.}{Supp.}}
\title{\bench: Do Multimodal LLMs Actually Look at All the Images?}
\author{Chanseul Cho \quad Dongwoo Kim \quad Noori Bae \quad Seungjae Moon \quad Seongsu Lee \quad Youngmin Ro$^{\dagger}$\\
Machine Intelligence Lab, University of Seoul\\
{\tt\small \{chanseul2001, shanghai110930\}@gmail.com}\\
{\tt\small \{2bee5, msj0243, ssulee0206, youngmin.ro\}@uos.ac.kr}}
\begin{document}
\etocdepthtag.toc{mtmain}
\twocolumn[{%
\renewcommand\twocolumn[1][]{#1}%
\maketitle
\begin{center}
  \centering
  \includegraphics[width=\linewidth]{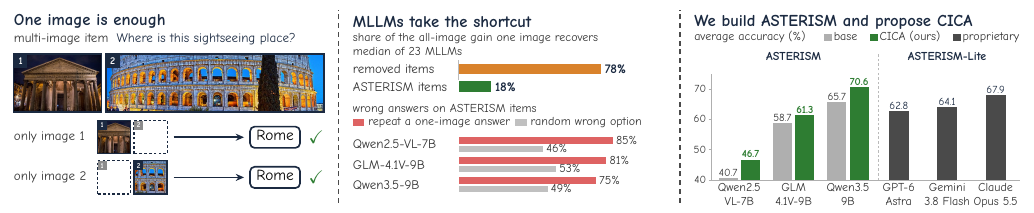}
  \captionof{figure}{\textbf{Many multi-image items reward a single-image shortcut that MLLMs take, and \cica moves credit to the tokens that need more than one image.} Either photo alone answers this item of a multi-image benchmark (left). With the other images grayed out, one image recovers most of the all-image gain, the accuracy that all images add over the text alone, on items the audit removes but little on \bench items, and most wrong answers on \bench items repeat a one-image answer (middle). \cica raises the \bench score of each base model, and on the hard subset \benchlite the strongest proprietary MLLM scores 68\% (right).}
  \label{fig:teaser}
\end{center}%
\vspace{2mm}
}]
\begingroup\renewcommand\thefootnote{\fnsymbol{footnote}}\footnotetext[2]{Corresponding author.}\endgroup
\begin{abstract}
Multi-image benchmarks aim to test whether a multimodal large language model (MLLM) combines evidence across images, but many of their items can be answered from a single image.
Across about 33k items from 19 public benchmarks, our audit and a two-stage single-image filter find that one image or the question alone settles 24.2\% of them.
On these items, MLLMs recover from one image a median 78\% of the accuracy that all images add over the text alone, so these items reward answering from one image, which we call the \emph{single-image shortcut}.
MLLMs take this shortcut even on items where one image is not enough. On such items, most of their wrong answers repeat an answer they give from one image, at a rate well above chance.
We build \bench, to our knowledge the largest audited multi-image benchmark, from these 19 benchmarks by keeping their items that need more than one image.
Its 18.7k items, 2.0k of them corrected or rewritten, share one format, one scorer and seven task categories.
We also release \bench-Train with 49k training items and \benchlite, a hard subset whose items we manually verify one by one. On \benchlite, the strongest proprietary MLLM scores 68\%.
Reinforcement learning with verifiable rewards (RLVR), now common in MLLM post-training, rewards the shortcut as well, because a correct answer read off one image earns as much reward as one that combines images.
In line with this, existing RLVR objectives add little to the \bench score.
To stop rewarding the shortcut in training as \bench does in evaluation, we propose cross-image credit assignment (\cica), an extension of RLVR that moves the credit toward the tokens that need more than one image.
On three open-source MLLMs, \cica raises the \bench score by 2.6 to 6.0 points, more than twice the gain of each general and vision-specific RL objective trained on the same data.
Our datasets, code and models are publicly available at \url{https://github.com/chanseul01/ASTERISM}.\looseness=-1
\end{abstract}

\section{Introduction}\label{sec:intro}
People make sense of the world by relating many views of it.
Each photograph shows only what one camera saw at one instant, so we look at a room from several sides to learn its layout, compare scans of a patient taken months apart to see how an illness has changed, and leaf through a long report to find the page that answers a question.
One image is not enough in any of these cases.
Multimodal large language models (MLLMs) now take interleaved images as input for such tasks~\cite{li2025llavainterleave,jiang2024mantis,bai2025qwen25vl}, yet MLLMs that understand single images well still struggle to link cues across images~\cite{chen2025mico,das2026mimic,guo2026ssmnbench,guo2026cvbench,tian2025position}.
Multi-image benchmarks~\cite{liu2021marvl,wang2025muirbench,zhao2024mirb,jiang2024mantis,song2024milebench,kazemi2024remi,yang2026mmsibench,wang2025mvmath,zhang2025vlm2bench,li2026mmrlife,zhang2024qbenchplus,ye2025m4bench,li2026dicobench,cheng2025mmrb} measure this skill and guide how MLLMs are built and trained~\cite{chen2025mico}. A multi-image item, however, tests the skill only if answering it requires more than one of its images, to combine evidence across them or to find the right image among them.
We ask how many benchmark items meet this condition and whether MLLMs answer from more than one image.\looseness=-1

Many benchmark items do not meet this condition, and MLLMs lean on one image even on the items that do.
We call answering a multi-image item from one of its images the \emph{single-image shortcut}, a form of shortcut learning~\cite{geirhos2020shortcut}.
It is the multi-image counterpart of the language prior, under which visual question answering models answer from the question alone~\cite{goyal2017making,agrawal2018dont} and MLLMs hallucinate content that the text makes likely~\cite{leng2024vcd,guan2024hallusionbench}.
Unlike the language prior, it uses visual evidence, so removing every image~\cite{chen2024mmstar,tong2024cambrian} does not expose it, while showing one image at a time does.
Across 32,688 items of 19 public benchmarks, our audit by two MLLMs with a human adjudicator and a two-stage single-image filter find that one image or the question alone settles 24.2\% of them, and more than a quarter of the items in eight of the benchmarks, among them MuirBench, Q-Bench2 and MIRB (\cref{sec:problems,tab:supp-ledger}).\footnote{Figures, tables and pages numbered with an S are in the supplementary material.}
These items reward the shortcut.
With one image visible and the others grayed out, 23 open-source MLLMs recover a median 78\% of their all-image gain, the accuracy that all images add over the text alone, on the items the audit removes, but 18\% on the items it keeps (\cref{sec:shortcut,fig:teaser}).\looseness=-1

\begin{figure*}[t]
  \centering
  \includegraphics[width=\textwidth]{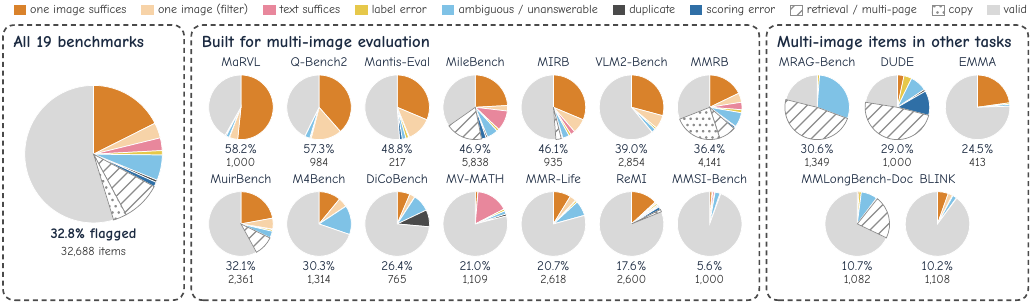}
  \caption{\textbf{One image or the question alone answers many multi-image items.} Verdicts of the audit and the single-image filter for the items of each source benchmark. The percentage below each chart is the share of flagged items, those that one image or the question answers or that have a defect. Hatched slices mark retrieval and multi-page document items that one image or the question answers, and dotted slices mark copies of other sources' items. The percentage counts neither kind of item, even when such items are removed.}
  \label{fig:audit}
\end{figure*}
MLLMs take the shortcut even on items that need more than one image.
On such items, 75--85\% of the wrong answers of Qwen2.5-VL-7B, Qwen3.5-9B and GLM-4.1V-9B-Thinking repeat an answer that the model gives from a single image, where chance would give about half (\cref{fig:teaser}).\looseness=-1

From the 19 benchmarks, we keep the items that need more than one image and build \bench from them (\cref{sec:bench}). They include the multi-image items of benchmarks for visual perception, multimodal reasoning, retrieval-augmented answering and documents~\cite{fu2024blink,hao2025emma,hu2025mragbench,vanlandeghem2023dude,ma2024mmlongbenchdoc}.
Beyond the shortcut items, the audit finds label or scoring errors, ambiguous or unanswerable questions and duplicates. In all, 32.8\% of the items do not test multi-image reasoning as posed.
\bench corrects the label, answer format or text of 569 items and adds 1,406 rewritten items whose options each pair evidence from different images, and human reviewers check every corrected label and every rewrite.
The sources differ in format, prompt and scorer and name the same skill differently, so \bench puts every item under one format, one scorer and seven task categories (\cref{sec:scorer-bench}).\looseness=-1

Reinforcement learning (RL) with verifiable rewards (RLVR), used in post-training, rewards the shortcut as well.
Through algorithms such as group relative policy optimization (GRPO)~\cite{shao2024deepseekmath} and DAPO~\cite{yu2025dapo}, RLVR has improved LLMs in mathematics and code~\cite{guo2025deepseekr1} and MLLMs in visual perception and reasoning~\cite{liu2025visualrft,huang2026visionr1}.
Recent MLLMs use it in post-training~\cite{hong2025glm41v,xiaomi2025mimovl,qwen2025qwen3vl,bytedance2025seed15vl}.
Its reward checks only the final answer, and every token of a response receives the same advantage, so the update does not tell which tokens earned the reward.
On a multi-image item, a correct answer read off one image therefore earns as much credit as one that combines images, and the reward does not push a model that already takes the shortcut (\cref{sec:shortcut}) toward combining them.
Consistent with this, existing RLVR objectives add little to multi-image accuracy.
Twenty-two public post-trained variants of Qwen2.5-VL-7B, most of them trained with RL, gain at most 1.6 points on \bench (\cref{tab:supp-leaderboard,tab:supp-basedelta}), and five RLVR objectives, two general and three designed for visual inputs, add little to any backbone that we train (\cref{tab:main}).\looseness=-1

Many multi-image items and RLVR thus both reward the shortcut.
\bench removes from evaluation the items that reward it, and in training, cross-image credit assignment (\cica) moves the credit of RLVR toward the tokens that need more than one image.
The audit asks of each item whether one image already gives the answer, and \cica asks the same question of each token of a response. \cica compares the policy's own likelihood of the token under all images with its likelihood under the best single image, the one image under which the token is most likely.
\cica keeps RLVR and its verifiable reward and changes only how that reward is credited to tokens (\cref{sec:method}).
Other RLVR methods weight or mask tokens by their visual activation, grounding or attention to key video frames~\cite{wang2026vgpo,du2026appo,cho2027grit} or push the policy away from its predictions without key visual cues~\cite{yu2026cfpo}.
Multi-image RL methods add tasks such as image contrast, jigsaw puzzles or cross-image grounding~\cite{chen2025mico,wu2026visualjigsaw,zheng2026mirgrl} (related work in \cref{sec:supp-related}).
\cica adds no task, parameter or rollout. Among the tokens that depend on the images, it upweights those that the best single image cannot explain.\looseness=-1

We make three contributions.
\begin{itemize}
\item We show that many multi-image items reward the single-image shortcut and that MLLMs take it. One image or the question alone settles 24.2\% of the items of 19 public benchmarks, and even on items that need more than one image, the wrong answers of each of 31 open-source MLLMs repeat a one-image answer more often than a matched chance level (\cref{sec:problems}).\looseness=-1
\item We release \bench, to our knowledge the largest audited multi-image benchmark, with 18.7k items that need more than one image, 2.0k of them corrected or rewritten, and evaluate 58 open-source MLLMs on it (\cref{sec:bench,sec:eval}).
\benchlite is a subset of 400 hard \bench items that we manually verify one by one. On \benchlite, the strongest proprietary MLLM scores 68\%, and the ranking of open-source MLLMs closely follows their ranking on \bench (\cref{sec:lite,tab:lite,tab:supp-lite}).
We also release \bench-Train, 49k training items with verifiable answers built from public datasets and images, with near-duplicates of \bench images removed (\cref{sec:train-set}).\looseness=-1
\item We propose \cica, an extension of RLVR that moves credit from the tokens of a response that one image explains to those that need more than one image. We show that it improves three backbones by 2.6 to 6.0 points, more than any of the five RL objectives trained on the same data, and that its gains over them are larger with all images visible than with one. Although it trains only on multi-image items, it keeps or raises the backbones' scores on ten single-image benchmarks (\cref{sec:method,sec:results}).\looseness=-1
\end{itemize}

\section{Many Multi-Image Items Reward a Single-Image Shortcut}\label{sec:problems}
\subsection{One image or the question alone answers more than one item in five}\label{sec:audit}
One image settles 21.2\% of the 32,688 items of 19 public benchmarks and the question alone another 3.0\%, counting the audit and both stages of the single-image filter (\cref{sec:supp-audit,sec:supp-ledger}), so these items reward single-image or text-only answering.
Most of these benchmarks were built for multi-image evaluation, and the others contain multi-image items within other tasks, such as retrieval-augmented answering or document understanding (\cref{fig:audit}).

Before the audit, the first stage of the filter removes 545 items that four open-source MLLMs from four model families~\cite{abouelenin2025phi4mini,wu2024deepseekvl2,kimi2025kimivl,deitke2025molmo} all answer correctly from one of the item's images under every rotation of the options~\cite{liu2024mmbench} (\cref{sec:supp-audit}).
Two MLLMs from different model families, Claude Fable~5~\cite{anthropic2026fable5} and GPT-5.6-Sol~\cite{openai2026gpt56sol}, audit each remaining item independently.
Each MLLM answers twice, first from the question and options alone and then with the numbered images, and compares both answers with the ground truth. It then names the single image that suffices, if any, and assigns one of eight verdicts by a fixed priority (\cref{sec:supp-audit}).
A human adjudicator resolves disagreements, and human reviewers check every label correction and a stratified sample of the verdicts (\cref{sec:human}).\looseness=-1

The share of these items varies across sources (\cref{tab:supp-ledger}).
One image or the question alone settles 55.8\% of MaRVL, whose statements are checked against a pair of images, and 28--54\% of MuirBench, Q-Bench2 and MIRB, while the question text settles 16.1\% of the multi-image math problems of MV-MATH.
Retrieval and multi-page document questions, in which finding the right image or page is the task, do not count toward these shares when one image or the question answers them, even when the audit removes them (hatched in \cref{fig:audit}; rule in \cref{sec:supp-groups}), so the shares are conservative.
Beyond these shortcut items, the audit finds items that are ambiguous or unanswerable (6.0\%), carry a label or scoring error (2.1\%) or duplicate another item (0.5\%). In all, 32.8\% of the items do not test multi-image reasoning as posed.\looseness=-1

\subsection{MLLMs take the single-image shortcut}\label{sec:shortcut}
\begin{figure*}[t]
\begin{minipage}[t]{\columnwidth}\vspace{0pt}
  \centering
  \includegraphics[width=\linewidth]{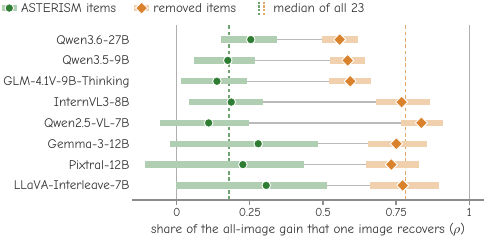}
  \caption{\textbf{One image recovers most of the all-image gain on the items that the audit removes and little on \bench items.} Recovery ratio $\rho$ on multiple-choice \bench items and removed items for the three backbones trained in \cref{sec:cica-results} and the most accurate of the 23 models in each family. Bands show 95\% intervals, and dashed lines mark the median over all 23 models.}
  \label{fig:reliance}
\end{minipage}\hfill
\begin{minipage}[t]{\columnwidth}\vspace{0pt}\inlinetables
\begin{table}[t]
\centering\footnotesize
\definecolor{secband}{HTML}{E3E8F0}
\setlength{\tabcolsep}{2pt}
\caption{\textbf{Wrong answers follow one image even on \bench items.} On the same multiple-choice items as \cref{fig:reliance}, Match is the share of each model's wrong option picks that repeat a one-image answer, and Chance (gray) is the same share for a wrong option drawn at random. The right block leaves out wrong answers equal to the text-only answer, and $\Delta$ is Match minus Chance in this block.}
\label{tab:match}
\newcommand{\mvc}[2]{\textbf{#1}\,/\,\textcolor{black!55}{#2}}
\begin{tabular*}{\linewidth}{@{\hspace{3pt}}l@{\extracolsep{\fill}}ccr@{\hspace{3pt}}}
\toprule
& \multicolumn{1}{c}{All wrong answers} & \multicolumn{2}{c}{Without text-only answers} \\
\cmidrule(lr){2-2}\cmidrule(l{4pt}r{3pt}){3-4}
Items & \mvc{Match}{Chance} & \mvc{Match}{Chance} & $\Delta$ \\
\midrule
\rowcolor{secband}\multicolumn{4}{@{\hspace{3pt}}l}{\textbf{Qwen2.5-VL-7B}} \\
\bench & \mvc{84.6}{46.2} & \mvc{68.7}{46.0} & \textbf{+22.7} \\
Removed & \mvc{81.5}{47.6} & \mvc{77.6}{49.0} & \textbf{+28.6} \\
\rowcolor{secband}\multicolumn{4}{@{\hspace{3pt}}l}{\textbf{GLM-4.1V-9B-Thinking}} \\
\bench & \mvc{81.1}{53.4} & \mvc{69.3}{51.7} & \textbf{+17.6} \\
Removed & \mvc{80.2}{51.1} & \mvc{73.8}{49.4} & \textbf{+24.4} \\
\rowcolor{secband}\multicolumn{4}{@{\hspace{3pt}}l}{\textbf{Qwen3.5-9B}} \\
\bench & \mvc{75.4}{48.8} & \mvc{65.7}{48.3} & \textbf{+17.4} \\
Removed & \mvc{81.0}{51.2} & \mvc{73.6}{44.5} & \textbf{+29.1} \\
\bottomrule
\end{tabular*}
\end{table}

\end{minipage}
\end{figure*}
\looseness=-1 We test whether MLLMs take this shortcut by hiding images from them.
We leave the prompt and the number, order and size of the images unchanged and turn hidden ones uniform gray, so the input keeps the layout of the item and loses only the content of the hidden images.
Each model is tested under three conditions. Under \emph{text only}, every image is hidden; under \emph{one image}, a single image is visible and accuracy is averaged over the choice of that image; under \emph{all images}, no image is hidden.
The \emph{all-image gain} is the accuracy that all images add over the text alone, and the recovery ratio
\begin{equation}
  \rho = \frac{\mathrm{Acc}_{\text{one}} - \mathrm{Acc}_{\text{text}}}{\mathrm{Acc}_{\text{all}} - \mathrm{Acc}_{\text{text}}} \label{eq:rho}
\end{equation}
is the share of the all-image gain that one image recovers.
A ratio near one means that one image gives about the accuracy of all images, and a ratio near zero means that one image adds little over the text alone.
We test 23 open-source MLLMs of the Qwen, InternVL, LLaVA, Gemma, Pixtral and GLM families, among them models trained for multi-image tasks. The test uses the multiple-choice items among 1,000 \bench items and 1,000 \emph{removed items}, items that the audit removed because one image or the question alone answers them. Every item has two to six images.\looseness=-1

\paragraph{One image recovers most of the all-image gain on removed items.}
The median $\rho$ over the 23 models is 0.78 on removed items and 0.18 on \bench items, a gap with a 95\% interval of [0.49, 0.74] over items and models, and every model recovers more on the removed items (\cref{fig:reliance,fig:supp-reliance-all}).
The gap holds when images of other items replace the gray ones, with median ratios of 0.65 on removed items and 0.14 on \bench items, and again every model recovers more on the removed items (\cref{tab:supp-swaprho}).
The removed items therefore reward the shortcut, and \bench items reward it little.\looseness=-1

\paragraph{Wrong answers follow one image.}
Even on items that need more than one image, most wrong answers repeat an answer the model gives from one image.
For each multiple-choice \bench item in this test, we collect a model's answers with each single image visible, its one-image answers, and check whether its wrong answer with all images visible repeats one of them.
For Qwen2.5-VL-7B, Qwen3.5-9B and GLM-4.1V-9B-Thinking, 75--85\% of the wrong answers repeat a one-image answer, against 46--53\% for a wrong option drawn at random (\cref{tab:match}). This chance level is the mean over items of $|S_i\cap W_i|/|W_i|$, with $S_i$ the set of one-image answers and $W_i$ the wrong options of item $i$. On two-option items the two shares coincide by construction, which dilutes this gap, and on items with three or more options it widens to 30.0--41.0 points (\cref{sec:supp-res-rho}).
The gap remains for each model among the wrong answers that differ from the text-only answer, with that answer left out of $W_i$.

A second, matched chance level uses the one-image answers of another item with the same correct letter and the same numbers of options and images.
The wrong answers of these three models match those answers in 47--53\% of cases, against 75--84\% for their own one-image answers in this comparison.
For each of 31 open-source MLLMs, the wrong answers also repeat the model's own one-image answers more often than this matched level, and the 95\% bootstrap interval of the difference lies above zero (\cref{tab:supp-match34}).\looseness=-1

Many multi-image items thus reward a single-image shortcut that MLLMs take.
We therefore build \bench from items that need more than one image (\cref{sec:bench}) and propose \cica, which moves the credit of RLVR toward the tokens of a response that one image does not explain (\cref{sec:method}).\looseness=-1

\begin{figure*}[t]
  \centering
  \includegraphics[width=\linewidth]{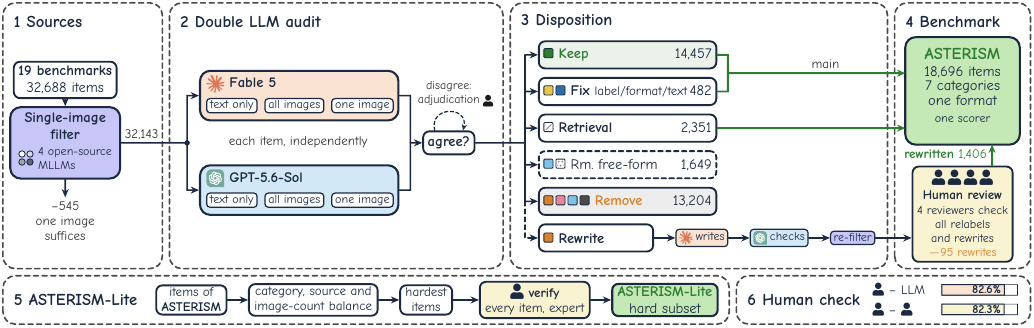}
  \caption{\textbf{Building \bench and \benchlite.} The three labels in each auditor box name its two answers, from the text alone and with all images, and the single image that suffices, if any. Each audited item is counted once, under its final disposition from Keep to Remove, and the 87 corrected retrieval items count under Retrieval. Keep, Fix and Retrieval hold the 17,290 source items of \bench, and the 1,406 rewrites are new items. Rm.\ free-form also counts copies of other sources' items, ambiguous items and source items that their rewrites replace (\cref{sec:supp-ledger}). Person icons mark the human steps. Chip colors follow \cref{fig:audit}; green marks Keep.}
  \label{fig:pipeline}
\end{figure*}
\section{\bench: One Benchmark of Items That Need More Than One Image}\label{sec:bench}
\bench keeps from each of the 19 sources only the items that need more than one image, scores every item with one scorer, and groups the items into seven task categories.
All items share one format, with the images in order, a prompt that follows the source's official instruction where one exists, and an answer format, so one run evaluates all 19 sources and reports a score per task category.
Among items with the same number of options, every letter is the correct answer about equally often.
\Cref{fig:pipeline} traces the items from the sources to the benchmark, \cref{sec:supp-ledger} gives the counts per source, and \cref{tab:compare} compares the item checks of \bench with those of other multi-image benchmarks.\looseness=-1

\subsection{Curation, rewriting and human review}\label{sec:curation}\label{sec:human}
The audit verdict decides what happens to each item.
Valid items, in which the audit finds no defect, keep their question, images and answer.
Items with a label error get the corrected answer, and items with a scoring error get an answer format that accepts the requested form.
Text-answerable, single-image-sufficient, ambiguous, unanswerable and duplicated items are removed.
Free-form items of MileBench enter as multiple-choice items only if the images support one verifiable answer and the converted item passes an independent audit (\cref{sec:supp-ledger}).\looseness=-1

Some items, mostly flagged ones, are rewritten so that their image sets still serve as multi-image tests (\cref{fig:supp-ex-rewrite}).
Claude Fable~5 writes a new question and options over the same images so that each option pairs evidence from different images, and GPT-5.6-Sol checks that exactly one option is correct. The rewrite keeps the category of its source item's operation (\cref{sec:supp-groups}).
The 1,406 rewrites pass the single-image filter and human review.
\looseness=-1

A human reviewer checks that every label the audit corrects and every rewrite has a correct answer and that each rewrite needs more than one image.
Before seeing the label, reviewers answer 94\% of the rewrites correctly and judge 98\% to need more than one image. Rewrites that fail their review or an automatic check are removed.
Two reviewers also label a 1,000-item sample stratified by verdict.
They agree with the audit verdict on 82.6\% of their answers and with each other on 82.3\% of the items, so the audit agrees with a reviewer as often as a second reviewer does (\cref{sec:supp-human}).\looseness=-1

\begin{figure}[t]
  \centering
  \includegraphics[width=\linewidth]{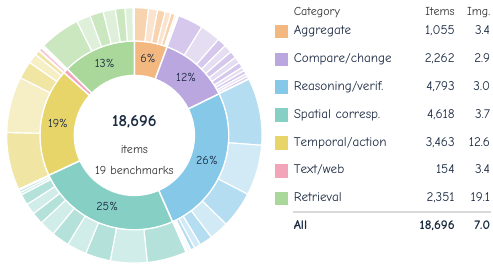}
  \caption{\textbf{\bench composition.} The inner ring shows the seven task categories and the outer ring their source benchmarks, by number of items; Img.\ is the mean number of images per item.}
  \label{fig:stats}
\end{figure}
\begin{figure*}[t]
  \centering
  \includegraphics[width=\linewidth]{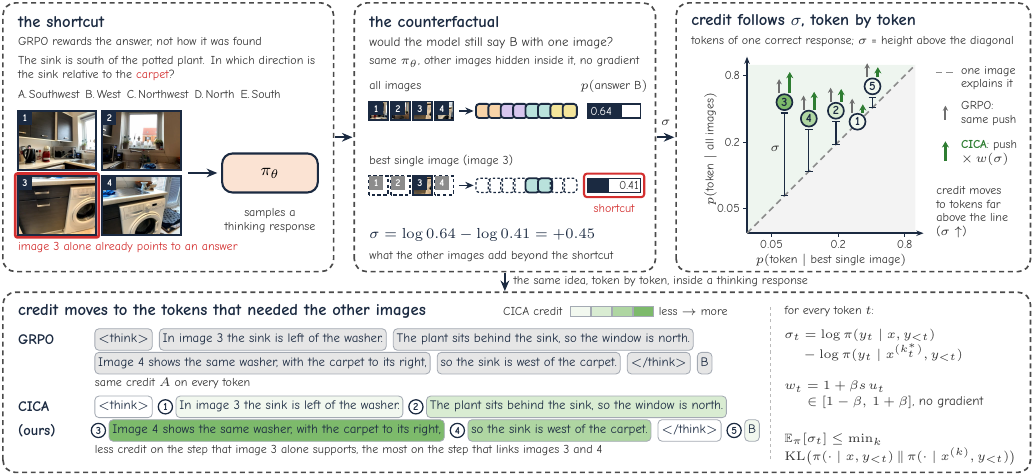}
  \caption{\textbf{Overview of our cross-image credit assignment (\cica).}}
  \label{fig:cica}
\end{figure*}
\subsection{Seven task categories and one scorer}\label{sec:scorer-bench}
The 19 sources name the same skills differently, so each item gets one of seven task categories by the cross-image operation that its answer requires, or by a rule for some subsets (\cref{sec:supp-groups}).
The categories are aggregate (counting or summarizing across images), compare/change, reasoning/verification, spatial correspondence, temporal/action, text/web and retrieval, which requires finding the image or page that holds the evidence (\cref{fig:stats}).
Items also have labels that cut across the categories, such as one of 11 cross-image operations, the relation between their images and the number of reasoning hops (\cref{tab:supp-axes,fig:supp-labels}).
The \bench score (Avg in \cref{tab:leaderboard}, entries of \cref{tab:main}) is the macro average over the seven task categories, so that the largest categories do not decide the score. \cref{tab:supp-leaderboard} also gives the micro average.\looseness=-1

\bench scores every item with one rule-based scorer that uses no LLM judge. Scoring the same generations of 21 models with the official code of eight benchmarks instead of the \bench scorer changes per-benchmark accuracy by up to 17.5 points and reverses the order of up to 25 of the 210 model pairs (\cref{sec:supp-scorer-shift}).
Each item carries an answer format taken from its prompt, and the scorer extracts the last explicit answer in that format and rejects answers that name several options. An annotator blind to its decisions agrees with it on 599 of 600 outputs (\cref{sec:supp-scorer}).\looseness=-1

\subsection{\benchlite}\label{sec:lite}

\benchlite takes the 400 \bench items that are hardest for open-source base models, under fixed quotas per task category and image count, and we manually check each item against its images and answer (\cref{sec:supp-lite}).
The ranking of 58 open-source MLLMs on \benchlite follows their ranking on \bench, with a Spearman correlation of 0.93 (\cref{tab:supp-lite}). We also evaluate proprietary MLLMs on \benchlite (\cref{tab:lite}).\looseness=-1

\subsection{\bench-Train}\label{sec:train-set}
Most of the 19 sources release only evaluation items, so we also collect training items.
\bench-Train holds 49,105 multi-image items with verifiable closed-form answers from 48 sources that span the training splits of ten public datasets, multi-image instruction collections and synthetic tasks built on public images (\cref{sec:supp-train}).
Items whose images match an \bench image by perceptual hashing and pixel correlation are removed. For sources with non-commercial licenses, we release item identifiers and build scripts.\looseness=-1

\section{Cross-Image Credit Assignment Upweights Tokens That Need More Than One Image}\label{sec:method}
{\renewcommand{\arraystretch}{1.0}\begin{table*}[t]
\centering\scriptsize\renewcommand{\arraystretch}{1.1}
\setlength{\tabcolsep}{1.9pt}
\definecolor{secband}{HTML}{E3E8F0}\def\famsep{\arrayrulecolor{black!40}\specialrule{0.4pt}{0.6pt}{2pt}\arrayrulecolor{black}}\def\ours{\oursrow}\def\m#1{\smash{$^{#1}$}}
\caption{\textbf{\bench leaderboard (representative models).} Columns are aggregate (Agg), compare/change (Cmp), reasoning/verification (Rsn), spatial correspondence (Spa), temporal/action (Tmp), text/web (Txt) and retrieval (Ret). Post-trained models (right) are grouped by the base model they start from, whose scores are repeated in gray at the top of each group; in each group the best value per column before rounding is in bold and the second best is underlined. $^{\S}$Built or fine-tuned for multi-image tasks.}
\label{tab:leaderboard}
\begin{tabular*}{\linewidth}{@{}llrrrrrrrr@{\hspace{0.6em}}@{\extracolsep{\fill}}c@{\extracolsep{0pt}}@{\hspace{0.6em}}llrrrrrrrr@{}}
\toprule
\multicolumn{10}{@{}>{\columncolor{secband}[0pt][0pt]}c@{\hspace{0.6em}}}{\textbf{Base models}} & & \multicolumn{10}{@{}>{\columncolor{secband}[0pt][0pt]}c@{}}{\textbf{Post-trained models}} \\
\cmidrule(r){1-10}\cmidrule(l{0pt}){12-21}
Model & Size & Agg & Cmp & Rsn & Spa & Tmp & Txt & Ret & Avg & & Model & Size & Agg & Cmp & Rsn & Spa & Tmp & Txt & Ret & Avg \\
\midrule
Gemma-4-12B~\cite{gemma2026gemma4} & 12B & 53.6 & 42.6 & 46.5 & 43.3 & 36.1 & 75.3 & 49.2 & 49.5 & & \multicolumn{2}{l}{\color{black!55}\textit{from} Qwen2.5-VL-7B~\cite{bai2025qwen25vl}} & {\color{black!55}45.8} & {\color{black!55}38.3} & {\color{black!55}40.0} & {\color{black!55}42.0} & {\color{black!55}36.1} & {\color{black!55}40.3} & {\color{black!55}\underline{42.2}} & {\color{black!55}40.7} \\
\noalign{\global\aboverulesep=0pt\global\belowrulesep=0pt}\arrayrulecolor{black!40}\cmidrule[0.4pt](l{0pt}r{2pt}){1-10}\arrayrulecolor{black}\noalign{\global\aboverulesep=0.4ex\global\belowrulesep=0.65ex}
GLM-4.1V-9B-Thinking~\cite{hong2025glm41v} & 9B & 64.3 & 58.2 & 66.4 & 51.8 & 44.7 & 70.1 & 55.3 & 58.7 & & Video-R1-7B~\cite{feng2025videor1} & 7B & 40.7 & 33.0 & 35.8 & 37.2 & 34.8 & 40.9 & 39.5 & 37.4 \\
\noalign{\global\aboverulesep=0pt\global\belowrulesep=0pt}\arrayrulecolor{black!40}\cmidrule[0.4pt](l{0pt}r{2pt}){1-10}\arrayrulecolor{black}\noalign{\global\aboverulesep=0.4ex\global\belowrulesep=0.65ex}
InternVL3-8B~\cite{zhu2025internvl3} & 8B & 49.5 & 43.5 & 39.9 & 37.7 & 42.8 & 44.8 & 38.3 & 42.4 & & Visual-Jigsaw-7B~\cite{wu2026visualjigsaw}\m{\S} & 7B & 44.0 & 40.6 & 40.2 & 40.7 & 36.1 & 40.3 & 40.4 & 40.3 \\
\noalign{\global\aboverulesep=0pt\global\belowrulesep=0pt}\arrayrulecolor{black!40}\cmidrule[0.4pt](l{0pt}r{2pt}){1-10}\arrayrulecolor{black}\noalign{\global\aboverulesep=0.4ex\global\belowrulesep=0.65ex}
LLaVA-Interleave-7B~\cite{li2025llavainterleave}\m{\S} & 7B & 37.0 & 37.8 & 30.4 & 29.6 & 25.9 & 35.7 & 17.0 & 30.5 & & Vision-Zero-7B~\cite{wang2026visionzero} & 7B & 44.7 & 37.1 & 40.1 & 42.2 & 36.2 & 40.3 & 41.9 & 40.4 \\
LLaVA-OneVision-2-8B~\cite{an2026llavaonevision2} & 8B & 40.4 & 40.1 & 41.5 & 38.7 & 37.0 & 42.9 & 34.2 & 39.3 & & ThinkLite-VL-7B~\cite{wang2025thinklitevl} & 7B & 45.9 & 40.0 & 40.0 & 42.3 & 36.1 & 39.0 & 41.8 & 40.7 \\
\noalign{\global\aboverulesep=0pt\global\belowrulesep=0pt}\arrayrulecolor{black!40}\cmidrule[0.4pt](l{0pt}r{2pt}){1-10}\arrayrulecolor{black}\noalign{\global\aboverulesep=0.4ex\global\belowrulesep=0.65ex}
MiMo-VL-7B~\cite{xiaomi2025mimovl} & 7B & 65.0 & 63.9 & 67.3 & 52.5 & 40.0 & 72.7 & 54.5 & 59.4 & & VLAA-Thinker-7B~\cite{chen2025vlaathinker} & 7B & 44.5 & 39.0 & 40.3 & 42.7 & 36.6 & 40.3 & 41.9 & 40.8 \\
\noalign{\global\aboverulesep=0pt\global\belowrulesep=0pt}\arrayrulecolor{black!40}\cmidrule[0.4pt](l{0pt}r{2pt}){1-10}\arrayrulecolor{black}\noalign{\global\aboverulesep=0.4ex\global\belowrulesep=0.65ex}
Pixtral-12B~\cite{agrawal2024pixtral} & 12B & 42.3 & 33.6 & 32.7 & 35.1 & 23.3 & 43.5 & 23.1 & 33.4 & & R1-ShareVL-7B~\cite{yao2025r1sharevl} & 7B & \underline{47.5} & 42.6 & \underline{42.3} & \underline{42.9} & \underline{36.8} & 39.6 & 41.4 & 41.9 \\
\noalign{\global\aboverulesep=0pt\global\belowrulesep=0pt}\arrayrulecolor{black!40}\cmidrule[0.4pt](l{0pt}r{2pt}){1-10}\arrayrulecolor{black}\noalign{\global\aboverulesep=0.4ex\global\belowrulesep=0.65ex}
Qwen2.5-VL-7B~\cite{bai2025qwen25vl} & 7B & 45.8 & 38.3 & 40.0 & 42.0 & 36.1 & 40.3 & 42.2 & 40.7 & & PeRL-7B~\cite{zhang2025perl}\m{\S} & 7B & 46.6 & \underline{44.0} & 41.7 & 40.8 & 36.8 & \underline{44.2} & 40.7 & \underline{42.1} \\
Qwen2.5-VL-32B~\cite{bai2025qwen25vl} & 32B & 54.8 & 58.8 & 52.1 & 48.2 & 37.7 & 53.9 & 45.9 & 50.2 & & \cellcolor{oursgreen}\cicaseven (ours) & \cellcolor{oursgreen}7B & \cellcolor{oursgreen}\textbf{52.1} & \cellcolor{oursgreen}\textbf{47.1} & \cellcolor{oursgreen}\textbf{44.3} & \cellcolor{oursgreen}\textbf{47.2} & \cellcolor{oursgreen}\textbf{44.0} & \cellcolor{oursgreen}\textbf{50.0} & \cellcolor{oursgreen}\textbf{42.4} & \multicolumn{1}{>{\columncolor{oursgreen}[\tabcolsep][0pt]}r@{}}{\textbf{46.7}} \\
\noalign{\global\aboverulesep=0pt\global\belowrulesep=0pt}\arrayrulecolor{black!40}\cmidrule[0.4pt](l{0pt}r{2pt}){1-10}\cmidrule[0.4pt](l{-\tabcolsep}r{0pt}){12-21}\arrayrulecolor{black}\noalign{\global\aboverulesep=0.4ex\global\belowrulesep=0.65ex}
Qwen3-VL-8B~\cite{qwen2025qwen3vl} & 8B & 50.9 & 46.8 & 44.8 & 47.0 & 40.0 & 54.5 & 47.7 & 47.4 & & \multicolumn{2}{l}{\color{black!55}\textit{from} GLM-4.1V-9B-Thinking~\cite{hong2025glm41v}} & {\color{black!55}\underline{64.3}} & {\color{black!55}\underline{58.2}} & {\color{black!55}\underline{66.4}} & {\color{black!55}\underline{51.8}} & {\color{black!55}\underline{44.7}} & {\color{black!55}\underline{70.1}} & {\color{black!55}\underline{55.3}} & {\color{black!55}\underline{58.7}} \\
Qwen3-VL-32B~\cite{qwen2025qwen3vl} & 32B & 62.1 & 56.8 & 66.0 & 50.8 & 46.0 & 64.3 & 54.1 & 57.2 & & \cellcolor{oursgreen}\cicaglm (ours) & \cellcolor{oursgreen}9B & \cellcolor{oursgreen}\textbf{69.3} & \cellcolor{oursgreen}\textbf{61.1} & \cellcolor{oursgreen}\textbf{67.1} & \cellcolor{oursgreen}\textbf{52.3} & \cellcolor{oursgreen}\textbf{46.4} & \cellcolor{oursgreen}\textbf{77.3} & \cellcolor{oursgreen}\textbf{55.4} & \multicolumn{1}{>{\columncolor{oursgreen}[\tabcolsep][0pt]}r@{}}{\textbf{61.3}} \\
\noalign{\global\aboverulesep=0pt\global\belowrulesep=0pt}\arrayrulecolor{black!40}\cmidrule[0.4pt](l{0pt}r{2pt}){1-10}\cmidrule[0.4pt](l{-\tabcolsep}r{0pt}){12-21}\arrayrulecolor{black}\noalign{\global\aboverulesep=0.4ex\global\belowrulesep=0.65ex}
Qwen3.5-9B~\cite{qwen2026qwen35} & 9B & 77.9 & 64.3 & 71.9 & 55.3 & 51.2 & 77.9 & 61.3 & 65.7 & & \multicolumn{2}{l}{\color{black!55}\textit{from} Qwen3.5-9B~\cite{qwen2026qwen35}} & {\color{black!55}\underline{77.9}} & {\color{black!55}\underline{64.3}} & {\color{black!55}\underline{71.9}} & {\color{black!55}\underline{55.3}} & {\color{black!55}\underline{51.2}} & {\color{black!55}\underline{77.9}} & {\color{black!55}\underline{61.3}} & {\color{black!55}\underline{65.7}} \\
Qwen3.6-27B~\cite{qwen2026qwen36} & 27B & 78.7 & 64.1 & 72.6 & 58.4 & 49.6 & 77.3 & 62.9 & 66.2 & & \cellcolor{oursgreen}\cicanine (ours) & \cellcolor{oursgreen}9B & \cellcolor{oursgreen}\textbf{80.0} & \cellcolor{oursgreen}\textbf{68.3} & \cellcolor{oursgreen}\textbf{80.7} & \cellcolor{oursgreen}\textbf{60.8} & \cellcolor{oursgreen}\textbf{58.4} & \cellcolor{oursgreen}\textbf{81.8} & \cellcolor{oursgreen}\textbf{64.0} & \multicolumn{1}{>{\columncolor{oursgreen}[\tabcolsep][0pt]}r@{}}{\textbf{70.6}} \\
\bottomrule
\end{tabular*}
\end{table*}

}
\begin{table*}[t]
\begin{minipage}[t]{\columnwidth}\vspace{0pt}\inlinetables
{\renewcommand{\arraystretch}{1.18}\begin{table}[t]
\centering\scriptsize
\setlength{\tabcolsep}{2.5pt}
\definecolor{secband}{HTML}{E3E8F0}
\caption{\textbf{Proprietary and open-source MLLMs on \benchlite.} Average over the seven task categories.}
\label{tab:lite}
\begin{tabularx}{\linewidth}{@{}>{\hsize=0.86\hsize}Xr@{\hspace{0.8em}}c@{\hspace{0.8em}}>{\hsize=1.14\hsize}Xr@{}}
\toprule
\multicolumn{2}{@{}>{\columncolor{secband}[0pt][0pt]}c@{\hspace{0.8em}}}{\textbf{Proprietary}} & & \multicolumn{2}{@{}>{\columncolor{secband}[0pt][0pt]}c@{}}{\textbf{Open-source, Lite only}} \\
Model & Avg & & Model & Avg \\
\midrule
Claude Opus 5.5~\cite{anthropic2026opus55} & 67.9 & & Qwen3.8-Flash-Next~\cite{qwen2026qwen38flashnext} & 60.7 \\
Gemini 3.8 Flash~\cite{google2026gemini38flash} & 64.1 & & Kimi K3.1 Pro~\cite{moonshot2026kimik31pro} & 53.2 \\
GPT-6-Astra~\cite{openai2026gpt6astra} & 62.8 & & GLM-5.3-Flash~\cite{zai2026glm53flash} & 48.9 \\
GPT-6.1-Sol~\cite{openai2026gpt61sol} & 61.6 & & Muse-Glimmer-30B~\cite{meta2026museglimmer30b} & 44.8 \\
Qwen3.8-Max~\cite{qwen2026qwen38max} & 60.9 & & MiMo-V2.6-Pro~\cite{xiaomi2026mimov26pro} & 39.3 \\
Muse Spark 1.3~\cite{meta2026musespark13} & 59.6 & & Nemotron-3.5-Super-VL~\cite{nvidia2026nemotron35supervl} & 33.6 \\
\bottomrule
\end{tabularx}
\end{table}
}
\end{minipage}\hfill
\begin{minipage}[t]{\columnwidth}\vspace{0pt}\inlinetables
{\renewcommand{\arraystretch}{1.0}\begin{table}[t]
\centering\footnotesize
\setlength{\tabcolsep}{4pt}
\caption{\textbf{\cica improves each backbone over other RL objectives on \bench.}}
\label{tab:main}
\resizebox{\linewidth}{!}{%
\begin{tabular}{lccc}
\toprule
Method & Qwen2.5-VL-7B & GLM-4.1V-9B-Thinking & Qwen3.5-9B \\
\midrule
Base & 40.7 & 58.7 & 65.7 \\
\midrule
GRPO~\cite{shao2024deepseekmath} & 41.7 & 57.8 & 66.7 \\
DAPO~\cite{yu2025dapo} & 43.0 & 58.1 & 65.7 \\
APPO~\cite{du2026appo} & 43.1 & 58.7 & 65.9 \\
VGPO~\cite{wang2026vgpo} & 42.7 & 58.7 & \underline{67.1} \\
CFPO~\cite{yu2026cfpo} & \underline{43.6} & \underline{59.0} & 64.0 \\
\oursrow \cica (ours) & \textbf{46.7} & \textbf{61.3} & \textbf{70.6} \\
\bottomrule
\end{tabular}}
\end{table}
}
\end{minipage}
\end{table*}
RLVR rewards the single-image shortcut of \cref{sec:problems} because its reward checks only the final answer, and \cica moves the credit for this reward toward the tokens of a sampled response that one image does not explain.
\cica keeps the data, the verifiable reward and the RLVR objective and changes only how advantages are divided among tokens, using the policy's own likelihoods under one image at a time (\cref{fig:cica}). \Cref{alg:cica} in \cref{sec:supp-cica} gives the full training step.\looseness=-1

\looseness=-1 GRPO~\cite{shao2024deepseekmath} samples a group of $G$ responses for each input $x$, scores each response $y_i$ with a verifiable reward $r_i$, and normalizes the rewards within the group,
\begin{equation}
  A_i=\frac{r_i-\bar r}{\mathrm{std}(r)}. \label{eq:adv}
\end{equation}
Every token $y_{i,t}$ receives this advantage in the policy term of the objective.
DAPO~\cite{yu2025dapo} also keeps one advantage per response, so neither objective separates the tokens that need more than one image from those one image explains.\looseness=-1

\subsection{Counterfactual cross-image contrast}\label{sec:cica-divergence}
\looseness=-1 For each token, \cica measures what the other images add beyond the best single image.
Let $x^{(k)}$ be the input $x$ with the visual tokens of every image except $k$ removed from the context.
The text, the image delimiters and all position indices stay as in $x$, so the model sees the same prompt and image slots and loses only the visual evidence of the other images.
Removing the other images this way reproduces the gap that gray masking shows in \cref{sec:shortcut} (\cref{tab:supp-mask}).
For each token of a sampled response $y_i$, \cica finds the single image under which that token is most likely, the best single image $k^*_{i,t}$, and contrasts the token's log-likelihood under all images with that under this image,
\begin{align}
  k^*_{i,t} &= \textstyle\arg\max_{k} \log\pi_\theta(y_{i,t}\mid x^{(k)},y_{i,<t}), \label{eq:kstar}\\
  \sigma_{i,t} &= \log\pi_\theta(y_{i,t}\mid x,y_{i,<t}) - \log\pi_\theta(y_{i,t}\mid x^{(k^*_{i,t})},y_{i,<t}). \label{eq:sigma}
\end{align}
A token counts as needing more than one image only if even the best single image cannot explain it.
In expectation over $y_{i,t}$ under $\pi_\theta$, $\sigma_{i,t}$ is at most $\min_k \mathrm{KL}\big(\pi_\theta(\cdot\mid x,y_{i,<t})\,\|\,\pi_\theta(\cdot\mid x^{(k)},y_{i,<t})\big)$ (\cref{fig:cica}).

Choosing the best single image per token keeps the contrast strict for a response that draws on a different image at each step.
With an average or a random single image as the reference, an answer read off one image would still receive a high $\sigma$, because the other images do not support it (\cref{sec:supp-cica-contrast}).
We measure $\sigma$ on the answers of two base models, with the probability of each answer renormalized over the options. Against the best single image, $\sigma$ is near zero for a correct answer to an item that one image answers (median 0.0004 for both models) and positive for a correct answer to an \bench item (medians 0.12 and 0.19).
A text-only reference ranks these answers in the opposite order, and a mean single-image reference separates them less (\cref{sec:supp-cica-contrast}).\looseness=-1

\subsection{Token-level advantage reweighting}\label{sec:cica-weights}
\looseness=-1 \cica maps $\sigma$ to a bounded weight on each token's advantage.
Over a set $\mathcal{T}_i$ of active tokens, it centers $\sigma$, squashes it and centers it again,\predisplaypenalty=0
\begin{equation}
  w_{i,t} = 1+\beta s_i u_{i,t},\ \ u_{i,t}=\tfrac12\big(h_{i,t}-\bar h_i\big),\ \ h_{i,t}=\tanh\tfrac{\sigma_{i,t}-\bar\sigma_i}{\tau}, \label{eq:w}
\end{equation}
\predisplaypenalty=10000 where $\beta\in[0,1)$ sets the strength, with $\beta=0$ giving GRPO, and $\tau$ sets the scale of the weights (\cref{tab:supp-cica-defaults}).
The indicator $s_i$ is 1 if the response has an answer in the format of its item (\cref{sec:supp-cica-answer}) and this answer is correct with $A_i>0$ or wrong with $A_i<0$, and $s_i=0$ otherwise.
$\mathcal{T}_i$ holds the active tokens of $y_i$ and of any response centered with it, which are all tokens except the answer marker and the tokens that frame a response, such as the end of turn and the thinking tags (\cref{sec:supp-cica-active}).
\cica multiplies each token's advantage by $w_{i,t}$ in the policy term of the underlying RLVR objective and sets $w_{i,t}=1$ outside $\mathcal{T}_i$ (\cref{sec:supp-cica}).

The means $\bar\sigma_i$ and $\bar h_i$ weight each response by its share of the update, so the update-weighted mean of $w$ over $\mathcal{T}_i$ is one and \cica moves credit among the tokens of $\mathcal{T}_i$ (\cref{fig:cica}).
With $s_i$, the tokens that need more than one image take a larger share of both the reward of a correct response and the penalty of a wrong one.
Since $|u|<1$ and $\beta<1$, \cica keeps the sign of every token's coefficient and bounds its weight to $[1-\beta,1+\beta]$.\looseness=-1

\section{Experiments}\label{sec:results}
\begin{figure*}[t]
  \centering
  \includegraphics[width=\textwidth]{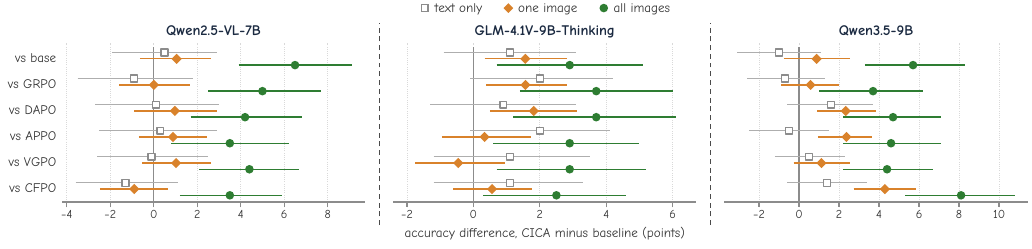}
\caption{\textbf{\cica's gains are largest with all images visible.} Accuracy of \cica minus that of each baseline on 1,000 \bench items under the gray masking of \cref{sec:shortcut}, with no image, one image or all images visible. Bars show 95\% intervals, and panel scales differ.}
  \label{fig:gain}
\end{figure*}

\subsection{Benchmarking 58 MLLMs on \bench}\label{sec:eval}
We evaluate 58 open-source MLLMs with the same prompts, each at its default image resolution or official multi-image setting (\cref{sec:supp-eval}).
\Cref{tab:leaderboard} lists representative models.\looseness=-1

Among models run in the same mode (\cref{sec:supp-eval}), accuracy rises with model size within each series, and temporal/action items score lowest of the seven categories for most models of 4B or more (\cref{tab:leaderboard,tab:supp-leaderboard}).
Across models, items whose cross-image operation merges images are harder than items whose operation compares them (\cref{tab:supp-axes}).\looseness=-1

\paragraph{Proprietary MLLMs on \benchlite.}
Claude Opus 5.5 scores highest, and among open-source MLLMs only Qwen3.8-Flash-Next reaches the proprietary range (\cref{tab:lite}).\looseness=-1

\subsection{\cica improves each backbone}\label{sec:cica-results}
We train Qwen2.5-VL-7B~\cite{bai2025qwen25vl}, GLM-4.1V-9B-Thinking~\cite{hong2025glm41v} and Qwen3.5-9B~\cite{qwen2026qwen35} from their public checkpoints for one epoch on \bench-Train (\cref{sec:supp-train}) and call the trained models \cicaseven, \cicaglm and \cicanine.
In these runs, \cica reweights the advantages of GRPO.
The five RL baselines of \cref{tab:main}, two general (GRPO, DAPO) and three vision-specific (APPO, VGPO, CFPO), each match \cica in data, the accuracy reward, prompts, group size, batch size and learning-rate schedule and otherwise follow their paper or official code (\cref{sec:supp-baselines}).
\cica adds no parameters, rollouts or backward pass. It adds only one gradient-free pass per image for each response, which takes about 9\% of a training step on Qwen2.5-VL-7B (\cref{sec:supp-cica-cost}).\looseness=-1

\paragraph{The gains exceed those of the RL baselines.}
\cica raises the \bench score of the base model by 2.6 to 6.0 points and scores above every RL baseline on all three backbones, with all paired 95\% intervals above zero (\cref{tab:main}; \cref{sec:supp-res-cica,sec:supp-res-reliability}).
On each backbone the RL baselines gain less than half as much as \cica over the base model.
\cica scores above the base model in each of the seven task categories on all three backbones (\cref{tab:leaderboard}).
Although GLM-4.1V-9B-Thinking and Qwen3.5-9B were already post-trained with RL, \cica improves both further (\cref{tab:main}).
\cicaseven scores above all other models post-trained from Qwen2.5-VL-7B (\cref{tab:supp-leaderboard,tab:supp-basedelta}).\looseness=-1

\paragraph{The gains are largest with all images visible.}
Under the gray masking of \cref{sec:shortcut}, the accuracy gap between \cica and each baseline is larger with all images visible than with one (\cref{fig:gain}).
On Qwen2.5-VL-7B, the same holds against the base model, GRPO and DAPO when the other images are hidden as in \cica or swapped for other items' images (\cref{tab:supp-maskgain,tab:supp-swapgain}).
After \cica training, a smaller share of each backbone's wrong answers repeats a one-image answer than after GRPO, by up to 7.7 points (\cref{fig:supp-shortcut}).
In a correct response of each trained model (\cref{fig:supp-tokw}), steps that relate the images carry larger weights than most steps that describe one image, as in \cref{fig:cica}.

\paragraph{Single-image scores are kept or raised.}
Although \cica trains only on multi-image items, it keeps or raises the score of each backbone on all ten single-image benchmarks (\cref{tab:supp-single}). It improves each backbone on the hallucination benchmark POPE~\cite{li2023pope} and on all three real-world benchmarks~\cite{xai2024grok15v,zhang2025mmerealworld,li2025rbench}.\looseness=-1

\subsection{\cica works best when $\sigma$ is measured against the best single image}\label{sec:ablation}
\begin{table}[t]
\centering\small
\setlength{\tabcolsep}{4pt}
\caption{\textbf{Ablations of \cica.} All $-$ one is the accuracy gain over GRPO with all images visible minus that with one image visible, on the items of \cref{fig:gain}.}
\label{tab:ablation}
\begin{tabular}{lrr}
\toprule
Training objective & \bench & All $-$ one \\
\midrule
GRPO ($\beta=0$) & 41.7 & --- \\
\cica, permuted $\sigma$ & 41.3 & $-1.0$ \\
\cica, $w=1$ on wrong responses & 44.1 & +1.5 \\
\cica, text-only reference & 44.4 & +0.8 \\
\cica, mean single-image reference & 45.6 & +2.2 \\
\oursrow \cica (best single image) & 46.7 & +5.0 \\
\bottomrule
\end{tabular}
\end{table}

We ablate \cica on Qwen2.5-VL-7B and train every row of \cref{tab:ablation} with the same trainer, training data and rewards (\cref{sec:supp-train}).
The table gives the \bench score and All~$-$~one, the accuracy gain over GRPO with all images visible minus that with one image visible.
The permuted-$\sigma$ variant shuffles the values of $\sigma$ in \cref{eq:sigma} before the weights are formed, within each set over which \cica centers them.
It falls below GRPO on \bench and on All~$-$~one, so the gain of \cica comes from the link of $\sigma$ to the images.
Setting $w=1$ on wrong responses leaves them in the sets over which \cica centers $\sigma$ and weights only the correct responses.
This variant falls below \cica on \bench and on All~$-$~one, so the weights on wrong responses add to both.
It still scores above all RL baselines of \cref{tab:main}, as does \cica with $\beta$ lowered to 0.5 or $\tau$ doubled (\cref{tab:supp-betatau}).
A text-only reference keeps about half of the \bench gain of \cica but little of the All~$-$~one gain, since it also credits the tokens of an answer read off one image.
A mean single-image reference keeps less than half of the All~$-$~one gain, so this gain is largest with the best single image as the reference (\cref{sec:supp-cica-contrast}).\looseness=-1

\section{Conclusion}\label{sec:conclusion}
Many multi-image items and RLVR both reward a single-image shortcut that MLLMs take.
One image or the question alone settles more than one in five of the items of 19 public benchmarks, and even where one image is not enough, most wrong answers of MLLMs repeat a one-image answer, well above chance.
\bench removes from evaluation the items that reward the shortcut and holds 18.7k items that need more than one image.
\benchlite takes 400 hard items that we manually verify one by one, and \bench-Train adds 49k training items with verifiable answers.
In training, \cica moves the credit of RLVR to tokens that need more than one image and gains more than the RL baselines trained on the same data.
Its gains are largest with all images visible and come without lowering single-image scores.\looseness=-1

\FloatBarrier
{\small\bibliographystyle{ieeenat_fullname}\bibliography{main}}
\clearpage
\setcounter{page}{1}\renewcommand{\thepage}{S\arabic{page}}
\appendix
\crefalias{section}{appendix}\crefalias{subsection}{subappendix}\crefalias{subsubsection}{subsubappendix}
\setcounter{table}{0}\renewcommand{\thetable}{S\arabic{table}}\renewcommand{\theHtable}{S\arabic{table}}
\setcounter{figure}{0}\renewcommand{\thefigure}{S\arabic{figure}}\renewcommand{\theHfigure}{S\arabic{figure}}
\setcounter{equation}{0}\renewcommand{\theequation}{S\arabic{equation}}\renewcommand{\theHequation}{S\arabic{equation}}
\etocdepthtag.toc{mtsupp}
\maketitlesupplementary
{\makeatletter\renewcommand\numberline[1]{#1\hskip 1em}\makeatother\etocsettagdepth{mtmain}{none}\etocsettagdepth{mtsupp}{subsection}\setcounter{tocdepth}{2}\small\tableofcontents}
\section{Related work}\label{sec:supp-related}
\subsection{Multi-image benchmarks}\label{sec:supp-related-bench}\label{sec:supp-compare}
Multi-image evaluation grew from verifying statements about pairs of photographs, as in NLVR2 and its multilingual, multicultural counterpart MaRVL~\cite{suhr2019nlvr2,liu2021marvl}, into suites that test many cross-image skills.
Mantis-Eval, MuirBench, MMIU, MIBench, MIRB and M4Bench cover comparison, ordering, correspondence, counting and multi-hop reasoning across images~\cite{jiang2024mantis,wang2025muirbench,meng2025mmiu,liu2024mibench,zhao2024mirb,ye2025m4bench}, MileBench tests long multi-image contexts~\cite{song2024milebench}, and ReMI, MMRB and MMR-Life target reasoning over several images~\cite{kazemi2024remi,cheng2025mmrb,li2026mmrlife}.
Other benchmarks focus on one kind of evidence, such as spatial relations between views in MMSI-Bench~\cite{yang2026mmsibench}, mathematics over several figures in MV-MATH~\cite{wang2025mvmath}, matching visual cues in VLM2-Bench~\cite{zhang2025vlm2bench}, differences and commonalities in DiCoBench~\cite{li2026dicobench} and low-level visual comparison in Q-Bench2~\cite{zhang2024qbenchplus}.
Benchmarks for visual perception, multimodal reasoning, retrieval-augmented answering and multi-page documents also contain multi-image items~\cite{fu2024blink,hao2025emma,hu2025mragbench,vanlandeghem2023dude,ma2024mmlongbenchdoc}.
\bench collects multi-image items from 19 of these benchmarks and keeps their official instructions where they exist.
One audit protocol tests every item against the text alone and against single images, and \bench keeps the items that need more than one image and scores all of them with one rule-based scorer (\cref{tab:compare}).

\begin{table}[t]
\centering\footnotesize
\setlength{\tabcolsep}{1.6pt}
\caption{\textbf{\bench audits public benchmarks item by item.} Entries follow each benchmark's paper (n/r, not reported), and \cref{tab:supp-ledger} counts the items that enter our pool from each source's public release.}
\label{tab:compare}
\begin{tabular}{@{}l r >{\raggedright\arraybackslash}p{0.18\linewidth} >{\raggedright\arraybackslash}p{0.20\linewidth} >{\raggedright\arraybackslash}p{0.18\linewidth}@{}}
\toprule
Benchmark & Items & Sources & Item check & Scoring \\
\midrule
BLINK~\cite{fu2024blink} & 3,807 & CV data + new & authors drop ambiguous ones & rules + GPT-3.5~\cite{openai2022chatgpt} \\
MuirBench~\cite{wang2025muirbench} & 2,600 & 7 datasets + new & 4 experts, every item & rules \\
MMIU~\cite{meng2025mmiu} & 11,698 & existing datasets & converted data checked & option match \\
MileBench~\cite{song2024milebench} & 6,440 & 21 datasets & 10\% sample; new sets all & accuracy, ROUGE-L~\cite{lin2004rouge} \\
MIRB~\cite{zhao2024mirb} & 925 & new + 3 datasets & image filter, 1 task & accuracy \\
Mantis-Eval~\cite{jiang2024mantis} & 217 & new & annotator-curated & n/r \\
MMSI-Bench~\cite{yang2026mmsibench} & 1,000 & new & 3 reviewers, every item & exact; LLM fallback \\
MV-MATH~\cite{wang2025mvmath} & 2,009 & new (K-12) & ${\geq}2$ annotators & LLM judge \\
\midrule
\bench & 18,696 & 19 benchmarks & 2 MLLMs, every item, human adjudication; 4 reviewers: relabels, rewrites; 1,000 by two & rules, no LLM \\
\bottomrule
\end{tabular}
\end{table}

\subsection{Benchmark audits and shortcuts}\label{sec:supp-related-audit}
Audits of test sets find label errors that change model rankings~\cite{northcutt2021pervasive}, and re-annotations of language and document benchmarks find wrong labels and ambiguous or unanswerable questions~\cite{gema2025mmluredux,vendrow2025platinum,zhang2026mmlongbenchdocv2}.
Statistics of model responses can flag such items for expert review~\cite{truong2025fantasticbugs}.
In visual question answering, questions whose answers follow from the text alone led to benchmarks that pair similar images with different answers or change the answer priors between training and test~\cite{goyal2017making,agrawal2018dont}, and to training on counterfactual samples~\cite{chen2020css}.
Many items of MLLM benchmarks can also be answered without the image~\cite{chen2024mmstar,tong2024cambrian}, and MMStar, MMMU-Pro and NaturalBench select, filter or pair items so that the image is needed~\cite{chen2024mmstar,yue2025mmmupro,li2024naturalbench}.
In multi-hop reading comprehension, models that read one paragraph or one sentence at a time answer many questions written to need two paragraphs~\cite{min2019compositional,chen2019understanding}, and adversarial documents that offer a second answer through the same shortcut lower the accuracy of strong models~\cite{jiang2019avoiding}.
A related effect appears in video benchmarks, where a single frame or frames in the wrong order often suffice~\cite{buch2022revisiting,lei2023singleframe,shangguan2025tomato}.
CVBench and SSMNBench build multi-camera benchmarks of human scenes whose items need several views, and they analyze the errors of MLLMs that rely on one view~\cite{guo2026cvbench,guo2026ssmnbench}.
\bench brings these tests to the multi-image items of 19 public benchmarks.
It checks each item that it draws from them against the text alone and against single images, together with label, ambiguity, duplicate and scoring checks (\cref{sec:supp-audit}).
\cica applies the same one-image comparison to the tokens of a response during training.

\subsection{RL with verifiable rewards for LLMs and MLLMs}\label{sec:supp-related-rlvr}
RLVR replaces a learned reward model with a rule that checks the final answer, such as an exact match or a test case.
T\"{u}lu~3 named the approach and applied it to mathematics and instruction following~\cite{lambert2024tulu3}, and DeepSeekMath introduced GRPO, which normalizes the rewards within a group of responses to the same prompt and needs no value model~\cite{shao2024deepseekmath}.
DeepSeek-R1 and Kimi k1.5 scaled RL with answer-checking rewards to long chain-of-thought reasoning in mathematics and coding~\cite{guo2025deepseekr1,kimi2025kimik15}.
For MLLMs, RLVR has improved visual perception~\cite{liu2025visualrft} and multimodal reasoning~\cite{huang2026visionr1,tan2025reasonrft,wang2025vlrethinker}, and recent MLLMs include an RL stage with verifiable rewards in their post-training, often together with rewards from preference models~\cite{hong2025glm41v,xiaomi2025mimovl,qwen2025qwen3vl,bytedance2025seed15vl,kimi2025kimivl}.
The reward checks the final answer, and GRPO and DAPO~\cite{yu2025dapo} give every token of a response the same advantage, so the update does not separate an answer read off one image from one that combines images.
\cica keeps this reward and changes how it is spread over the tokens (\cref{sec:method}).\looseness=-1

\subsection{Token-level credit assignment}\label{sec:supp-related-credit}
With one reward per response, credit assignment decides how much of it each token receives.
PPO learns a value function that gives each token its own advantage~\cite{schulman2017ppo}, and VinePPO replaces the value function with Monte Carlo estimates from rollouts that restart at intermediate steps~\cite{kazemnejad2025vineppo}.
Process reward models, learned from human step labels~\cite{lightman2024verify} or from outcome labels alone~\cite{cui2026prime}, score intermediate steps.
In RLHF, the attention weights of the reward model can spread the sequence reward over tokens~\cite{chan2024dense}, and in RLVR, restricting the update to the high-entropy tokens of a response matches or exceeds updating all tokens~\cite{wang2025beyond8020}.
Counterfactual credit assignment measures what an action contributes by comparing it with an alternative, for example by marginalizing out one agent's action in a multi-agent critic~\cite{foerster2018coma} or by conditioning a baseline on hindsight information about the trajectory~\cite{mesnard2021counterfactual}.
\cica takes its counterfactual from the images of the input.
It keeps the sampled response fixed, changes which images the policy sees, and compares the likelihood of each token under all images with that under the best single image, so it needs no value model, no reward model and no extra rollouts.\looseness=-1

\subsection{RL toward visual evidence}\label{sec:supp-related-visual}
Several methods change the RLVR update of MLLMs so that it depends on the visual input.
VGPO locates visual cues by similarity and reweights advantages toward tokens with high visual activation and toward trajectories that accumulate more of it~\cite{wang2026vgpo}, and APPO gives token-level rewards to tokens that attend to the same key video frame across the responses of a group~\cite{du2026appo}. Another method masks the gradients of visually misgrounded tokens in text-rich reasoning~\cite{cho2027grit}. VPPO scales the advantage of each response by its tokens' mean dependence on the image, measured against a randomly masked image, and updates only the most dependent tokens~\cite{huang2026vppo}.
CFPO regularizes the policy to differ from its predictions in a counterfactual state in which key visual cues are suppressed~\cite{yu2026cfpo}, PAPO adds a loss that pushes the policy's outputs for the original image away from those for a masked one~\cite{wang2026papo}, and NoisyRollout mixes rollouts from clean and distorted images~\cite{liu2025noisyrollout}.
Other methods train the model to anchor its reasoning steps to image coordinates or boxes~\cite{sarch2025vigorl,fan2025grit}, or reward perception directly, by judging the visual content of a response against reference annotations~\cite{xiao2026perceptionr1} or by checking whether a self-generated visual description suffices to answer~\cite{li2026visionsr1}.
Counterfactual visual inputs also appear outside RL, in VQA training samples that mask critical objects~\cite{chen2020css}.
\cica defines its counterfactual over the images of a multi-image input.
It uses this counterfactual to reweight the advantage of GRPO, with no extra loss or reward.

\subsection{Training for multi-image inputs}\label{sec:supp-related-multi}
Interleaved instruction data such as Mantis-Instruct and M4-Instruct teach MLLMs to take several images~\cite{jiang2024mantis,li2025llavainterleave}, Migician trains grounding across images on a large instruction set~\cite{li2025migician}, and Das et al.~\cite{das2026mimic} compose single-image annotations into multi-image training examples and mask attention for multi-image inputs.
RL methods add multi-image tasks with verifiable answers, such as telling two augmented views of an image from a similar image~\cite{chen2025mico}, solving jigsaw puzzles over images, videos and 3D data~\cite{wu2026visualjigsaw}, grounding objects across images~\cite{zheng2026mirgrl,bai2025univgr1} and answering spatial questions over several views~\cite{wang2026mindcube,oi2026hatch}.
\cica adds no task.
It trains on multi-image items with their own verifiable answers and changes only how the reward of a response is credited to its tokens.\looseness=-1

\section{Construction of \bench}\label{sec:supp-build}
\subsection{Audit protocol}\label{sec:supp-audit}
\paragraph{Verdicts.}
Each auditor assigns one of the eight verdicts below, and when several apply, the first in this order wins.
\emph{Label error} means that the ground truth is false given the images and the question, and the auditor can name the correct answer.
\emph{Unanswerable} means that no answer follows from the inputs.
\emph{Scoring error} means that the label is right but a correct answer in the requested format is rejected by the official scorer.
\emph{Ambiguous} means that two or more answers are defensible.
\emph{Duplicate} means that the item repeats the images and question of another item.
\emph{Text-answerable} means that the auditor determines the answer confidently from the text alone, and it matches the ground truth.
\emph{Single-image-sufficient} means that one image alone answers the question, and the auditor names the image.
\emph{Valid} means that none of the above applies.

\paragraph{Procedure.}
Two MLLMs, Claude Fable~5~\cite{anthropic2026fable5} and GPT-5.6-Sol~\cite{openai2026gpt56sol}, annotate every audited item independently.
They answer each item twice, first from the question and the options alone and then with the numbered images, and write one verdict, both answers, any image that suffices, a confidence rating and a short reason per item.
The two MLLMs agree on whether to keep or remove an item in 92.2\% of cases.
When their verdicts differ, a human adjudicator reads both verdicts, answers and reasons next to the images and makes the final decision.
Escalated verdicts (those reporting a defect in the label or the item, and valid verdicts given with low confidence) also go to a human adjudicator.
Every label correction goes to human review (\cref{sec:supp-human}).
\looseness=-1

\paragraph{Single-image filter.}
Both stages of the single-image filter use one committee of four open-source MLLMs from four model families, Phi-4-multimodal-instruct~\cite{abouelenin2025phi4mini}, DeepSeek-VL2-Small~\cite{wu2024deepseekvl2}, Kimi-VL-A3B-Instruct~\cite{kimi2025kimivl} and Molmo-7B-O~\cite{deitke2025molmo}.
The first stage runs on source items before the audit and on each rewrite before human review, and the second stage runs once labels, formats and option orders are final, on all items including the rewrites.
The committee shares no model family with the backbones we train and no model with the leaderboard, so no evaluated model takes part in the filter.
For every image of an item, each member answers from that image alone in one presentation per cyclic rotation of the $C$ options, as in CircularEval~\cite{liu2024mmbench}.
In each presentation, the member receives the evaluation prompt of the item with only that image and decodes greedily, and the rule-based scorer of \cref{sec:supp-scorer} checks its answer.
Items whose options refer to other options, such as ``none of the above'', keep their original order and are presented once to each member.
An item is removed if, for some image, every member answers correctly in every presentation and a one-sided binomial test of the committee's correct presentations against the chance rate $1/C$ is significant.
The test is Bonferroni-corrected over the item's images, with Benjamini--Hochberg control of the false discovery rate at 5\% over all items of the stage~\cite{benjamini1995fdr}.
For items without options the chance rate is zero, so all four members must answer correctly from the same image.\looseness=-1

\subsection{Per-source item counts}\label{sec:supp-ledger}
\begin{table*}[t]
\centering\footnotesize
\setlength{\tabcolsep}{3pt}
\caption{\textbf{Where every source item goes.} For each source benchmark, the columns count all items (Items), items removed by the single-image filter before the audit (Filter), audited items (Audited), items kept in the six task categories other than retrieval (Main) or in the retrieval category (Ret.), items removed by a first-pass audit verdict (Rm.\ verdict) or by a rule (Rm.\ rule), free-form items that fail the conversion, its audit or the first filter stage, together with copies, ambiguous items and source items that their rewrites replace (Rm.\ free), items removed by the second audit pass (Rm.\ 2nd), items removed by the second stage of the single-image filter because one image answers them (1-img), and rewritten items derived from the source that pass human review and the final steps (R). Each row satisfies Items = Main + Ret. + Rm.\ verdict + Rm.\ rule + Rm.\ free + Rm.\ 2nd + 1-img + Filter. R items are new items and are not part of this sum. Main includes the 83 converted free-form items of MileBench. Rules remove the single-page items of DUDE, other items with free-form answers, control tasks and a subtask whose labels failed the audit. The last three columns give the percentage of the row's items that one image or the question alone settles, as judged by the two audit passes (Audit), after adding the items that the first stage of the single-image filter removes (+Filter), and after further adding the 1-img items (+1-img). Retrieval and multi-page document items, whose task is to find the image or page that holds the answer, and copies of other sources' items are not counted. A dash marks zero.}
\label{tab:supp-ledger}
\begin{tabular}{@{}lrrrrrrrrrrr@{\hspace{6pt}}rrr@{}}
\toprule
 & & & & & & & & & & & & \multicolumn{3}{c}{One image or text (\%)} \\
\cmidrule(l){13-15}
Source & Items & Filter & Audited & Main & Ret. & Rm.\ verdict & Rm.\ rule & Rm.\ free & Rm.\ 2nd & 1-img & R & Audit & +Filter & +1-img \\
\midrule
MileBench~\cite{song2024milebench} & 5,838 & 2 & 5,836 & 1,508 & 318 & 2,522 & 260 & 519 & 382 & 327 & 218 & 34.3 & 34.3 & 37.0 \\
MMRB~\cite{cheng2025mmrb} & 4,141 & 112 & 4,029 & 1,069 & 191 & 1,373 & 68 & 249 & 955 & 124 & -- & 21.6 & 23.4 & 26.2 \\
VLM2-Bench~\cite{zhang2025vlm2bench} & 2,854 & 37 & 2,817 & 1,728 & -- & 521 & -- & 29 & 381 & 158 & 82 & 29.2 & 30.5 & 36.0 \\
MMR-Life~\cite{li2026mmrlife} & 2,618 & 81 & 2,537 & 2,093 & -- & 394 & 1 & 30 & 16 & 3 & 158 & 9.0 & 12.1 & 12.2 \\
ReMI~\cite{kazemi2024remi} & 2,600 & -- & 2,600 & 1,956 & -- & 213 & -- & 259 & 153 & 19 & 273 & 13.3 & 13.3 & 14.0 \\
MuirBench~\cite{wang2025muirbench} & 2,361 & 121 & 2,240 & 979 & 226 & 710 & 227 & 22 & 15 & 61 & -- & 22.4 & 26.3 & 27.8 \\
MRAG-Bench~\cite{hu2025mragbench} & 1,349 & 83 & 1,266 & -- & 291 & 947 & 4 & -- & 5 & 19 & -- & 0.2 & 0.2 & 0.2 \\
M4Bench~\cite{ye2025m4bench} & 1,314 & -- & 1,314 & 865 & -- & 150 & 230 & 3 & 4 & 62 & -- & 11.0 & 11.0 & 15.7 \\
MV-MATH~\cite{wang2025mvmath} & 1,109 & -- & 1,109 & 857 & -- & 206 & 1 & 39 & 5 & 1 & 36 & 17.3 & 17.3 & 17.4 \\
BLINK~\cite{fu2024blink} & 1,108 & -- & 1,108 & 879 & -- & 62 & -- & 141 & 1 & 25 & 145 & 5.4 & 5.4 & 7.7 \\
MMLongBench-Doc~\cite{ma2024mmlongbenchdoc} & 1,082 & -- & 1,082 & -- & 963 & 91 & 1 & 24 & -- & 3 & -- & 0.8 & 0.8 & 0.8 \\
DUDE~\cite{vanlandeghem2023dude} & 1,000 & -- & 1,000 & -- & 334 & 93 & 529 & 35 & -- & 9 & -- & 3.3 & 3.3 & 3.3 \\
MaRVL~\cite{liu2021marvl} & 1,000 & -- & 1,000 & 316 & -- & 522 & -- & 120 & -- & 42 & 147 & 51.6 & 51.6 & 55.8 \\
MMSI-Bench~\cite{yang2026mmsibench} & 1,000 & -- & 1,000 & 928 & -- & 19 & 1 & 37 & 12 & 3 & 23 & 2.1 & 2.1 & 2.4 \\
Q-Bench2~\cite{zhang2024qbenchplus} & 984 & 64 & 920 & 391 & -- & 405 & 2 & 33 & 1 & 88 & 102 & 38.6 & 45.1 & 54.1 \\
MIRB~\cite{zhao2024mirb} & 935 & 45 & 890 & 432 & 4 & 345 & 2 & 50 & 34 & 23 & 177 & 33.6 & 38.4 & 40.9 \\
DiCoBench~\cite{li2026dicobench} & 765 & -- & 765 & 547 & -- & 119 & -- & 46 & 32 & 21 & 22 & 6.4 & 6.4 & 9.2 \\
EMMA~\cite{hao2025emma} & 413 & -- & 413 & 309 & -- & 96 & 4 & 4 & -- & -- & -- & 22.8 & 22.8 & 22.8 \\
Mantis-Eval~\cite{jiang2024mantis} & 217 & -- & 217 & 82 & 24 & 69 & 3 & 9 & 3 & 27 & 23 & 31.8 & 31.8 & 44.2 \\
\midrule
Total & 32,688 & 545 & 32,143 & 14,939 & 2,351 & 8,857 & 1,333 & 1,649 & 1,999 & 1,015 & 1,406 & 20.6 & 21.8 & 24.2 \\
\bottomrule
\end{tabular}
\end{table*}

\cref{tab:supp-ledger} traces all 32,688 source items. Each source enters with the labeled items of its public evaluation data, except that BLINK enters with the multi-image items of its validation split, EMMA with its multi-image items, MV-MATH with its multiple-choice items, VLM2-Bench with its image items and MileBench without its GPR1200 task, and that MaRVL~\cite{liu2021marvl} and DUDE~\cite{vanlandeghem2023dude} each enter with a random sample of 1,000 items, stratified by language and label for MaRVL and by answer type for the validation split of DUDE. The pool also leaves out 573 items whose images appear in public training data (249 of MMRB, 239 of MuirBench, 34 of MIRB, 27 of MMR-Life, 16 of Q-Bench2, 4 of MRAG-Bench and 2 each of VLM2-Bench and MileBench), 359 MMRB items that share image files with items of other sources, 9 MMLongBench-Doc items whose document is missing from our copy, 2 items whose images are missing from the release (1 each of MMRB and MMR-Life) and 4 VLM2-Bench rows that repeat the identifier of another row. The single-page items of DUDE enter with its sample and are removed by a rule.\looseness=-1

The first stage of the single-image filter removes 545 items before the audit.
Of the 32,143 audited items, the verdicts of the first audit pass remove 8,857 (one image suffices 6,434, text suffices 1,007, ambiguous or unanswerable 1,233, duplicate 170, removed by the adjudicator 13) and rules remove 1,333 (the caption of \cref{tab:supp-ledger} lists the rules).
The first pass corrects the label of 250 items and gives 350 items with a scoring error an answer format that accepts the requested form, set in the item or by a rule of the scorer.
Another 1,092 items have free-form answers, copy another source's kept items or are ambiguous.\looseness=-1

The free-form items of the Spot-the-Diff~\cite{jhamtani2018spotthediff}, IEdit~\cite{tan2019iedit} and ALFRED~\cite{shridhar2020alfred} tasks of MileBench~\cite{song2024milebench} are converted into multiple-choice items when the images support one verifiable answer.
An MLLM writes four options for each such item, a correct option that states one change, edit or next step of the reference answer that it verifies in the images and three distractors that the images show to be false.
An MLLM of another model family audits each converted item with the verdicts of \cref{sec:supp-audit}, and a converted item enters \bench only if it passes this independent audit.
Of the 404 free-form items among the 1,092, 91 pass the conversion, the independent audit and the first stage of the single-image filter, and 83 of them also pass the later checks and enter \bench.
The 313 that fail the first three steps and the 688 ambiguous items and copies, 1,001 items in all, are removed.\looseness=-1
\begin{table*}[t]
\centering\footnotesize
\setlength{\tabcolsep}{4pt}
\caption{\textbf{Task categories.} Operation each category covers, an example question from \bench (source, number of images $K$) and the number of \bench items, including the rewritten items of each category.}
\label{tab:supp-groups}
\begin{tabular}{@{}l >{\raggedright\arraybackslash}p{0.27\linewidth} >{\raggedright\arraybackslash}p{0.47\linewidth} r@{}}
\toprule
Category & Operation & Example & Items \\
\midrule
Aggregate & count or combine values over images & ``How many total bottles are there in the images?'' (MuirBench~\cite{wang2025muirbench}, $K{=}2$) & 1,055 \\
Compare/change & judge a property across images, or find what differs between aligned views & ``Compared to the first image, how is the sharpness of the second image?'' (Q-Bench2~\cite{zhang2024qbenchplus}, $K{=}2$) & 2,262 \\
Reasoning/verification & combine quantities from figures, charts or diagrams, or verify a statement about a set of images & ``The following images show two graphs. \dots\ Are these graphs isomorphic?'' (ReMI~\cite{kazemi2024remi}, $K{=}2$) & 4,793 \\
Spatial correspondence & relate camera views in 3D, or match objects, regions or styles across images & ``The images were taken in succession by a vehicle-mounted camera. In which direction is the vehicle moving?'' (MMSI-Bench~\cite{yang2026mmsibench}, $K{=}2$) & 4,618 \\
Temporal/action & order frames or steps, track state over time, choose the next action & ``Please arrange these images in the correct order of events.'' (MMR-Life~\cite{li2026mmrlife}, $K{=}4$) & 3,463 \\
Text/web & answer from text and images together, with the images as evidence & ``Do the Black Lion and the Hop Poles in Hammersmith both have a black background on the main name sign above the ground floor?'' (MileBench~\cite{song2024milebench} WebQA~\cite{chang2022webqa}, $K{=}2$) & 154 \\
Retrieval & find one image among many, or read a multi-page document & ``Please find the most relevant picture among the candidate images for this description.'' (MMRB~\cite{cheng2025mmrb}, $K{=}4$) & 2,351 \\
\bottomrule
\end{tabular}
\end{table*}

Claude Fable~5 rewrites items, and 1,546 rewrites pass the GPT-5.6-Sol check and the first stage of the single-image filter.
After human review and an automatic check, we remove the 95 of the 1,546 rewrites that fail either (\cref{sec:human}).

A second, targeted audit pass applies the verdicts and procedure of \cref{sec:supp-audit} to the kinds of items that human reviewers flagged. It removes 2,008 items (1,999 source items and 9 rewrites), among them 805 MMRB items that copy another source's kept items, and corrects the label of 86 items and the question or options of 321 of the source items that \bench keeps.
The second stage of the single-image filter, which runs once labels, formats and option orders are final, removes 1,038 items that one image answers (1,015 source items, 1-img in \cref{tab:supp-ledger}, and 23 rewrites).
Last, so that no source item enters together with its rewrite, we remove the 648 source items whose rewrites enter \bench (part of Rm.\ free in \cref{tab:supp-ledger}) and drop the 13 rewrites whose source items belong to \benchlite. \bench then holds 18,696 items in seven task categories, including 1,406 rewrites and the 2,351 items of the retrieval category.
Over all steps, the two stages of the single-image filter remove 1,560 source items, the two audit passes 10,856, rules 1,333, the handling of free-form items, copies and ambiguous items 1,001 and the replacement of source items by their rewrites 648, so 17,290 of the 32,688 source items enter \bench.
After the later removal steps, 222 source items of \bench carry a label that we correct, and 569 in all, including the 83 converted free-form items, differ beyond option order in their label, answer format, question or options (Fix in \cref{tab:stats}).\looseness=-1

The shares in \cref{sec:audit} and \cref{fig:audit} are computed over these 32,688 items and count the verdicts of both audit passes and both filter stages. Retrieval and multi-page document items that one image or the question answers and copies of other sources' items do not count, even when they are removed. The last three columns of \cref{tab:supp-ledger} give the shares that one image or the question settles per source.

We balance the position of the correct option. Within each option count, every letter is the answer about equally often, which we obtain by exchanging the correct option with another option in 834 items.
Items whose options refer to other options, whose answer reads like ``none of the above'', or whose options are bare letters keep their order.

\subsection{Human verification}\label{sec:supp-human}
Four reviewers take part. A reviewer checks every corrected label and every rewritten item, and two reviewers check each item of the 1,000-item agreement sample independently.
Reviewers work in a web interface that shows the images, the question and the options.
A reviewer first answers the item without seeing the label, then sees the label and records a verdict and an optional note.
On the rewritten items, reviewers answer 94\% correctly before seeing the label and judge 98\% to need more than one image, and the items they flag are removed from \bench (\cref{sec:supp-ledger}).
Reviewers agree with the audit verdict on 82.6\% of their answers over the 1,000-item sample (Wilson~\cite{wilson1927probable} 95\% interval [80.9, 84.2]) and with each other on 82.3\% of the items.

\subsection{Task categories}\label{sec:supp-groups}
Each item has one of seven task categories, named for the cross-image operation that its answer requires (\cref{tab:supp-groups}).
GPT-5.6-Sol labeled the operation of the items of the 19 sources, and a fixed crosswalk maps its operation types to the categories.
Five structural rules settle the types that the crosswalk splits (galleries searched for one image, a needle found and then read, tracking over three or more frames, a table in the prompt, and science and coding problems).
In some source subsets, a rule on the subset, its source category or the question pattern sets the category instead, as for the WebQA source items of MileBench, which are text/web unless the second audit pass corrects their category.
Rewritten items keep the crosswalk category of the item they were rewritten from.\looseness=-1

Items whose task is finding one image among many or reading a multi-page document form the retrieval category. The rules above, not the audit verdict, decide which items are retrieval items. A retrieval item asks the model to find the image or page that holds the answer among the others, so the statistics of \cref{sec:audit} do not count such items as answered by one image or the question, even when an audit verdict removes them (hatched in \cref{fig:audit}). Retrieval thus tests finding evidence among the images, and the other six categories test combining it across them. The MIRB item of \cref{fig:supp-ex-single} asks where its two landmark photos were taken, which the rules do not count as retrieval. Either photo answers it, so the shares of \cref{sec:audit} count it. The MMLongBench-Doc item of \cref{fig:supp-cat-ret}, in contrast, asks about the pages of one document, which the rules count as retrieval.
For every item except the 83 converted free-form items, the same labeling records the cross-image operation, the relation between the images, the number of operations and reasoning hops, and the time scale. \cref{tab:supp-axes,fig:supp-labels} use these labels. We release these per-item labels with \bench.\looseness=-1

\newpage
\section{Example items}\label{sec:supp-examples}
\subsection{Removed and corrected items}\label{sec:supp-ex-removed}
\begin{figure*}[t]
  \centering
  \includegraphics[width=\linewidth]{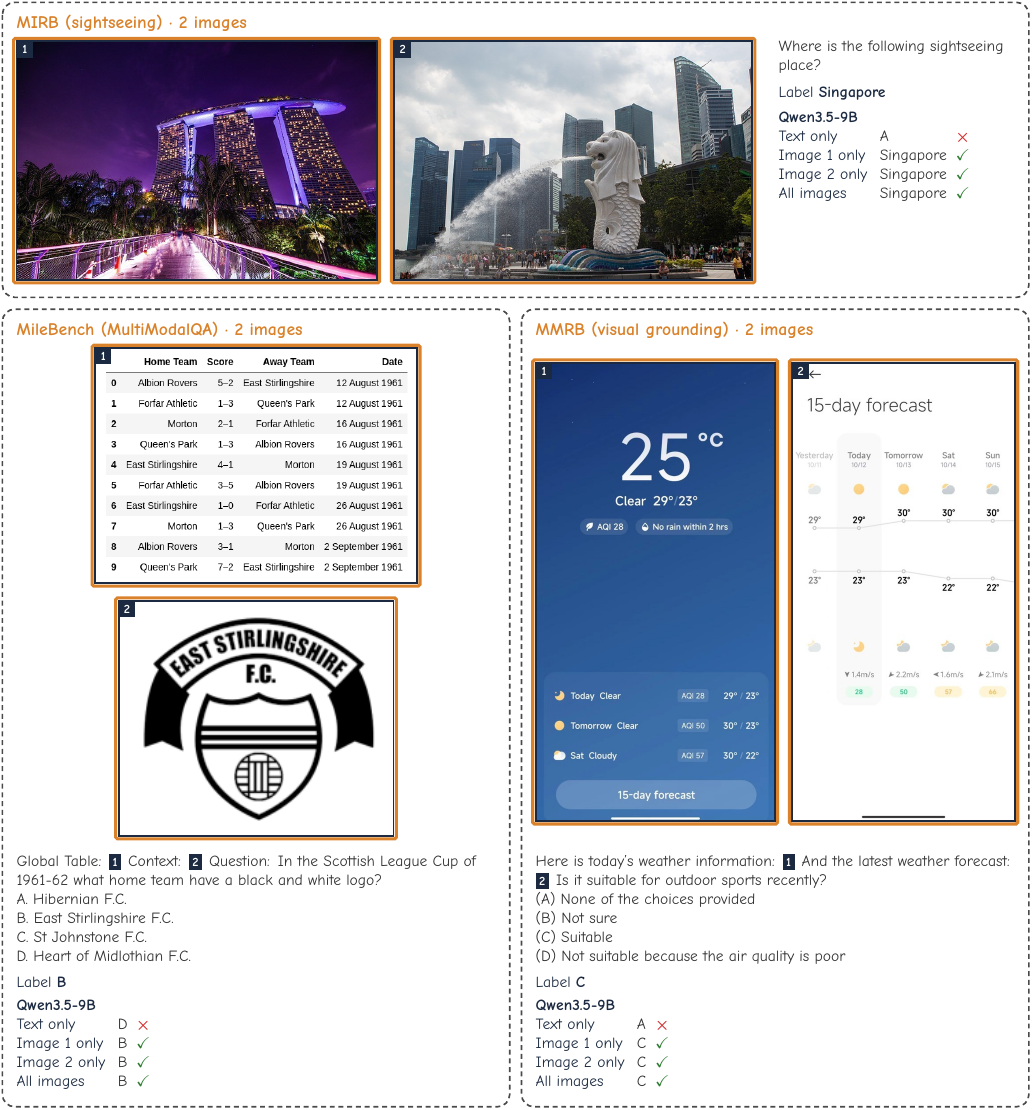}
  \caption{\textbf{Items removed because one image answers them.} Each panel shows a source item with its images, question, options if any and label. An orange frame marks each image that answers the item alone, and the symbols $\checkmark$ and $\times$ show whether Qwen3.5-9B answers correctly or incorrectly with the text only, with one image kept and with all images, under the gray masking of \cref{sec:shortcut}. Answer-format instructions and generic preambles of the prompts are not shown here or in \crefrange{fig:supp-ex-text}{fig:supp-ex-rewrite}. Either landmark of the MIRB item, Marina Bay Sands in Image~1 or the Merlion in Image~2, gives Singapore, and with the text only Qwen3.5-9B answers this open question with the letter A. In the MileBench MultiModalQA item, the crest in Image~2 is black and white and names East Stirlingshire F.C., and among the clubs of the options only East Stirlingshire appears as a home team in the table of Image~1. Image~1 of the MMRB item shows clear weather with an air-quality index of 28, and Image~2 shows a forecast of sunny or partly cloudy days, so either screen gives Suitable.}
  \label{fig:supp-ex-single}
\end{figure*}

\begin{figure*}[t]
  \centering
  \includegraphics[width=\linewidth]{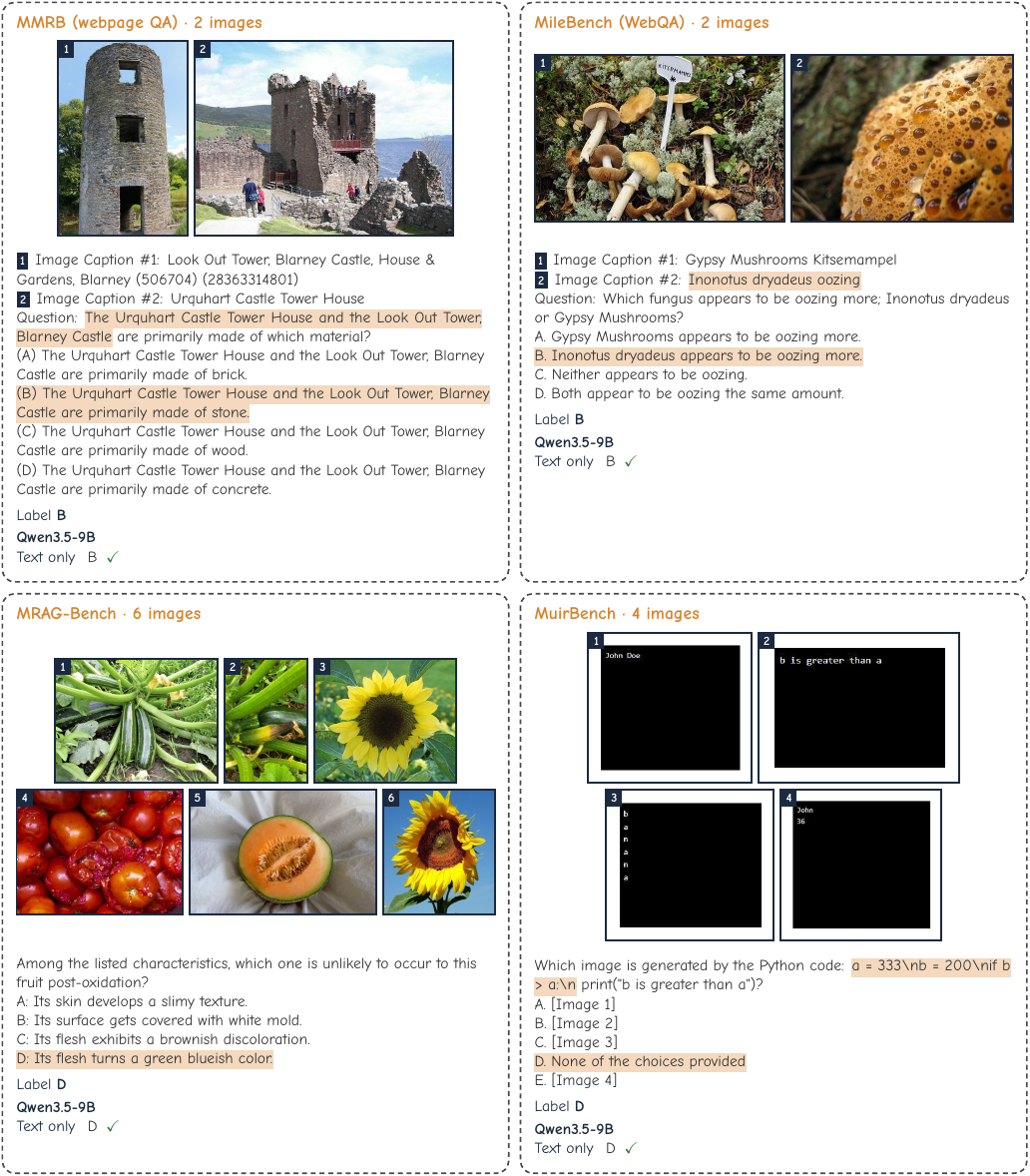}
  \caption{\textbf{Items removed because the text answers them.} Each panel shows a source item with its images, question, options and label, with any part of the text that gives the answer away and the correct option highlighted, and the answer of Qwen3.5-9B with all images replaced by gray (text only). In the MMRB item, the names of the two castle towers point to stone without the images. In the MileBench item, the caption of Image~2 says that Inonotus dryadeus is oozing. In the MRAG-Bench item, whose first image is the query and whose other images are retrieved examples, three options describe common spoilage, which leaves option D. In the MuirBench item, the code prints nothing because 200 is not greater than 333, which makes option D the likely answer without the images.}
  \label{fig:supp-ex-text}
\end{figure*}

\begin{figure*}[t]
  \centering
  \includegraphics[width=\linewidth]{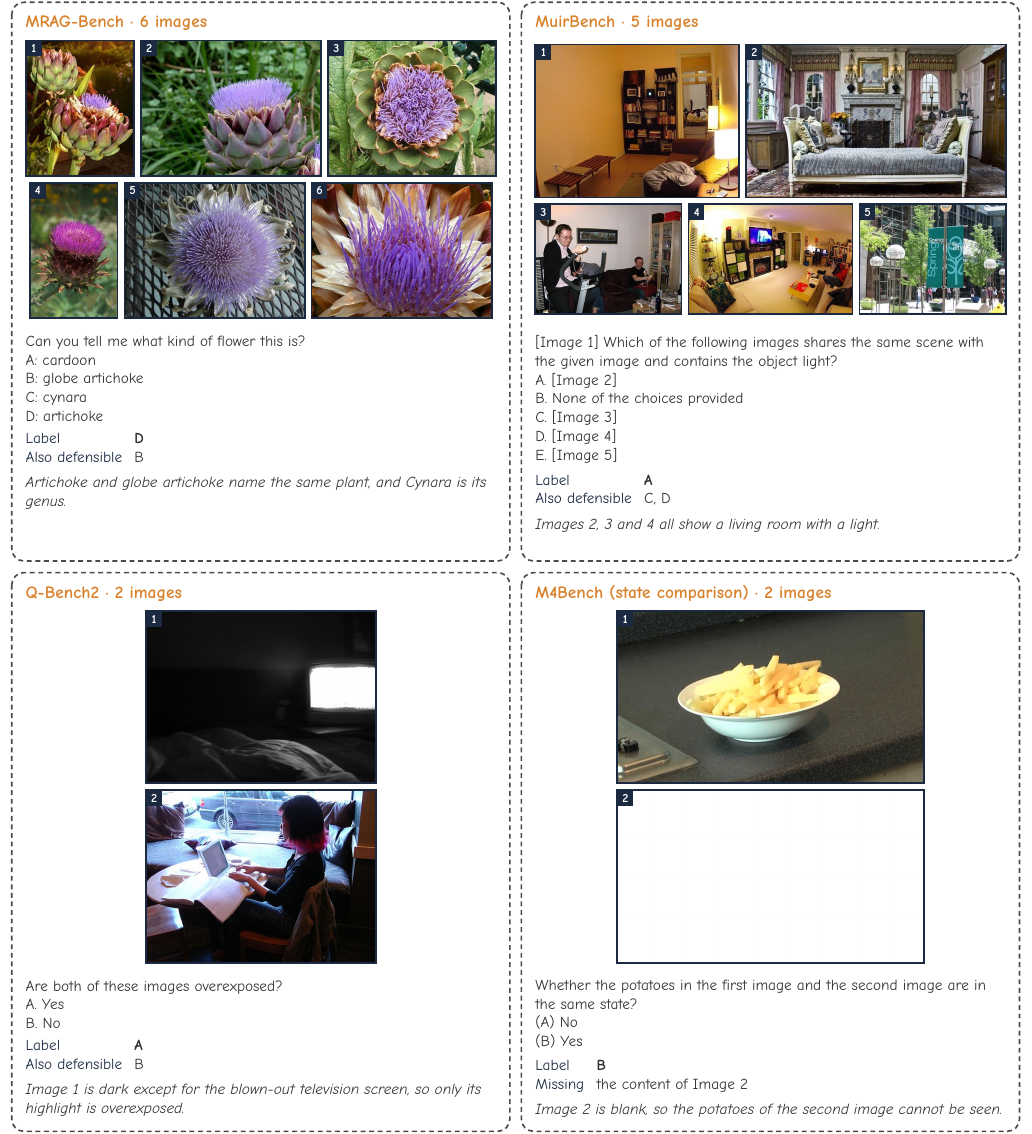}
  \caption{\textbf{Ambiguous or unanswerable items.} The audit removes source items for which two or more answers are defensible or no answer follows from the inputs (\cref{sec:supp-audit}). Each panel shows a source item with its images, question, options and label, the other defensible answers or the missing information, and the image evidence. In the MRAG-Bench item, whose first image is the query and whose other images are retrieved examples, artichoke and globe artichoke name the same plant. In the MuirBench item, Images~2, 3 and~4 all show a living room with a light, like the room in Image~1. In the Q-Bench2 item, Image~1 is dark apart from the blown-out television screen, so it counts as overexposed under one reading of the question and not under another. Image~2 of the M4Bench item is blank, so the potatoes it should show cannot be compared.}
  \label{fig:supp-ex-ambig}
\end{figure*}

\begin{figure*}[t]
  \centering
  \includegraphics[width=\linewidth]{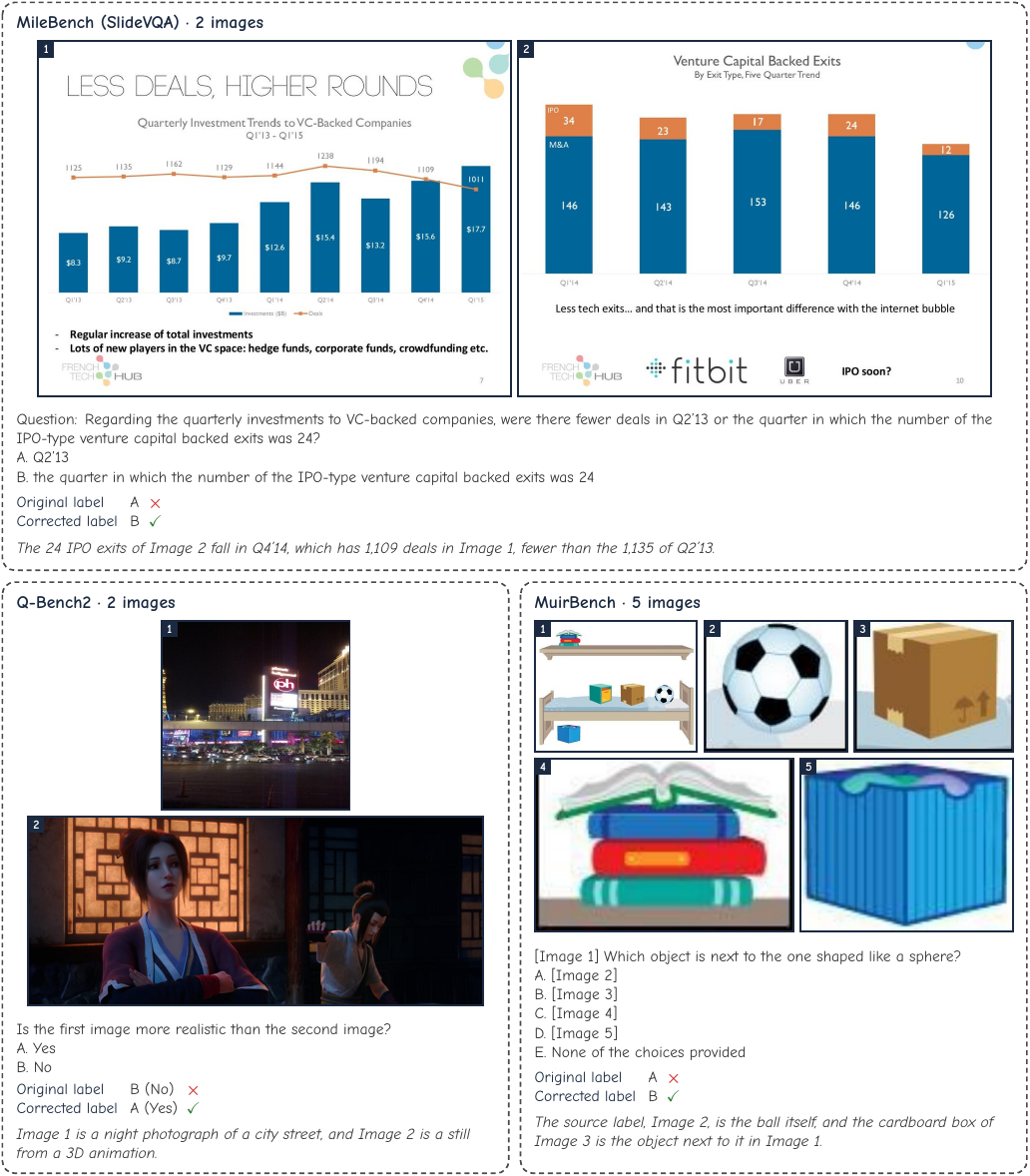}
  \caption{\textbf{Items kept with a corrected label.} Each panel shows an \bench item with its images, question and options, the label of the source benchmark ($\times$), the corrected label ($\checkmark$) and the image evidence for the correction. In the MileBench item, the source label is Q2'13, but the 24 IPO exits of Image~2 fall in Q4'14, which has 1,109 deals in Image~1, fewer than the 1,135 of Q2'13. In the Q-Bench2 item, Image~1 is a night photograph and Image~2 a frame of a 3D animation, so the first image is the more realistic one. In the MuirBench item, the source label is the ball itself, and the object next to the ball in Image~1 is the cardboard box of Image~3. A human reviewer accepted each correction (\cref{sec:supp-human}).}
  \label{fig:supp-ex-label}
\end{figure*}

\begin{figure*}[t]
  \centering
  \includegraphics[width=\linewidth]{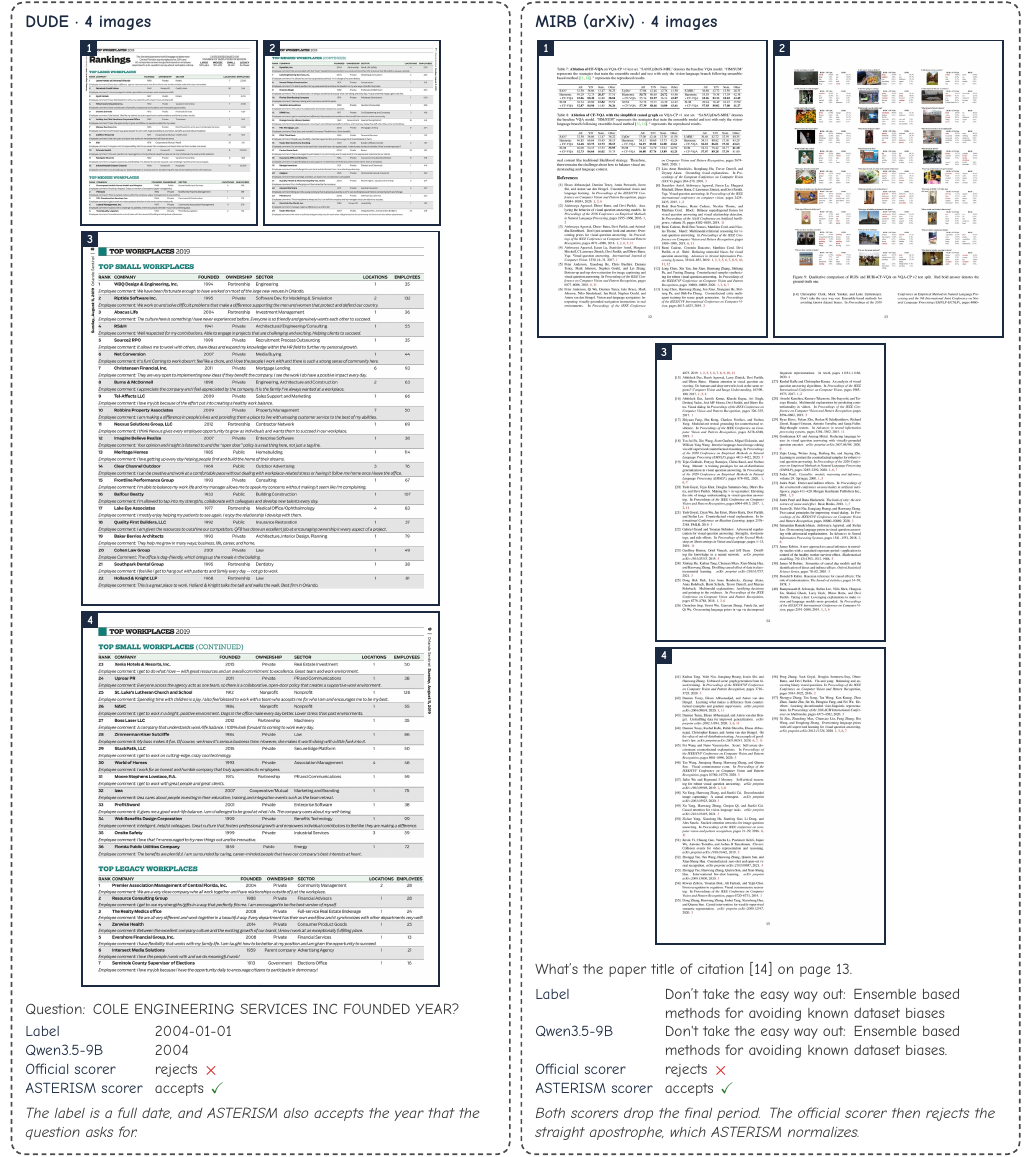}
  \caption{\textbf{Items kept with a corrected answer format.} The label of each item is right, but the official scorer of its source rejects a correct answer given in the form the question asks for, and \bench scores the item with an answer format that accepts this form, set in the item or by a rule of the scorer (\cref{sec:supp-audit,sec:supp-ledger}). Each panel shows an \bench item with its images, question and label, an answer of Qwen3.5-9B, and whether the official scorer and the \bench scorer accept that answer. The DUDE question asks for the founding year, the label is the full date, and \bench also accepts the year. The answer to the MIRB item gives the cited title with a straight apostrophe where the label has a curly one, which the exact match of the official MIRB scorer rejects and \bench normalizes.}
  \label{fig:supp-ex-format}
\end{figure*}

The figures of this section cover items that one image answers (\cref{fig:supp-ex-single}), items that the text answers (\cref{fig:supp-ex-text}), ambiguous or unanswerable items (\cref{fig:supp-ex-ambig}), items kept with a corrected label (\cref{fig:supp-ex-label}) and items kept with a corrected answer format (\cref{fig:supp-ex-format}). They show the items of each kind with the images, the question, any options and the label of the source benchmark.
In each of \cref{fig:supp-ex-single,fig:supp-ex-text,fig:supp-ex-ambig,fig:supp-ex-label,fig:supp-ex-format}, the items come from different source benchmarks or subsets.

\subsection{Rewritten items}\label{sec:supp-ex-rewrite}
\begin{figure*}[t]
  \centering
  \includegraphics[width=\linewidth]{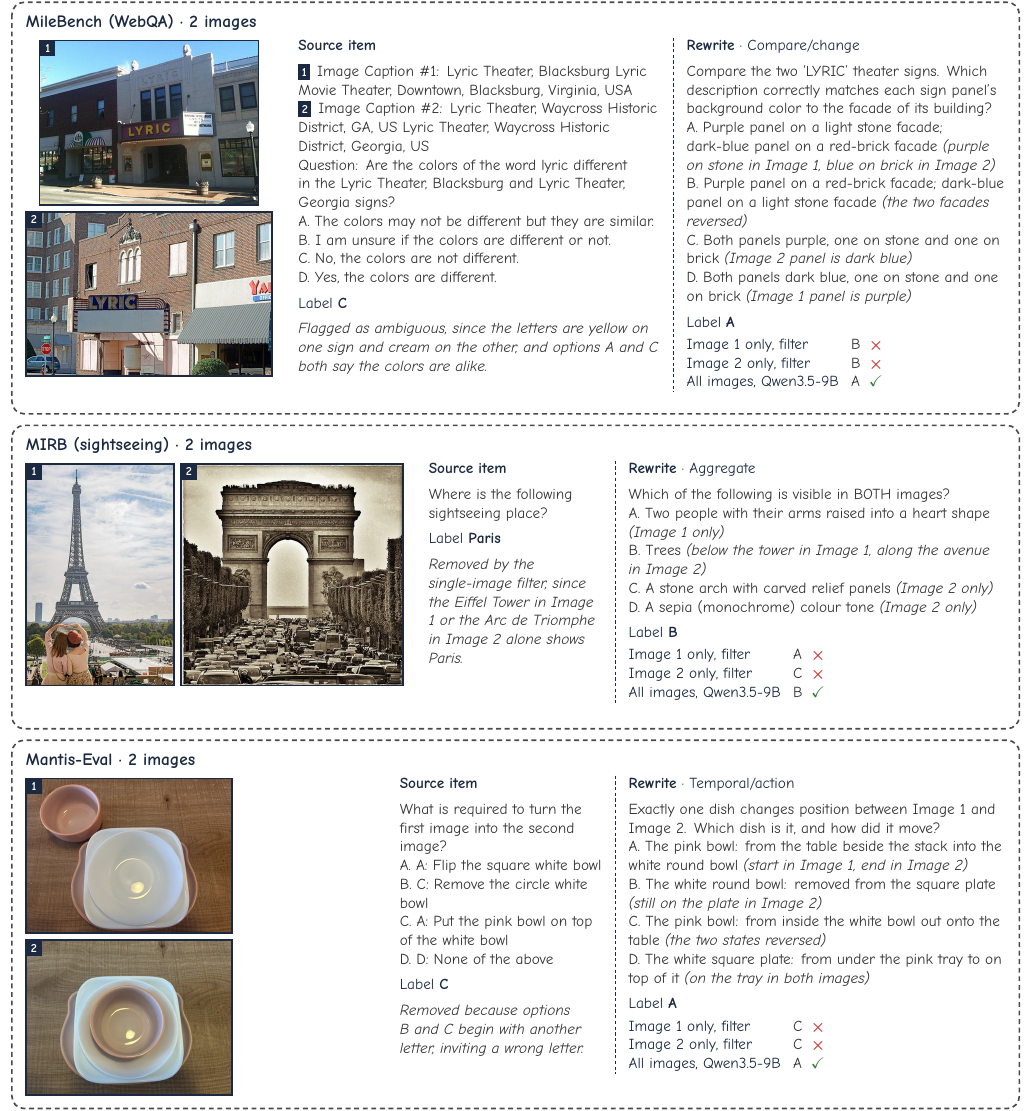}
  \caption{\textbf{Rewritten items next to their source items.} Each panel shows a source item, why it was flagged or removed, and its rewrite in \bench with the task category of the rewrite. Rows marked filter give the most common answer of the filter committee from that image alone. Italic notes after the options give the image evidence behind each option. The MileBench source is ambiguous, since LYRIC is yellow in Image~1 and cream in Image~2 and options A and C both say the colors are alike. The rewrite matches each sign panel to its facade, purple on light stone in Image~1 and dark blue on red brick in Image~2. The Eiffel Tower in Image~1 and the Arc de Triomphe in Image~2 each answer the MIRB source alone. Of the four options in the rewrite, only the trees appear in both images, while the heart gesture shows only in Image~1 and the carved arch and the sepia tone only in Image~2. Options B and C of the Mantis-Eval source begin with the letter of another option, and the rewrite asks which dish moved and in which direction, so it needs the start in Image~1 and the end in Image~2.}
  \label{fig:supp-ex-rewrite}
\end{figure*}

A rewrite keeps the images of its source item and replaces the question and options so that each option pairs evidence from different images (\cref{sec:curation}).
\cref{fig:supp-ex-rewrite} shows rewrites next to their source items. The italic notes on the image evidence behind each option are ours and are not part of the items.

\clearpage

\subsection{Items per task category}\label{sec:supp-ex-cat}
\begin{figure*}[t]
  \centering
  \includegraphics[width=\linewidth]{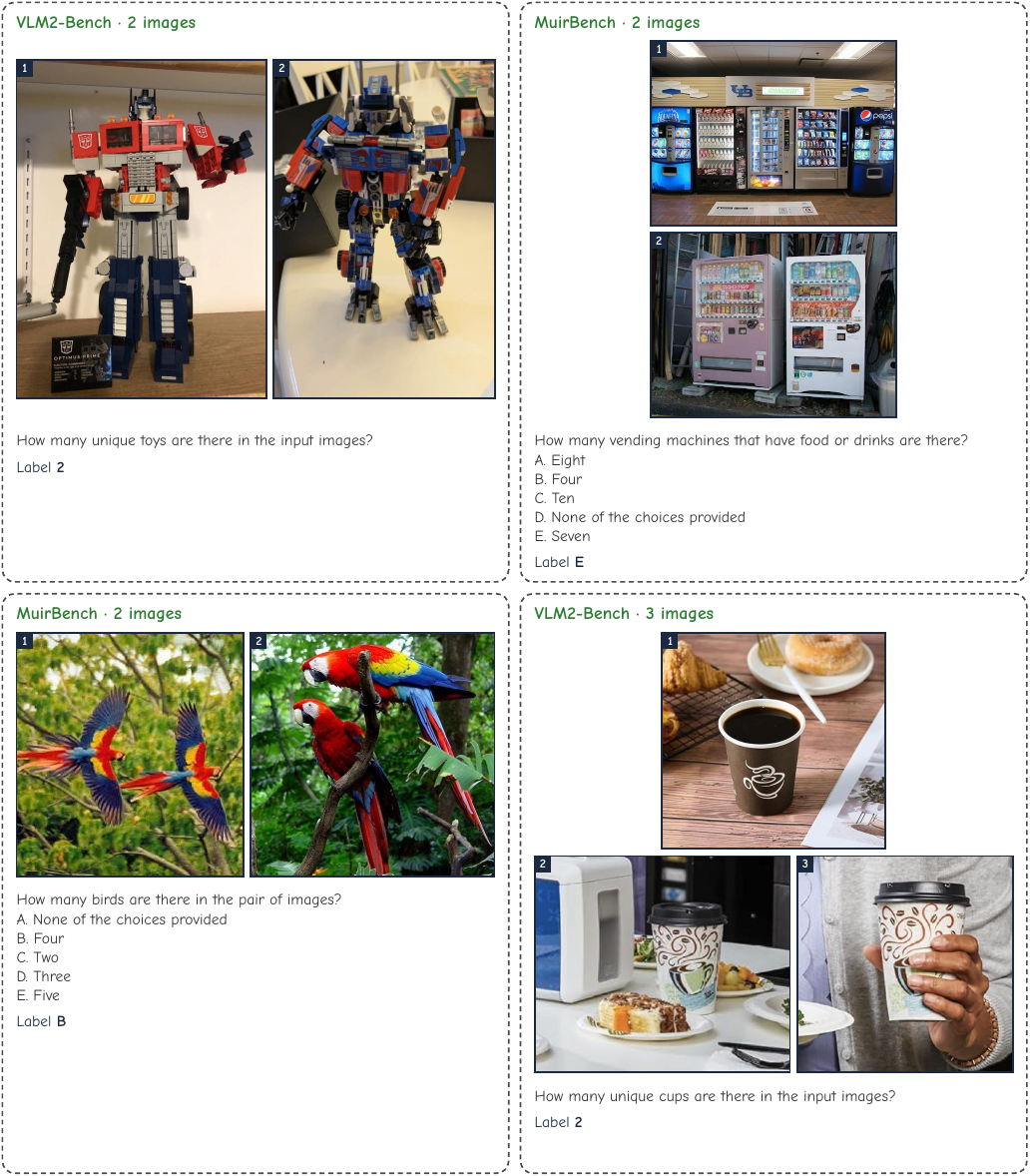}
  \caption{\textbf{Aggregate items.} Four \bench items of the category, each with its source, number of images, question, options if any and label. In the first VLM2-Bench item, Images~1 and~2 show two different Optimus Prime models. In the first MuirBench item, Image~1 shows five vending machines and Image~2 two. In the second MuirBench item, each image shows two scarlet macaws, four in all. In the second VLM2-Bench item, Images~2 and~3 show the same swirl-print paper cup and Image~1 a brown one, so the images hold two different cups. Answer-format instructions and generic preambles of the prompts are not shown here or in \crefrange{fig:supp-cat-cmp}{fig:supp-cat-ret}.}
  \label{fig:supp-cat-agg}
\end{figure*}

\begin{figure*}[t]
  \centering
  \includegraphics[width=\linewidth]{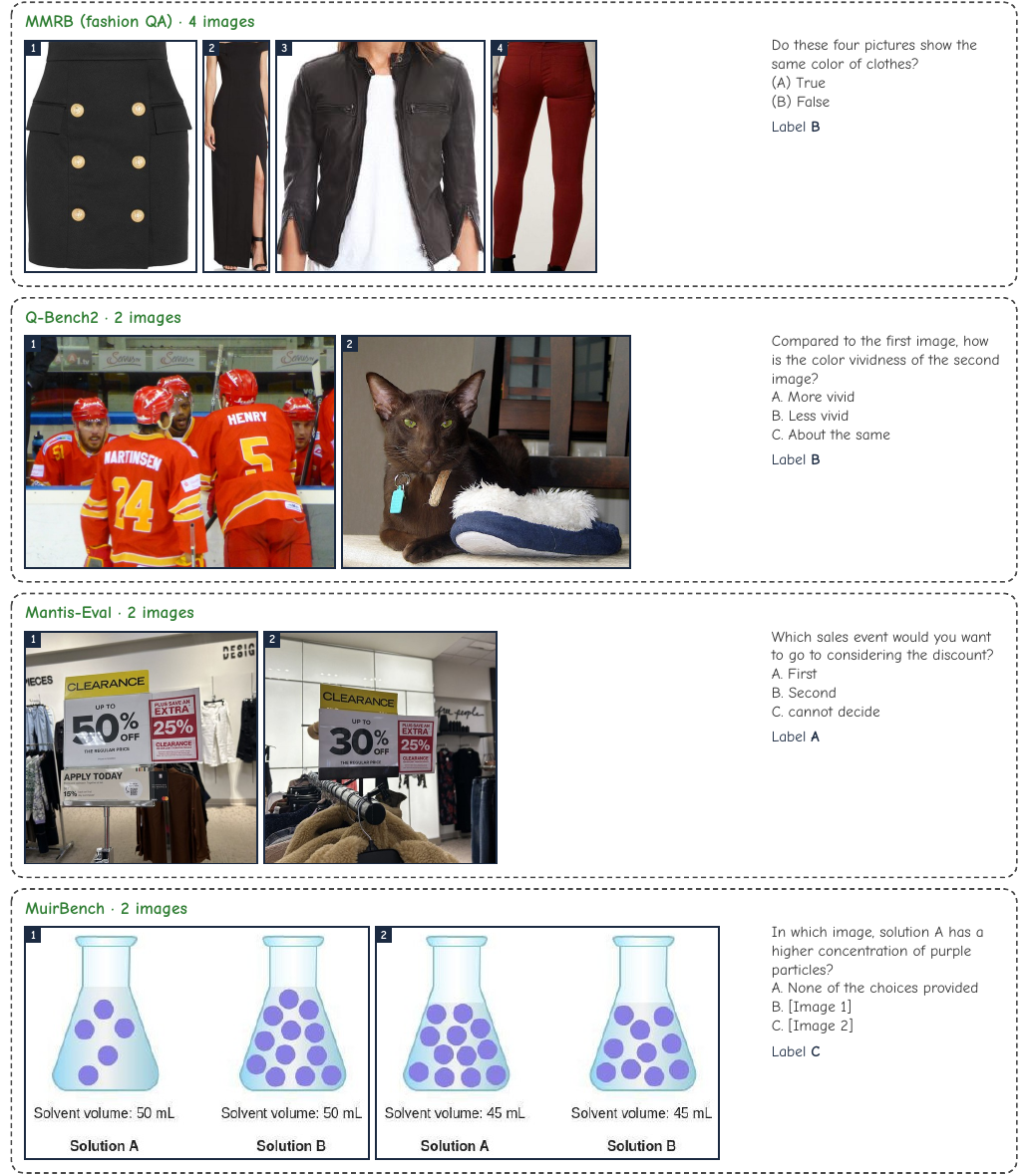}
  \caption{\textbf{Compare/change items.} Four \bench items of the category, shown as in \cref{fig:supp-cat-agg}. In the MMRB item, Images~1 to~3 show black garments and Image~4 dark red jeans, so the clothes do not share one color. In the Q-Bench2 item, the red and yellow jerseys of Image~1 are more vivid than the brown cat of Image~2. In the Mantis-Eval item, the first sale offers up to 50\% off and the second up to 30\% off, each with an extra 25\% off. In the MuirBench item, solution~A has more particles in less solvent in Image~2 than in Image~1.}
  \label{fig:supp-cat-cmp}
\end{figure*}

\begin{figure*}[t]
  \centering
  \includegraphics[width=\linewidth]{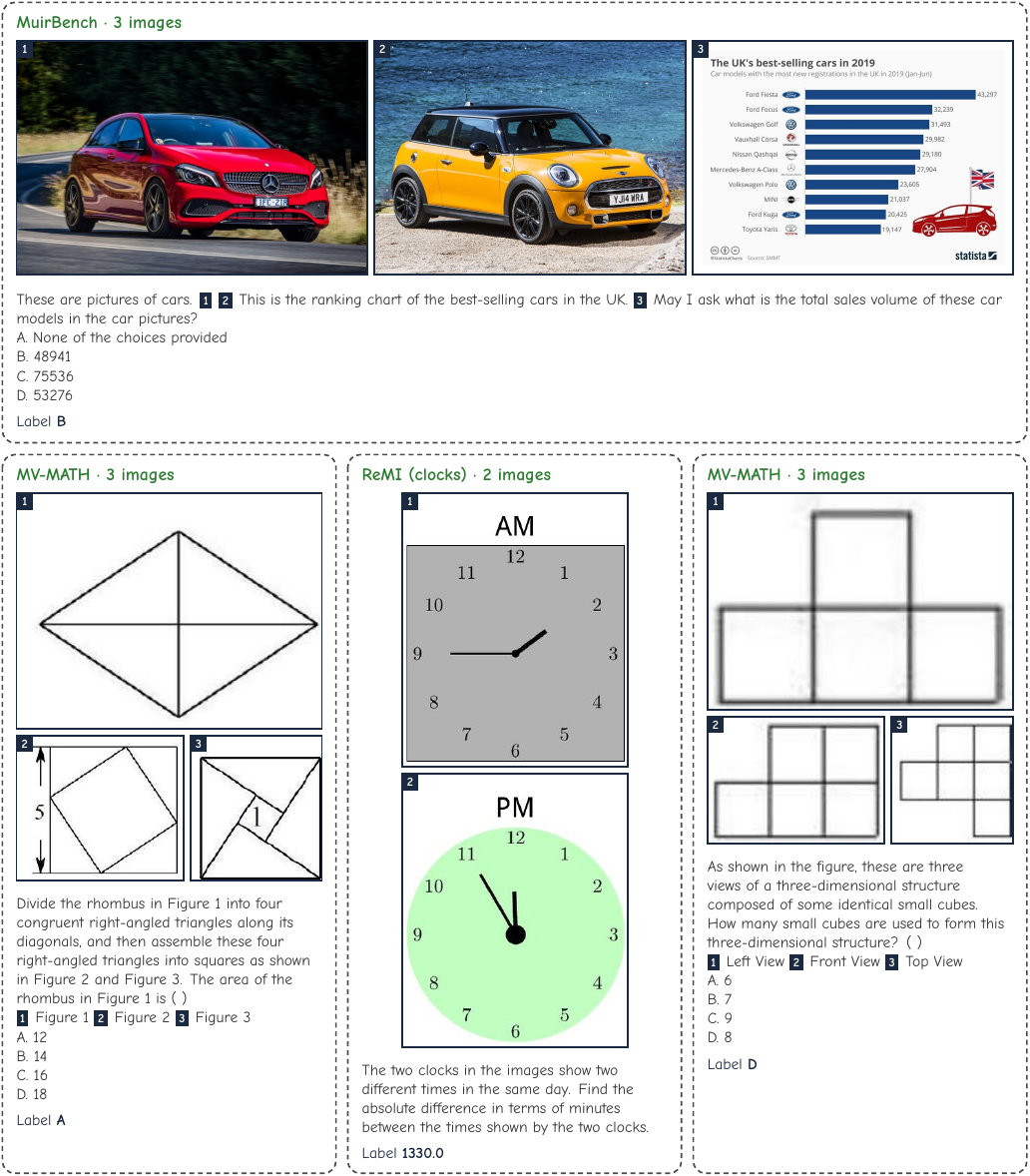}
  \caption{\textbf{Reasoning/verification items.} Four \bench items of the category, shown as in \cref{fig:supp-cat-agg}. In the MuirBench item, Image~1 shows a Mercedes-Benz A-Class and Image~2 a MINI, and the sales that the chart in Image~3 lists for them, 27,904 and 21,037, add up to 48,941. In the first MV-MATH item, the square of side 5 in Image~2 gives the sum of the two legs of each triangle and the unit square in Image~3 their difference, so the legs are 3 and~2 and the rhombus has area 12. In the ReMI item, the clocks show 1:45 in the morning and 11:55 at night. In the second MV-MATH item, the left, front and top views in the three images together fix the number of cubes.}
  \label{fig:supp-cat-rsn}
\end{figure*}

\begin{figure*}[t]
  \centering
  \includegraphics[width=\linewidth]{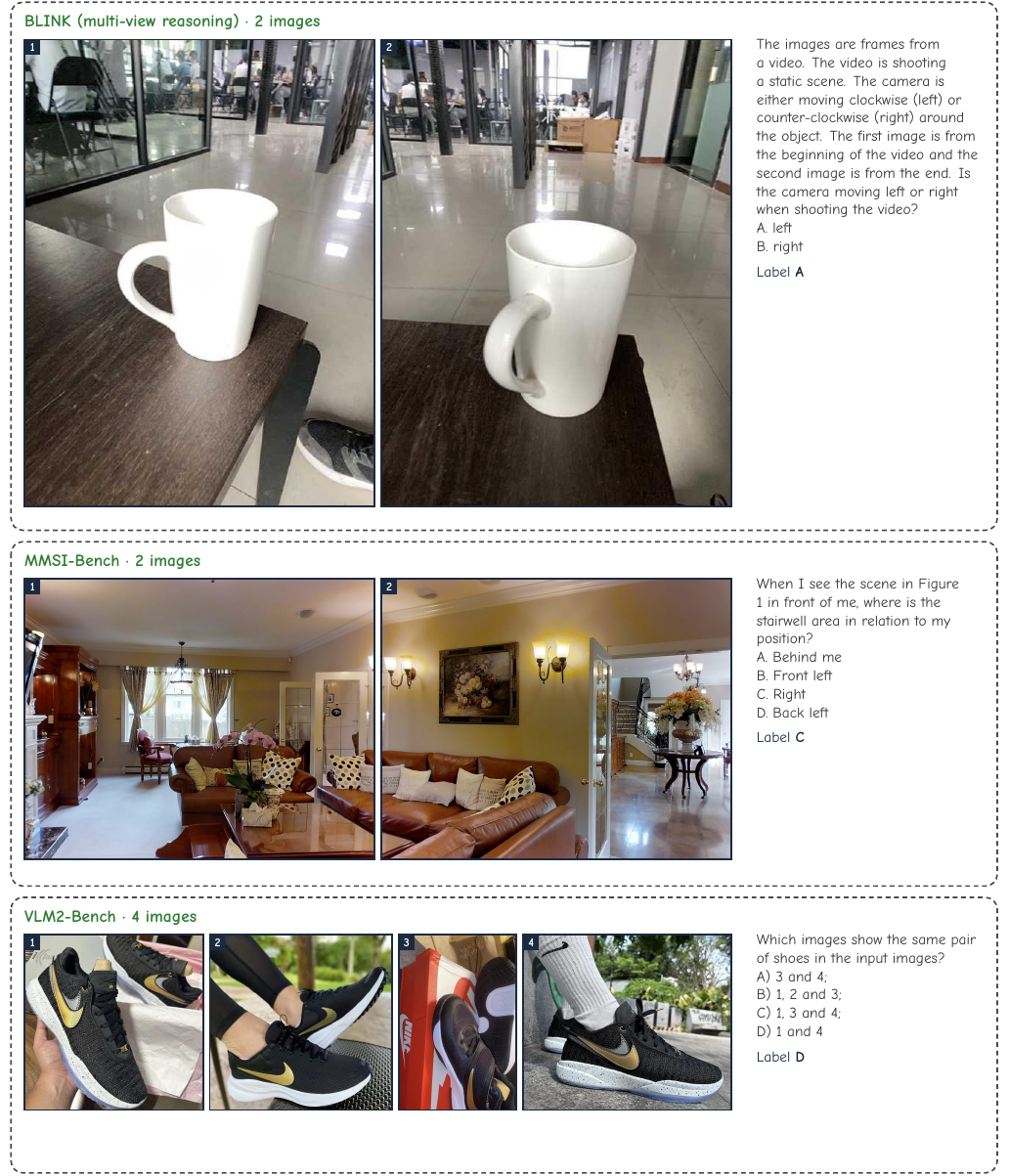}
  \caption{\textbf{Spatial correspondence items.} Three \bench items of the category, shown as in \cref{fig:supp-cat-agg}. In the BLINK item, the direction of the camera shows only in how the view changes between the two frames. In the MMSI-Bench item, the leather sofa with dotted cushions at the right of Image~1 reappears in Image~2, where the open door beside it shows the hall with the staircase, so the stairwell lies to the right of the first view. In the VLM2-Bench item, only Images~1 and~4 show the knit shoe with the speckled white sole.}
  \label{fig:supp-cat-spa}
\end{figure*}

\begin{figure*}[t]
  \centering
  \includegraphics[width=\linewidth]{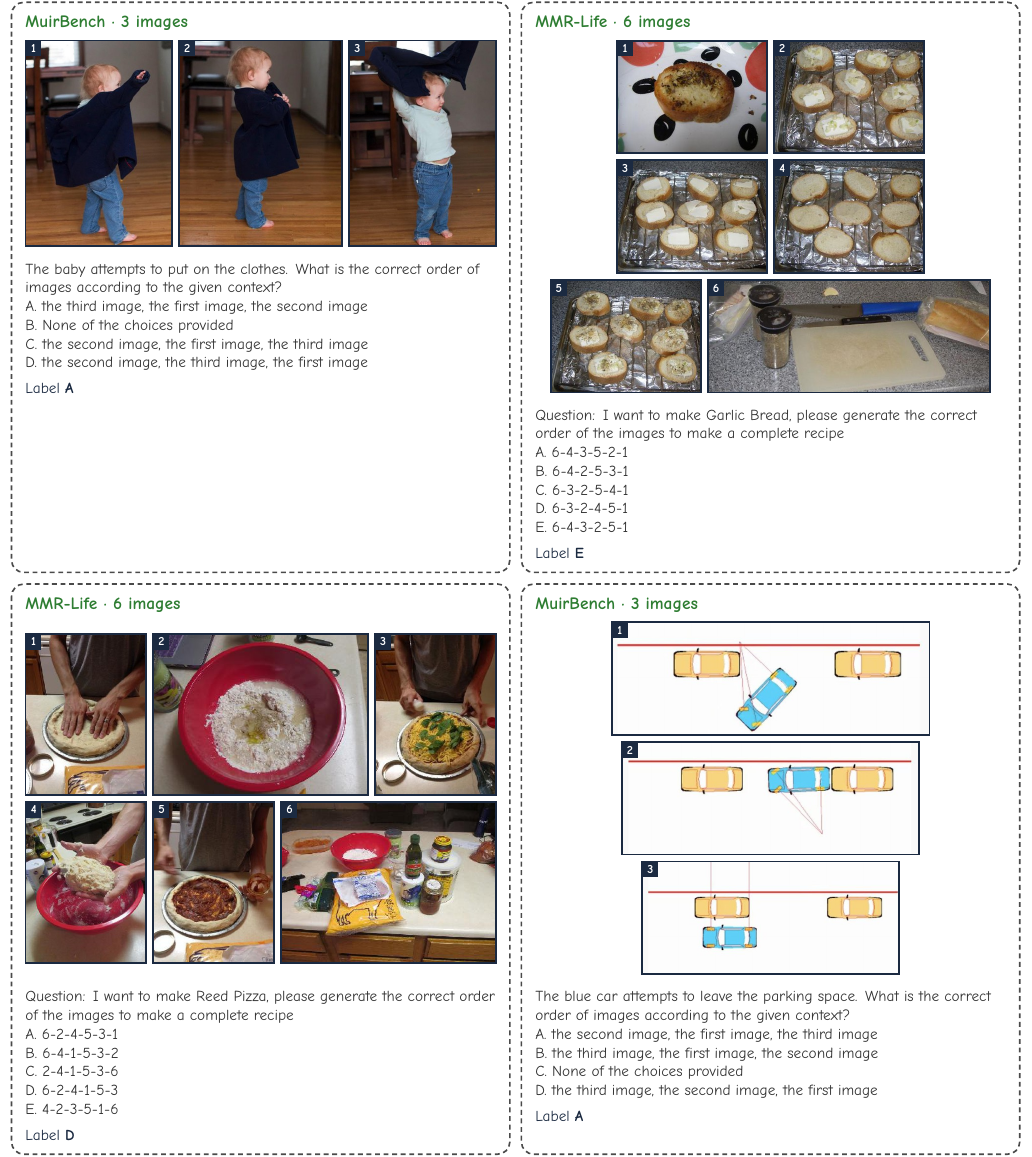}
  \caption{\textbf{Temporal/action items.} Four \bench items of the category, shown as in \cref{fig:supp-cat-agg}. In the first MuirBench item, the coat is lifted over the head in Image~3, half on in Image~1 and on in Image~2. In the first MMR-Life item, the ingredients of Image~6 come first, then the bare slices of Image~4, the butter of Image~3, the garlic of Image~2, the herbs of Image~5 and the toasted slice of Image~1. In the second MMR-Life item, the pizza runs from the ingredients in Image~6 through the flour, kneading and pressed crust of Images~2, 4 and~1 to the sauce in Image~5 and the toppings in Image~3. In the second MuirBench item, the blue car is parked in Image~2, angles out of the space in Image~1 and has left it in Image~3.}
  \label{fig:supp-cat-tmp}
\end{figure*}

\begin{figure*}[t]
  \centering
  \includegraphics[width=\linewidth]{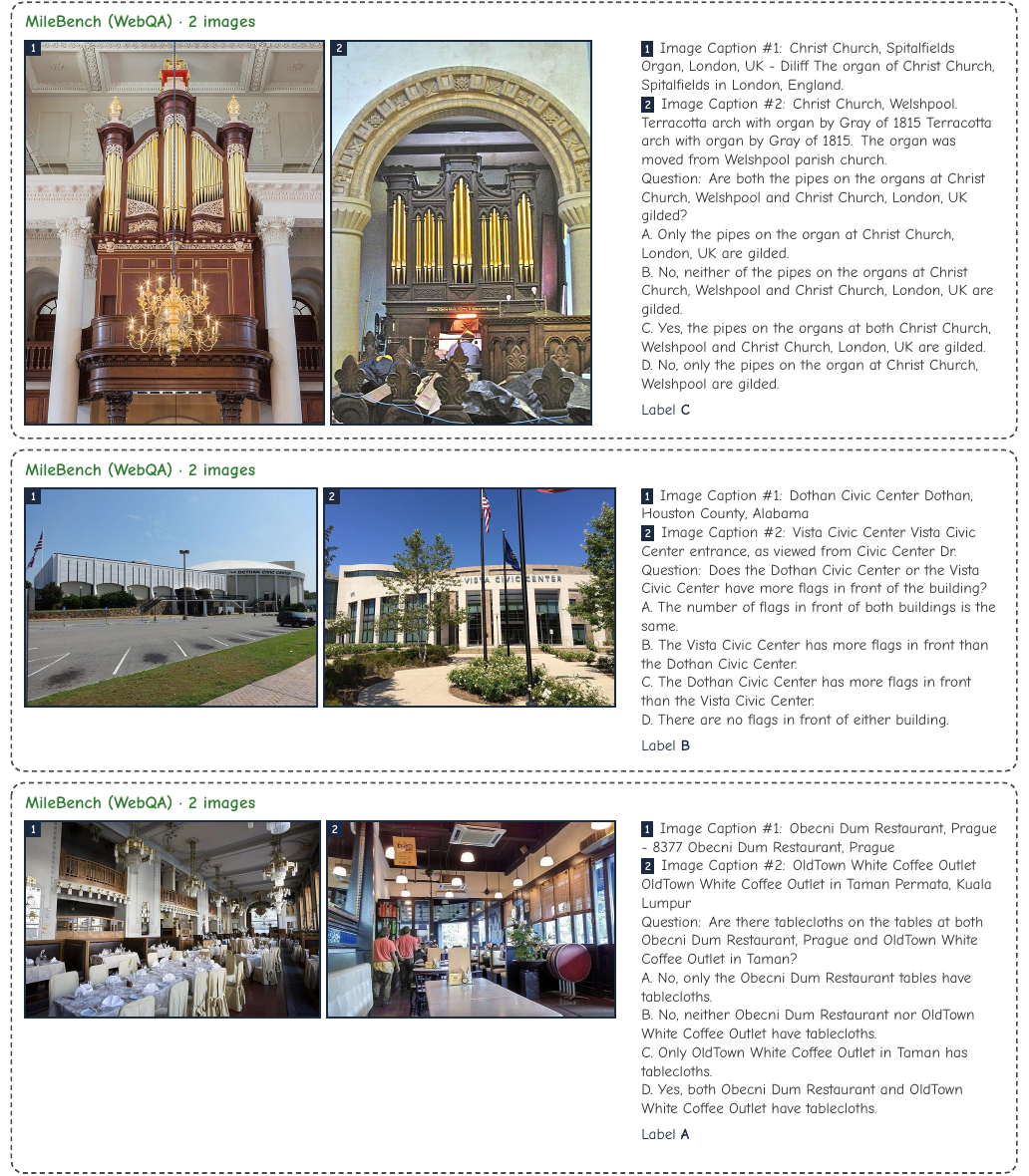}
  \caption{\textbf{Text/web items.} Three \bench items of the category, shown as in \cref{fig:supp-cat-agg}. In the first MileBench item, the captions name the two organs, and only the images show that both have gilded pipes, in London in Image~1 and in Welshpool in Image~2. In the second MileBench item, the captions name the two civic centers, and only the images show one flag in Image~1 and three flags in Image~2. In the third MileBench item, the captions name the two restaurants, and only the images show tablecloths in Image~1 and bare tables in Image~2.}
  \label{fig:supp-cat-txt}
\end{figure*}

\begin{figure*}[t]
  \centering
  \includegraphics[width=\linewidth]{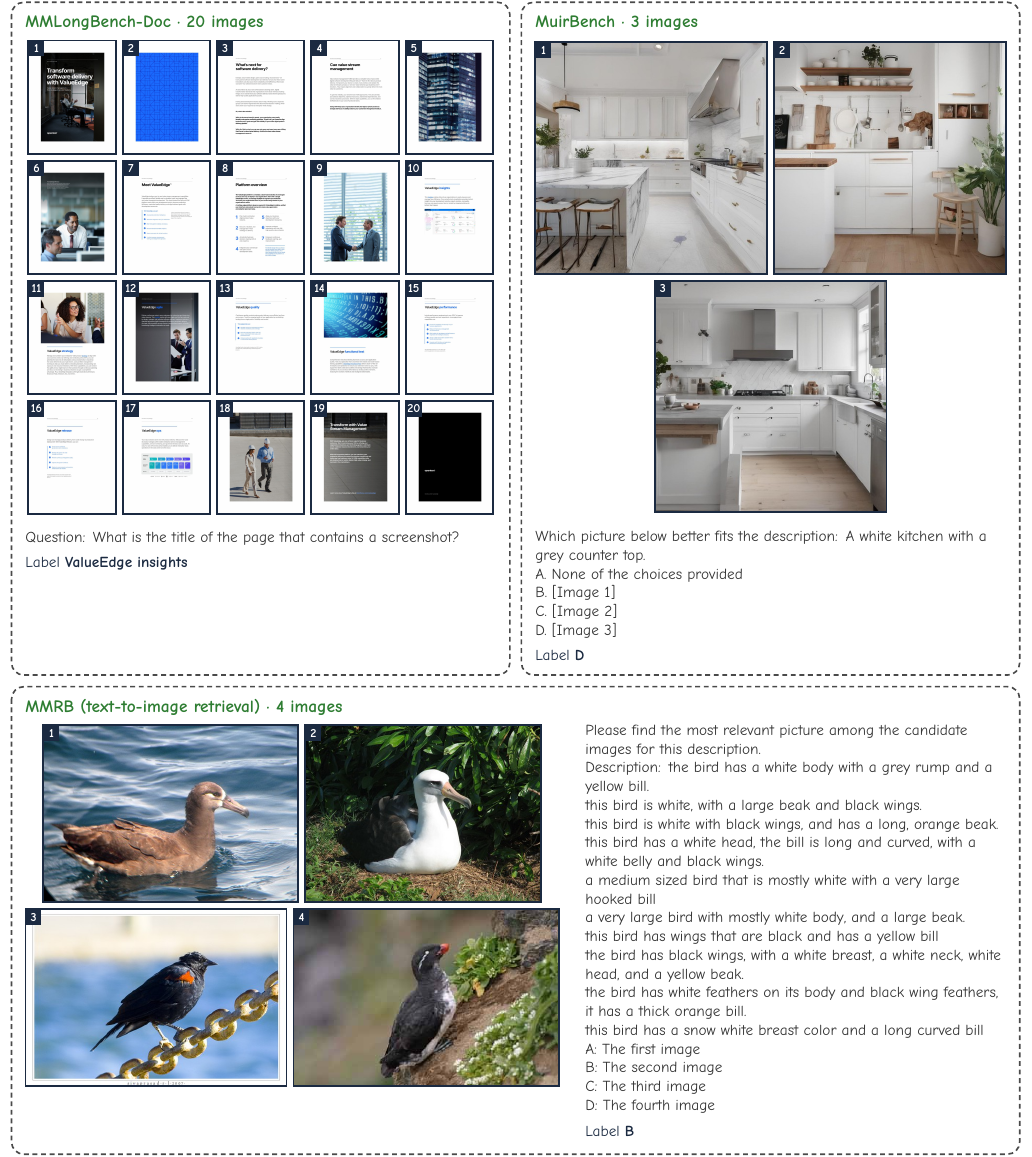}
  \caption{\textbf{Retrieval items.} Three \bench items of the category, shown as in \cref{fig:supp-cat-agg}. In the MMLongBench-Doc item, only Image~10 of the 20 pages shows a product screenshot, and its title is ValueEdge insights. In the MuirBench item, all three kitchens are white, and only the one in Image~3 has a grey counter top. In the MMRB item, Image~2 shows the only one of the four birds with a white body, dark wings and a long, hooked bill.}
  \label{fig:supp-cat-ret}
\end{figure*}

For each of the seven task categories (\cref{sec:supp-groups}), \cref{fig:supp-cat-agg,fig:supp-cat-cmp,fig:supp-cat-rsn,fig:supp-cat-spa,fig:supp-cat-tmp,fig:supp-cat-txt,fig:supp-cat-ret} show \bench items of that category.
We draw only from items that the audit judged valid and whose label and text did not change after the audit.
\clearpage

\section{\benchlite}\label{sec:supp-lite}
\begin{table*}[tp]
\centering\footnotesize
\setlength{\tabcolsep}{4pt}
\definecolor{secband}{HTML}{E3E8F0}
\caption{\textbf{\benchlite leaderboard.} The top block gives the accuracy per task category, its macro average (Avg) and the 95\% interval of the macro average for the MLLMs of \cref{tab:lite}, with total and active parameters for mixture-of-experts models. The bottom block gives the average accuracy (Avg) on \benchlite of the 58 open-source MLLMs of \cref{tab:supp-leaderboard,tab:supp-basedelta} and the three models trained with \cica, ordered by accuracy. Columns are as in \cref{tab:leaderboard}, and the best value per column of the top block is in bold. Qwen3.5-9B and \cicanine answer with thinking enabled (\cref{sec:supp-eval}).}
\label{tab:supp-lite}\label{tab:supp-lite-cat}
{\setlength{\tabcolsep}{4.5pt}%
\begin{tabular}{lrrrrrrrr>{\hspace{4pt}}r>{\hspace{4pt}}c}
\toprule
Model & Size & Agg & Cmp & Rsn & Spa & Tmp & Txt & Ret & Avg & 95\% interval \\
\specialrule{0.5pt}{1pt}{1.2pt}\specialrule{0.5pt}{0pt}{0pt}
\rowcolor{secband}\multicolumn{11}{l}{\textbf{Proprietary}} \\
\specialrule{0.5pt}{0pt}{1pt}
Claude Opus 5.5~\cite{anthropic2026opus55} & -- & 80.3 & 77.0 & 75.4 & 31.1 & \textbf{44.3} & \textbf{100.0} & 67.2 & \textbf{67.9} & [64.2, 71.9] \\
Gemini 3.8 Flash~\cite{google2026gemini38flash} & -- & \textbf{82.0} & 80.3 & 67.2 & 21.3 & 36.1 & 91.2 & 70.5 & 64.1 & [60.1, 68.1] \\
GPT-6-Astra~\cite{openai2026gpt6astra} & -- & 75.4 & 83.6 & \textbf{78.7} & 31.1 & 39.3 & 85.3 & 45.9 & 62.8 & [58.6, 67.3] \\
GPT-6.1-Sol~\cite{openai2026gpt61sol} & -- & 73.8 & \textbf{85.2} & 77.0 & 26.2 & 37.7 & 85.3 & 45.9 & 61.6 & [57.4, 65.8] \\
Qwen3.8-Max~\cite{qwen2026qwen38max} & -- & 73.8 & 77.0 & 72.1 & 32.8 & 31.1 & 85.3 & 54.1 & 60.9 & [56.6, 65.2] \\
Muse Spark 1.3~\cite{meta2026musespark13} & -- & 73.8 & 75.4 & 52.5 & 23.0 & 26.2 & 91.2 & \textbf{75.4} & 59.6 & [55.4, 63.8] \\
\specialrule{0.5pt}{1pt}{1.2pt}\specialrule{0.5pt}{0pt}{0pt}
\rowcolor{secband}\multicolumn{11}{l}{\textbf{Open-source, evaluated on \benchlite only}} \\
\specialrule{0.5pt}{0pt}{1pt}
Qwen3.8-Flash-Next~\cite{qwen2026qwen38flashnext} & 125B-A6B & 68.9 & 72.1 & 63.9 & \textbf{36.1} & 31.1 & 94.1 & 59.0 & 60.7 & [56.1, 65.2] \\
Kimi K3.1 Pro~\cite{moonshot2026kimik31pro} & 2.8T-A104B & 65.6 & 70.5 & 54.1 & 19.7 & 27.9 & 85.3 & 49.2 & 53.2 & [48.9, 57.8] \\
GLM-5.3-Flash~\cite{zai2026glm53flash} & 320B-A18B & 54.1 & 70.5 & 31.1 & 23.0 & 42.6 & 73.5 & 47.5 & 48.9 & [44.0, 53.6] \\
Muse-Glimmer-30B~\cite{meta2026museglimmer30b} & 30B & 29.5 & 62.3 & 18.0 & 24.6 & 27.9 & 94.1 & 57.4 & 44.8 & [40.7, 49.0] \\
MiMo-V2.6-Pro~\cite{xiaomi2026mimov26pro} & 1T-A42B & 39.3 & 54.1 & 26.2 & 13.1 & 21.3 & 85.3 & 36.1 & 39.3 & [35.0, 43.6] \\
Nemotron-3.5-Super-VL~\cite{nvidia2026nemotron35supervl} & 120B-A12B & 29.5 & 34.4 & 21.3 & 11.5 & 11.5 & 94.1 & 32.8 & 33.6 & [29.7, 37.3] \\
\bottomrule
\end{tabular}}

\vspace{8pt}

\makebox[\textwidth][c]{%
\begin{tabular}[t]{lr}
\toprule
Model & Avg \\
\midrule
\oursrow \cicanine (ours) & 47.0 \\
Qwen3.5-9B~\cite{qwen2026qwen35} & 43.3 \\
\oursrow \cicaglm (ours) & 40.2 \\
Qwen3.6-27B~\cite{qwen2026qwen36} & 40.0 \\
GLM-4.1V-9B-Thinking~\cite{hong2025glm41v} & 37.7 \\
MiMo-VL-7B~\cite{xiaomi2025mimovl} & 36.3 \\
Qwen3-VL-4B-Thinking~\cite{qwen2025qwen3vl} & 35.9 \\
Qwen3-VL-32B~\cite{qwen2025qwen3vl} & 35.8 \\
Qwen3-VL-30B-A3B~\cite{qwen2025qwen3vl} & 32.0 \\
\oursrow \cicaseven (ours) & 29.9 \\
Gemma-4-12B~\cite{gemma2026gemma4} & 29.6 \\
Qwen2.5-VL-32B~\cite{bai2025qwen25vl} & 28.3 \\
Time-R1-7B~\cite{wang2025timer1} & 27.9 \\
PAPO-D-7B~\cite{wang2026papo} & 27.7 \\
Migician-7B~\cite{li2025migician} & 27.5 \\
NoisyRollout-7B~\cite{liu2025noisyrollout} & 27.3 \\
Qwen2.5-VL-7B~\cite{bai2025qwen25vl} & 27.2 \\
VisionReasoner-7B~\cite{liu2026visionreasoner} & 27.2 \\
PeRL-7B~\cite{zhang2025perl} & 27.0 \\
Qwen3-VL-8B~\cite{qwen2025qwen3vl} & 26.9 \\
Vision-Zero-7B~\cite{wang2026visionzero} & 26.3 \\
Perception-R1-7B~\cite{xiao2026perceptionr1} & 26.2 \\
\bottomrule
\end{tabular}\hspace{1.5em}%
\begin{tabular}[t]{lr}
\toprule
Model & Avg \\
\midrule
ThinkLite-VL-7B~\cite{wang2025thinklitevl} & 26.2 \\
SophiaVL-R1-7B~\cite{fan2026sophiavlr1} & 26.1 \\
VLAA-Thinker-7B~\cite{chen2025vlaathinker} & 25.7 \\
Qwen3-VL-4B~\cite{qwen2025qwen3vl} & 25.5 \\
PAPO-G-H-7B~\cite{wang2026papo} & 25.3 \\
VPPO-7B~\cite{huang2026vppo} & 25.3 \\
InternVL3-8B~\cite{zhu2025internvl3} & 24.7 \\
VideoRFT-7B~\cite{wang2025videorft} & 24.6 \\
Spatial-SSRL-7B~\cite{liu2026spatialssrl} & 24.4 \\
R1-ShareVL-7B~\cite{yao2025r1sharevl} & 24.2 \\
Video-R1-7B~\cite{feng2025videor1} & 24.0 \\
Visual-Jigsaw-7B~\cite{wu2026visualjigsaw} & 23.6 \\
ArrowRL-7B~\cite{xue2025arrowrl} & 23.4 \\
Gemma-3-12B~\cite{gemma2025gemma3} & 22.9 \\
Vision-SR1-7B~\cite{li2026visionsr1} & 22.7 \\
TON-7B-Math~\cite{wang2025ton} & 22.6 \\
Qwen3-VL-2B~\cite{qwen2025qwen3vl} & 21.0 \\
Open-o3-Video-7B~\cite{meng2026openo3video} & 20.1 \\
LLaVA-OneVision-2-8B~\cite{an2026llavaonevision2} & 19.5 \\
Qwen2-VL-7B~\cite{wang2024qwen2vl} & 18.7 \\
UniVG-R1-7B~\cite{bai2025univgr1} & 18.1 \\
SpatialReasoner-7B~\cite{ma2025spatialreasoner} & 17.9 \\
\bottomrule
\end{tabular}\hspace{1.5em}%
\begin{tabular}[t]{lr}
\toprule
Model & Avg \\
\midrule
Pixtral-12B~\cite{agrawal2024pixtral} & 17.5 \\
Mulberry-7B~\cite{yao2025mulberry} & 16.6 \\
TVC-7B~\cite{sun2025tvc} & 16.4 \\
InternVL3-2B~\cite{zhu2025internvl3} & 15.1 \\
mPLUG-Owl3-7B~\cite{ye2025mplugowl3} & 14.7 \\
Gemma-3-4B~\cite{gemma2025gemma3} & 14.2 \\
Embodied-R1-3B~\cite{yuan2026embodiedr1} & 13.3 \\
SpatialLadder-3B~\cite{li2026spatialladder} & 12.7 \\
LLaVA-OneVision-7B~\cite{li2025llavaonevision} & 12.5 \\
LLaVA-Interleave-7B~\cite{li2025llavainterleave} & 12.3 \\
Idefics2-8B~\cite{laurencon2024idefics2} & 11.5 \\
CapRL-3B~\cite{xing2026caprl} & 11.2 \\
GRIT-3B~\cite{fan2025grit} & 10.5 \\
Qwen2.5-VL-3B~\cite{bai2025qwen25vl} & 10.5 \\
HATCH-3B~\cite{oi2026hatch} & 10.1 \\
Mantis-8B-Idefics2~\cite{jiang2024mantis} & 10.0 \\
MindCube-3B~\cite{wang2026mindcube} & 9.6 \\
\bottomrule
\end{tabular}}
\end{table*}

\cref{tab:supp-lite-cat} gives the accuracy of the MLLMs of \cref{tab:lite} per task category, and \cref{tab:supp-lite-comp} lists the composition of \benchlite.

\subsection{Selection}\label{sec:supp-lite-select}
\benchlite takes the hardest items of \bench under fixed quotas.
We measure the difficulty of an item by the mean accuracy on it of the open-source base models that we evaluate on \bench, that is, the models that are not post-trained from another evaluated model.
Items enter hardest first, with ties broken in a fixed random order.
Post-trained MLLMs, including those trained in this paper, and proprietary MLLMs take no part in the selection.
Each task category receives 61 items, except text/web, the smallest category of \bench, which receives 34.
Within each category the items fill target shares of five image-count bins (2, 3--4, 5--8, 9--16 and 17 or more images; \cref{tab:supp-lite-comp}), and these shares are relaxed only where too few items qualify.
No source benchmark supplies more than 19 items of a category or 11 of text/web.

\begin{table}[t]
\centering\footnotesize
\setlength{\tabcolsep}{3.2pt}
\caption{\textbf{Composition of \benchlite.} Items per task category and per number of images, and the number of source benchmarks per category.}
\label{tab:supp-lite-comp}
\begin{tabular}{@{}lrrrrrrr@{}}
\toprule
 & & \multicolumn{5}{c}{Images per item} & \\
\cmidrule(lr){3-7}
Category & Items & 2 & 3--4 & 5--8 & 9--16 & 17+ & Sources \\
\midrule
Aggregate & 61 & 28 & 21 & 11 & 1 & 0 & 8 \\
Compare/change & 61 & 43 & 11 & 4 & 3 & 0 & 10 \\
Reasoning/verification & 61 & 39 & 11 & 10 & 0 & 1 & 9 \\
Spatial correspondence & 61 & 29 & 15 & 14 & 3 & 0 & 8 \\
Temporal/action & 61 & 1 & 10 & 33 & 7 & 10 & 4 \\
Text/web & 34 & 23 & 3 & 5 & 3 & 0 & 6 \\
Retrieval & 61 & 8 & 11 & 19 & 4 & 19 & 5 \\
\midrule
Total & 400 & 171 & 82 & 96 & 21 & 30 & 19 \\
\bottomrule
\end{tabular}
\end{table}

\subsection{Verification}\label{sec:supp-lite-verify}
We manually verify all 400 items of \benchlite against their images and labels, checking that the label holds, that no second answer is defensible, that no single image answers the item and that no image is defective.
An item that fails is replaced by the next item of its category under the same quotas, and we verify the replacement in the same way.
\subsection{Rank agreement and paired intervals}\label{sec:supp-lite-rank}
On \benchlite, the ranking of the 58 open-source MLLMs evaluated on both benchmarks, including the post-trained ones that take no part in the selection, follows their ranking on \bench, with a Spearman rank correlation of 0.93 between their accuracies on the two (\cref{tab:supp-lite,tab:supp-leaderboard,tab:supp-basedelta}).
In paired comparisons over the 400 items (bootstrap~\cite{efron1979bootstrap} over items stratified by category, 10,000 resamples), Claude Opus 5.5~\cite{anthropic2026opus55} leads Qwen3.8-Flash-Next~\cite{qwen2026qwen38flashnext}, the strongest open-source MLLM, by 7.2 points (95\% interval $[2.0, 12.4]$).

\subsection{Proprietary MLLMs}\label{sec:supp-lite-prop}
The six proprietary MLLMs receive the prompts and image files of the open-source runs through the APIs and with the identifiers and settings of \cref{tab:supp-lite-eval}, and the rule-based scorer of \cref{sec:supp-scorer} extracts every answer from the returned text.
Each of them runs with the high reasoning-effort setting of its API.
\begin{table*}[t]
\centering\footnotesize
\setlength{\tabcolsep}{4pt}
\caption{\textbf{Proprietary MLLMs on \benchlite.} API identifier and interface of the proprietary MLLMs of \cref{tab:lite}. Each request carries the prompt and image files of the open-source runs, allows at most 65,536 output tokens and sets no temperature, so each API samples at its default.}
\label{tab:supp-lite-eval}
\begin{tabular}{lll}
\toprule
Model & API identifier & Interface \\
\midrule
Claude Opus 5.5~\cite{anthropic2026opus55} & \texttt{claude-opus-5-5} & Anthropic API \\
Gemini 3.8 Flash~\cite{google2026gemini38flash} & \texttt{google/gemini-3.8-flash} & OpenRouter (Google AI Studio) \\
GPT-6-Astra~\cite{openai2026gpt6astra} & \texttt{gpt-6-astra} & OpenAI API \\
GPT-6.1-Sol~\cite{openai2026gpt61sol} & \texttt{gpt-6.1-sol} & OpenAI API \\
Qwen3.8-Max~\cite{qwen2026qwen38max} & \texttt{qwen/qwen3.8-max-0902} & OpenRouter (Alibaba) \\
Muse Spark 1.3~\cite{meta2026musespark13} & \texttt{meta/muse-spark-1.3} & OpenRouter (Meta) \\
\bottomrule
\end{tabular}
\end{table*}

\section{Scoring and evaluation}\label{sec:supp-evaluation}
\subsection{Rule-based scorer}\label{sec:supp-scorer}
Each item carries an answer format derived from its prompt.
The format is an option letter (with the option texts), yes/no or true/false, an integer or a number with the units the question permits, or a short text with a list of accepted aliases, and rules derive it from the source task, the question and the options.
A number matches its label exactly unless the official rules of its task allow a tolerance, as the rules of ReMI~\cite{kazemi2024remi} do for numbers (1\%), geometry (3\%) and clock readings (ten minutes) and those of MMLongBench-Doc~\cite{ma2024mmlongbenchdoc} for decimal answers (1\%).
A list answer scores only if it names exactly the items of its label, in any order.
The scorer takes the last explicit answer (such as an \texttt{Answer:} line, a boxed answer, or a bare answer when the prompt asks for one), rejects hedged answers that name several options, and never consults the ground truth when extracting.
It uses no LLM judge.
It makes no extraction error on 2.66M synthetic adversarial outputs (correct, wrong, empty, hedged and permuted answers).
On 600 outputs of GLM-4.1V-9B-Thinking~\cite{hong2025glm41v}, which reasons at length before each answer, an annotator blind to the scorer's decisions agrees with it on 599.

\subsection{Official scoring code}\label{sec:supp-scorer-shift}
\begin{table}[t]
\centering\footnotesize
\setlength{\tabcolsep}{1.5pt}
\caption{\textbf{Official scoring code shifts per-benchmark accuracy in both directions.} Official minus \bench scorer accuracy on the same generations of 21 models on a fixed sample of $n$ items of each benchmark, the same for all 21 models, with the mean, the range over models, and the model pairs (of 210) whose order reverses (Rev.). $^{\dagger}$ marks a mean whose 95\% interval excludes zero.}
\label{tab:scorer}
\begin{tabular}{@{}l >{\raggedright\arraybackslash}p{0.258\linewidth} r@{\hspace{6.5pt}}r r@{}}
\toprule
Benchmark & Main difference & Mean & Min\,/\,max & Rev. \\
\midrule
\begin{tabular}[t]{@{}l@{}}EMMA~\cite{hao2025emma}\\[-1pt]{\scriptsize $n=63$}\end{tabular} & letter from the reasoning text & $+3.3$\rlap{$^{\dagger}$} & $-1.6$\,/\,$+17.5$ & 17 \\
\begin{tabular}[t]{@{}l@{}}MMLongBench-Doc~\cite{ma2024mmlongbenchdoc}\\[-1pt]{\scriptsize $n=198$}\end{tabular} & fuzzy match passes wrong lists & $+2.7$\rlap{$^{\dagger}$} & $-0.5$\,/\,$+10.1$ & 5 \\
\begin{tabular}[t]{@{}l@{}}MV-MATH~\cite{wang2025mvmath}\\[-1pt]{\scriptsize $n=177$}\end{tabular} & letter from the reasoning text & $+1.5$\rlap{$^{\dagger}$} & $0.0$\,/\,$+5.1$ & 5 \\
\begin{tabular}[t]{@{}l@{}}DUDE~\cite{vanlandeghem2023dude}\\[-1pt]{\scriptsize $n=73$}\end{tabular} & near misses pass, reformatted answers fail & $+0.3$ & $-5.5$\,/\,$+8.2$ & 5 \\
\begin{tabular}[t]{@{}l@{}}ReMI~\cite{kazemi2024remi}\\[-1pt]{\scriptsize $n=473$}\end{tabular} & rejects alternative golds, passes hedges & $-0.2$ & $-3.6$\,/\,$+9.3$ & 25 \\
\begin{tabular}[t]{@{}l@{}}MIRB~\cite{zhao2024mirb}\\[-1pt]{\scriptsize $n=107$}\end{tabular} & rejects number words, explained yes/no & $-2.3$\rlap{$^{\dagger}$} & $-10.3$\,/\,$0.0$ & 8 \\
\begin{tabular}[t]{@{}l@{}}MMRB~\cite{cheng2025mmrb}\\[-1pt]{\scriptsize $n=469$}\end{tabular} & matches empty image-option text & $-2.7$\rlap{$^{\dagger}$} & $-4.9$\,/\,$-1.3$ & 3 \\
\begin{tabular}[t]{@{}l@{}}VLM2-Bench~\cite{zhang2025vlm2bench}\\[-1pt]{\scriptsize $n=464$}\end{tabular} & rejects \texttt{A)}, number words & $-3.4$\rlap{$^{\dagger}$} & $-8.6$\,/\,$0.0$ & 20 \\
\bottomrule
\end{tabular}
\end{table}

On the same generations of 21 models, the official code shifts per-benchmark accuracy in both directions relative to the rule-based scorer of \cref{sec:supp-scorer} and reverses the order of model pairs within a benchmark (\cref{tab:scorer}).
String similarity, which scores free-form answers in DUDE~\cite{vanlandeghem2023dude} and in the CLEVR-Change~\cite{park2019clevrchange} subset of MileBench~\cite{song2024milebench}, compares the wording of an answer with the reference, so a correct answer in another format can score below a near miss.
On multiple-choice benchmarks, the official code and the rule-based scorer also extract some answers differently.

\subsection{Evaluation settings}\label{sec:supp-eval}
All models use the same prompts. Each prompt follows the official instruction of its source where one exists and asks for the answer in the format of the item (\cref{sec:supp-scorer}).
Every open-source model evaluated on \bench has a budget of at least 8,192 output tokens per answer.
Qwen3.5-9B~\cite{qwen2026qwen35} and the models trained from it are evaluated with thinking enabled, using thinking-mode sampling from the model card (temperature 0.6, top-p 0.95, top-k 20).
They use the same decoding in the gray-masking tests of \cref{sec:shortcut} and \cref{fig:gain}.
All other open-source models evaluated on \bench, including Qwen2.5-VL-7B, GLM-4.1V-9B-Thinking and the models trained from them, decode greedily.
GLM-4.1V-9B-Thinking~\cite{hong2025glm41v}, Qwen3-VL-4B-Thinking~\cite{qwen2025qwen3vl} and MiMo-VL-7B~\cite{xiaomi2025mimovl} think before every answer.
InternVL3~\cite{zhu2025internvl3} models use the multi-image tiling rule of VLMEvalKit~\cite{duan2024vlmevalkit}, LLaVA-OneVision~\cite{li2025llavaonevision} its official multi-image setting, and the other models their default image processing.

\section{Full specification of \texorpdfstring{\cica}{CICA}}\label{sec:supp-cica}
This section states the rules of \cref{alg:cica}, and \cref{tab:supp-cica-defaults} lists the settings.

\begin{algorithm}[t]
\caption{One training step with \cica.}
\label{alg:cica}
\small
\begin{algorithmic}[1]
\Require policy $\pi_\theta$, items $\mathcal{D}$, group size $G$, items per step $B$, attempt limit $M$, minimum $B_{\min}$ (\cref{tab:supp-cica-defaults}), $\beta$, $\tau$
\State $\mathcal{S}\gets\emptyset$ \Comment{kept groups}
\While{$|\mathcal{S}|<B$ and fewer than $M$ attempts}
  \State draw an item $x$ from $\mathcal{D}$
  \State sample $y_{1..G}$ from the rollout sampler $\mu(\cdot\mid x)$ with rewards $r_{1..G}$
  \State \textbf{if} all $r_i$ are equal \textbf{then continue}
  \State add the group $\{(x,y_i,A_i)\}_{i=1}^{G}$ to $\mathcal{S}$, with $A_i=(r_i-\bar r)/(\mathrm{std}(r)+\epsilon)$
\EndWhile
\State \textbf{if} $|\mathcal{S}|<B_{\min}$ \textbf{then} skip the step
\For{each response $y_i$ in $\mathcal{S}$, of an input $x_i$ with $K_i$ images}
  \State $\ell_{i,t}\gets\log\pi_\theta(y_{i,t}\mid x_i,y_{i,<t})$ \Comment{with gradient}
  \State $\ell^{(k)}_{i,t}\gets\log\pi_\theta(y_{i,t}\mid x_i^{(k)},y_{i,<t})$, $k{=}1..K_i$ \Comment{no grad.}
  \State $k^*_{i,t}$ and $\sigma_{i,t}$ by \cref{eq:kstar,eq:sigma}
\EndFor
\State $w\gets$ \cref{eq:w}, centered over each set $\mathcal{T}_i$ of active tokens
\State AdamW~\cite{loshchilov2019adamw} step on \cref{eq:loss9} with $w$ fixed
\end{algorithmic}
\end{algorithm}

\subsection{\texorpdfstring{\cica}{CICA} objective}\label{sec:supp-cica-objective}
Each sampled response $i$ of a step has the advantage $A_i=(r_i-\bar r)/(\mathrm{std}(r)+\epsilon)$ over its group, with $\epsilon=10^{-6}$ and the population standard deviation over the $G$ responses (\cref{eq:adv} omits $\epsilon$). The reward $r_i\in[0,1]$ is the accuracy of the extracted answer, and a response counts as correct when $r_i=1$.
Let $N$ be the number of responses, $L_i$ the length of $y_i$ including end-of-turn, and $\ell_{i,t}=\log\pi_\theta(y_{i,t}\mid x_i,y_{i,<t})$.
Each backbone takes one update per step on the GRPO~\cite{shao2024deepseekmath} objective with \cica weights,
\begin{equation}
  \mathcal{L}=-\frac1N\sum_i\frac{A_i}{L_i}\sum_t \eta_{i,t}w_{i,t}\ell_{i,t}
  +\frac{\beta_{\mathrm{KL}}}{T}\sum_i\sum_t\big(e^{\delta_{i,t}}-\delta_{i,t}-1\big), \label{eq:loss9}
\end{equation}
where $\delta_{i,t}$ is the log-ratio of the reference model $\pi_{\mathrm{ref}}$ to $\pi_\theta$ clipped to $[-5,5]$, $\beta_{\mathrm{KL}}=0.02$, and $T$ is the number of response tokens before end-of-turn.
The sum over $t$ runs over all $L_i$ tokens of $y_i$ in the first term and over those before end-of-turn in the second, and $w_{i,t}$ carries no gradient.
Here $\eta_{i,t}=\min(\pi_\theta/\mu,2)$ is the truncated importance weight of each token, without gradient, where $\mu$ is the probability that the rollout sampler gave the token, so $\eta_{i,t}=1$ for a sampler that matches the policy.
\cica changes only the policy term.
The KL term uses a reference pass on the same responses and is not weighted by $w$.

Groups with equal accuracy rewards are discarded and new items drawn, steps with too few kept groups are skipped, and gradients are clipped to norm 1 (limits in \cref{tab:supp-cica-defaults}). A step draws items without replacement from a shuffled pool, skips items whose prompt alone exceeds 32,768 tokens and drops a group if the prompt with any of its responses would exceed that length.

\begin{table*}[t]
\centering\small
\setlength{\tabcolsep}{5pt}
\caption{\textbf{Settings of the \cica training step.} \cica builds on GRPO on every backbone, and a value spanning several columns applies to each of them.}
\label{tab:supp-cica-defaults}
\begin{tabular}{@{}llll@{}}
\toprule
 & Qwen2.5-VL-7B~\cite{bai2025qwen25vl} & Qwen3.5-9B~\cite{qwen2026qwen35} & GLM-4.1V-9B-Thinking~\cite{hong2025glm41v} \\
\midrule
Items $B$, group size $G$ & 8, 8 & \multicolumn{2}{c}{32, 8} \\
Rollout sampler & Transformers~\cite{wolf2020transformers} & \multicolumn{2}{c}{vLLM~\cite{kwon2023vllm}} \\
Temperature; top-$p$ & \multicolumn{3}{c}{1; 0.95} \\
Top-$k$; repetition penalty & 50; 1.05 & \multicolumn{2}{c}{20; none} \\
Rollout budget & up to 3,072 tokens & \multicolumn{2}{c}{8,192 tokens} \\
Learning rate (cosine schedule~\cite{loshchilov2017sgdr}) & $5\times10^{-7}$ & \multicolumn{2}{c}{$3\times10^{-7}$} \\
\cica $\beta$; $\tau$ & \multicolumn{3}{c}{0.95, ramped over the first 32 steps; 0.5} \\
Single-image input & \multicolumn{3}{c}{other images deleted, batched ($\le$32,768 tokens)} \\
Updates per step & \multicolumn{3}{c}{1} \\
Policy surrogate & \multicolumn{3}{c}{no ratio; sampler weight $\min(\pi_\theta/\mu,2)$} \\
Normalization & \multicolumn{3}{c}{per response, then $N$} \\
KL to reference model & \multicolumn{3}{c}{k3~\cite{schulman2020klapprox}, $\beta_{\mathrm{KL}}=0.02$} \\
Overlong penalty & \multicolumn{3}{c}{none} \\
Equal-accuracy groups & \multicolumn{3}{c}{discarded, next item drawn} \\
Attempt limit $M$ per step & 256 & \multicolumn{2}{c}{128} \\
Minimum kept groups $B_{\min}$ & 4 & \multicolumn{2}{c}{12} \\
Maximum length & \multicolumn{3}{c}{32,768} \\
Gradient clip; optimizer & \multicolumn{3}{c}{norm 1; AdamW, no weight decay} \\
Frozen & \multicolumn{3}{c}{vision encoder and projector} \\
\bottomrule
\end{tabular}
\end{table*}

\subsection{Single-image inputs of \texorpdfstring{\cica}{CICA}}\label{sec:supp-cica-single}
The input $x^{(k)}$ keeps the visual tokens of image $k$ and removes those of the other images from the context.
The text, the tokens that open and close each image, and the position indices of all kept tokens are those of $x$, so only the other images are missing.
The vision encoder thus encodes only image $k$.
In exact arithmetic, deleting the visual tokens of the other images gives the same log-probabilities as masking them in attention layers, and it also keeps them out of the recurrent state of linear-attention layers.
The sequences of all responses and images are batched up to 32,768 tokens including padding, and all single-image passes use the parameters being trained and run without gradients.

\subsection{Answer tokens in \texorpdfstring{\cica}{CICA}}\label{sec:supp-cica-answer}
For each response, $\mathrm{ans}_i$ is the answer that the reward scores.
If the response has a line \texttt{Answer:}\,\ldots{} (also \texttt{Final answer:}, with optional bullet or bold), $\mathrm{ans}_i$ covers the extracted answer on the last such line, after any \texttt{**}, parenthesis, quote, brace or \texttt{\textbackslash boxed\{} and without a closing period or bracket.
Without such a line, $\mathrm{ans}_i$ is the content of the last \texttt{\textbackslash boxed\{\}}, or the bare answer when the whole response is one line of at most 40 characters.
A response whose answer cannot be located, such as one that never ends its thinking, or whose bare answer is not in the canonical format of its item, gets $s_i=0$.
In a response that thinks before answering, $\mathrm{ans}_i$ is read from the text after the end of thinking, from the answer box that the chat template marks, if any.
The answer marker runs from the start of the last answer line to $\mathrm{ans}_i$.
Every token takes its own best single image (\cref{eq:kstar}), and $\mathrm{ans}_i$ serves to locate the answer and its marker.

\subsection{Active tokens and \texorpdfstring{\cica}{CICA} weights}\label{sec:supp-cica-active}
The active tokens of a response that reasons before its answer are its tokens except the answer marker, the answer tags, the end-of-turn and stop tokens, the tokens that open or close thinking, and the token right after the end of thinking, and $\mathcal{T}_i$ holds these tokens.
A response that gives only its answer has its answer $\mathrm{ans}_i$ as active tokens, too few to compare.
The set $\mathcal{T}_i$ therefore pools them with those of the other answer-only responses of its group that write their answer the same way, either as an answer line (including \texttt{\textbackslash boxed\{\}}) or as a bare answer.
The responses of an item are processed together, and a pool with a single response is given $u=0$.

The indicator $s_i$ is 1 for a response that is correct (accuracy 1) with $A_i>0$ or wrong with $A_i<0$.
It is 0 otherwise, as for a partially correct answer above its group mean or a response without a usable answer.
A response with $s_i=0$ keeps $w=1$ and joins no pool.
In $\mathcal{T}_i$, $\bar\sigma_i$ and $\bar h_i$ are means in which every token of response $j$ counts with $a_j=|A_j|/(NL_j)$, its share of the update, so within one response they are plain means.
Since $\sum a_ju_{j,t}=0$ over $\mathcal{T}_i$, the weights do not change the total size $\sum a_jw_{j,t}$ of the update over $\mathcal{T}_i$ before the importance weight $\eta$, and they average to one over the active tokens of a response that reasons.
A pooled answer is thus reweighted against the other answers to its item, and an answer carries more of the item's update when the full input favors it more than its best single image does, relative to these answers.
At step $n$ the weight uses $\beta\min(1,(n-1)/32)$, which reaches 0.95 at step 33.

\subsection{Choice of contrast for \texorpdfstring{\cica}{CICA}}\label{sec:supp-cica-contrast}
\cica contrasts the full input with its best single image, and not with an average or a random single image or with the full input minus one image.
Because $k^*_{i,t}$ maximizes the likelihood of each token (\cref{eq:kstar}), $\sigma_{i,t}$ is the smallest contrast that any single image gives at that token, and it is positive only if the full input makes the token more likely than every single image does.
A response that reasons across several images can draw on a different image at each step.
For a one-token answer, the best single image of the token is the image that best explains the answer.
An average or a random single image, like the text alone, would give a high $\sigma$ to an answer read off one image, because neither the other images nor the text alone supports that answer.
A leave-one-out contrast, which hides one image, is high for any token that depends on that image, including an answer read off it alone.
Each of these contrasts would credit the single-image shortcut.
Against the best single image, $\sigma$ is high only where the other images add to the most informative one.
Hiding the other images keeps the text, the image delimiters and all positions in place, so a shift in log-probability that the reduced input causes equally on every active token cancels in $\sigma-\bar\sigma$.

The answers of the base models bear this out.
On the items of \cref{tab:supp-mask}, we compute $\sigma$ as in \cref{eq:sigma} for the answer that Qwen2.5-VL-7B or Qwen3.5-9B gives with all images, with the probability of each answer renormalized over the valid answers of the item.
For a correct answer, the median $\sigma$ is 0.0004 for both models on the items that the audit removed because one image answers them, against 0.12 and 0.19 on \bench items.
In 72\% and 73\% of the pairs of a correct answer on a removed item and a correct answer on an \bench item, the \bench answer has the larger $\sigma$.
With the mean single-image log-likelihood as the reference, these shares are 62\% and 66\%, and with the text alone 39\% and 44\%, since both references give a high $\sigma$ to an answer read off one image.
Two rows of \cref{tab:ablation} replace the best single image with the text alone and with the mean single-image log-likelihood.

\subsection{\texorpdfstring{\cica}{CICA} weights on real responses}\label{sec:supp-cica-examples}
\begin{figure*}[tp]
  \centering
  \includegraphics[width=\linewidth]{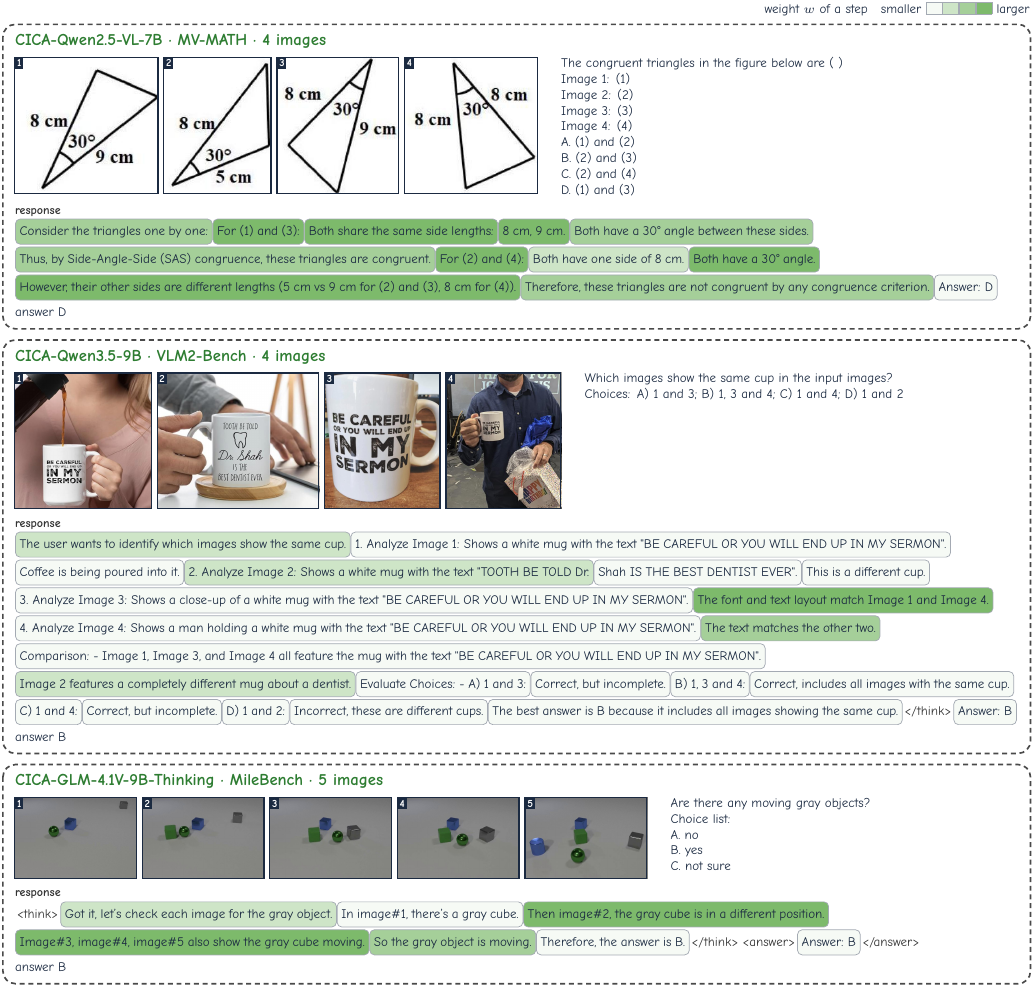}
  \caption{\textbf{Weights of \cica on whole responses.} A correct response of \cicaseven (top), \cicanine (middle) and \cicaglm (bottom) on \bench items, cut into steps as in \cref{fig:cica}, each step shaded by the mean weight $w$ (\cref{eq:w}) of its tokens, darker for a larger weight. Gray text and the answer marker that opens the last box are not weighted. Answer-format instructions of the prompts are not shown.}
  \label{fig:supp-tokw}
\end{figure*}

\cref{fig:supp-tokw} shows a whole correct response of each of \cicaseven, \cicanine and \cicaglm on a different \bench item, cut into steps as in \cref{fig:cica}, with each step shaded by the mean weight $w$ (\cref{eq:w}) of its tokens, as the training code computes it with the trained model.
Steps that relate the images, such as comparing side lengths and angles across triangles, matching the text on a cup across photos or tracking a cube across frames, are among the darkest, and most steps that describe one image are lighter.

\subsection{Cost of \texorpdfstring{\cica}{CICA}}\label{sec:supp-cica-cost}

\begin{figure*}[tp]
  \centering
  \includegraphics[width=\linewidth]{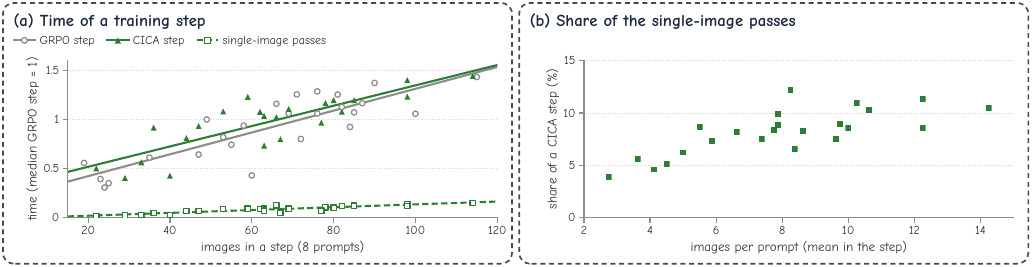}
  \caption{\textbf{Training time and the number of images.} Steps of GRPO and \cica on Qwen2.5-VL-7B, run side by side on one node, against the number of images of the eight prompts of each step. (a) Time of a step and of the single-image passes of \cica, in units of the median step of GRPO, with least-squares lines. (b) Share of the single-image passes in each step of \cica, against the mean number of images per prompt in the step.}
  \label{fig:supp-cost}
\end{figure*}

\cica adds no parameters and no backward pass.
Per response, it evaluates the $K$ single-image inputs without gradients.
Their visual tokens add up to those of $x$, and the text is repeated $K$ times.
A step of GRPO already samples the responses and runs a reference pass and a forward and backward pass over each of them.
The single-image passes reuse these sampled responses, so a training step needs no further sampling and no further backward pass, and only these forward passes add to its time.
Run as one batch per prompt, they take about 9\% of a training step on Qwen2.5-VL-7B.
With more images, a step of \cica grows at about the rate of a step of GRPO rather than $K$-fold (\cref{fig:supp-cost}).
The vision encoder runs once per prompt for all of its sampled responses.
Each image enters only one of the $K$ single-image inputs, so the single-image passes process each image once per response, while a step of GRPO already runs each image through the reference pass and the forward and backward pass of every sampled response.
Per image in a step, a GRPO step grows by 1.1\% of the median GRPO step, and the single-image passes add 0.14\%, about one eighth of that.

The single-image passes keep no activations and run in bounded batches, and on Qwen3.5-9B the peak allocated memory of \cica does not exceed that of GRPO.

\section{Training details}\label{sec:supp-train}
All runs start from the public checkpoints and keep the vision encoder and projector frozen, and the runs on one backbone share rewards and prompts.
On every backbone \cica trains for one epoch over all 49,105 items of \bench-Train with the settings of \cref{tab:supp-cica-defaults}, and \cref{sec:supp-baselines} describes the RL baselines.
\subsection{Backbones}\label{sec:supp-backbones}
We train Qwen2.5-VL-7B~\cite{bai2025qwen25vl}, Qwen3.5-9B~\cite{qwen2026qwen35} and GLM-4.1V-9B-Thinking~\cite{hong2025glm41v} with \cica.
On each backbone \cica reweights the advantages of GRPO~\cite{shao2024deepseekmath} with a KL penalty of 0.02 to the base model (\cref{eq:loss9}).
The three backbones differ in the language model, the vision encoder, the position encoding and the pre-training.
Unlike Qwen2.5-VL-7B, Qwen3.5-9B uses linear attention in three of every four language-model layers, another vision encoder and position encoding, and early-fusion multimodal pre-training.
GLM-4.1V-9B-Thinking comes from another model family and uses full attention in every language-model layer, as Qwen2.5-VL-7B does.
\cica forms the single-image inputs of every backbone in the same way, by deleting the visual tokens of the other images (\cref{sec:supp-cica}).

\subsection{Training data}\label{sec:supp-train-data}
\bench-Train (\cref{sec:train-set}) has 49,105 multi-image items with verifiable answers from 48 sources, with near-duplicates of \bench images removed. It holds 26,443 items from the training splits of ten public datasets (MindCube~\cite{wang2026mindcube}, Perception Test~\cite{patraucean2023perceptiontest}, MagicBrush~\cite{zhang2023magicbrush}, NLVR2~\cite{suhr2019nlvr2}, WebQA~\cite{chang2022webqa}, SAT~\cite{ray2025sat}, CLEVRER~\cite{yi2020clevrer}, STAR~\cite{wu2021star}, CLEVR-Change~\cite{park2019clevrchange} and Spot-the-Diff~\cite{jhamtani2018spotthediff}).
Another 9,570 items come from multi-image instruction collections (M4-Instruct~\cite{li2025llavainterleave} 6,467, Mantis-Instruct~\cite{jiang2024mantis} 1,049, LEOPARD~\cite{jia2025leopard} 2,039, other 15), and 13,092 are synthetic items built on public images (jigsaw, forensic, art-style and similarity matching, visual and semantic correspondence, counting, needle-in-a-haystack, odd-one-out, size comparison and frame ordering tasks, and question templates on common-object images).

A training image and an \bench image form a candidate pair when their 64-bit difference hashes~\cite{krawetz2013dhash} differ in at most 6 bits, and a candidate pair is a near-duplicate when the two images, converted to grayscale and resized to $32{\times}32$ pixels, correlate at 0.99 or more after mean subtraction. Near-duplicates include identical frames and other frames of one scene, such as frames of one fixed camera in Spot-the-Diff or renders of one CLEVR~\cite{johnson2017clevr} scene.
This check removes 1,756 items from the training pool.
\newpage
\subsection{RL baselines}\label{sec:supp-baselines}
On each backbone the five RL baselines match \cica in data, the accuracy reward, prompts, group size, batch size and learning-rate schedule, and each otherwise follows the settings of its paper or official code.
Every run is reported at its last checkpoint.
On a multi-image item, VGPO~\cite{wang2026vgpo}, APPO~\cite{du2026appo} and CFPO~\cite{yu2026cfpo} act on the visual tokens of all its images.
Their official implementations handle several images or video frames per input.
VPPO~\cite{huang2026vppo} and PAPO~\cite{wang2026papo} enter \cref{tab:supp-leaderboard} with their released checkpoints.
GRPO~\cite{shao2024deepseekmath} and DAPO~\cite{yu2025dapo} give every token of a response the same advantage, and the visual terms of VGPO, APPO and CFPO favor tokens that draw on the images, on key frames or on key visual cues, with no term that sets apart an answer read off one image.
\begin{table*}[tp]
\centering\footnotesize
\setlength{\tabcolsep}{4.5pt}
\renewcommand{\arraystretch}{0.94}
\definecolor{secband}{HTML}{E3E8F0}\def\famsep{\arrayrulecolor{black!40}\specialrule{0.4pt}{0.6pt}{2pt}\arrayrulecolor{black}}\def\ours{\oursrow}\def\m#1{\smash{$^{#1}$}}
\caption{\textbf{\bench leaderboard by task category.} Most models of \cref{tab:leaderboard} and further base and post-trained models, with the average on \benchlite (Lite). Columns, groups and the $^{\S}$ mark are as in \cref{tab:leaderboard}, and in each group the best value per column before rounding is in bold and the second best is underlined. $^{\dagger}$The micro average, the accuracy over all items of \bench. Qwen3.5-9B and \cicanine answer with thinking enabled.}
\label{tab:supp-leaderboard}
\begin{tabular}{llrrrrrrr>{\hspace{4pt}}rr>{\hspace{6pt}}r}
\toprule
& & \multicolumn{9}{c}{\bench} & \multicolumn{1}{c}{Lite} \\
\cmidrule(lr){3-11}\cmidrule(l){12-12}
Model & Size & Agg & Cmp & Rsn & Spa & Tmp & Txt & Ret & Avg & Avg$^{\dagger}$ & Avg \\
\specialrule{0.5pt}{1pt}{1.2pt}\specialrule{0.5pt}{0pt}{0pt}
\rowcolor{secband}\multicolumn{12}{l}{\textbf{Base models}} \\
\specialrule{0.5pt}{0pt}{1pt}
Gemma-3-4B~\cite{gemma2025gemma3} & 4B & 33.1 & 25.9 & 38.8 & 31.1 & 25.7 & 41.6 & 30.8 & 32.4 & 31.6 & 14.2 \\
Gemma-3-12B~\cite{gemma2025gemma3} & 12B & 45.4 & 34.9 & 44.2 & 37.7 & 28.6 & 46.8 & 38.5 & 39.4 & 38.0 & 22.9 \\
Gemma-4-12B~\cite{gemma2026gemma4} & 12B & 53.6 & 42.6 & 46.5 & 43.3 & 36.1 & 75.3 & 49.2 & 49.5 & 44.3 & 29.6 \\
\famsep
GLM-4.1V-9B-Thinking~\cite{hong2025glm41v} & 9B & 64.3 & 58.2 & 66.4 & 51.8 & 44.7 & 70.1 & 55.3 & 58.7 & 56.3 & 37.7 \\
\famsep
Idefics2-8B~\cite{laurencon2024idefics2} & 8B & 25.1 & 25.1 & 31.6 & 27.3 & 25.0 & 33.1 & 18.1 & 26.5 & 26.5 & 11.5 \\
\famsep
InternVL3-2B~\cite{zhu2025internvl3} & 2B & 34.2 & 30.1 & 30.2 & 27.9 & 34.7 & 41.6 & 27.6 & 32.3 & 30.4 & 15.1 \\
InternVL3-8B~\cite{zhu2025internvl3} & 8B & 49.5 & 43.5 & 39.9 & 37.7 & 42.8 & 44.8 & 38.3 & 42.4 & 40.7 & 24.7 \\
\famsep
LLaVA-Interleave-7B~\cite{li2025llavainterleave}\m{\S} & 7B & 37.0 & 37.8 & 30.4 & 29.6 & 25.9 & 35.7 & 17.0 & 30.5 & 29.0 & 12.3 \\
LLaVA-OneVision-7B~\cite{li2025llavaonevision} & 7B & 36.4 & 30.8 & 33.0 & 31.1 & 22.1 & 35.7 & 18.9 & 29.7 & 28.7 & 12.5 \\
LLaVA-OneVision-2-8B~\cite{an2026llavaonevision2} & 8B & 40.4 & 40.1 & 41.5 & 38.7 & 37.0 & 42.9 & 34.2 & 39.3 & 38.8 & 19.5 \\
\famsep
MiMo-VL-7B~\cite{xiaomi2025mimovl} & 7B & 65.0 & 63.9 & 67.3 & 52.5 & 40.0 & 72.7 & 54.5 & 59.4 & 56.5 & 36.3 \\
\famsep
mPLUG-Owl3-7B~\cite{ye2025mplugowl3}\m{\S} & 7B & 29.1 & 26.5 & 30.4 & 27.8 & 30.6 & 33.1 & 20.0 & 28.2 & 28.0 & 14.7 \\
\famsep
Pixtral-12B~\cite{agrawal2024pixtral} & 12B & 42.3 & 33.6 & 32.7 & 35.1 & 23.3 & 43.5 & 23.1 & 33.4 & 31.1 & 17.5 \\
\famsep
Qwen2-VL-7B~\cite{wang2024qwen2vl} & 7B & 40.6 & 33.6 & 36.9 & 36.9 & 34.2 & 40.3 & 38.9 & 37.3 & 36.5 & 18.7 \\
\famsep
Qwen2.5-VL-3B~\cite{bai2025qwen25vl} & 3B & 32.8 & 24.2 & 33.2 & 29.7 & 34.6 & 44.2 & 33.0 & 33.1 & 31.5 & 10.5 \\
Qwen2.5-VL-7B~\cite{bai2025qwen25vl} & 7B & 45.8 & 38.3 & 40.0 & 42.0 & 36.1 & 40.3 & 42.2 & 40.7 & 40.2 & 27.2 \\
Qwen2.5-VL-32B~\cite{bai2025qwen25vl} & 32B & 54.8 & 58.8 & 52.1 & 48.2 & 37.7 & 53.9 & 45.9 & 50.2 & 48.7 & 28.3 \\
\famsep
Qwen3-VL-2B~\cite{qwen2025qwen3vl} & 2B & 38.0 & 30.8 & 34.7 & 35.2 & 32.6 & 46.8 & 37.4 & 36.5 & 34.6 & 21.0 \\
Qwen3-VL-4B~\cite{qwen2025qwen3vl} & 4B & 44.3 & 45.4 & 42.5 & 43.6 & 37.0 & 47.4 & 44.7 & 43.6 & 42.5 & 25.5 \\
Qwen3-VL-4B-Thinking~\cite{qwen2025qwen3vl} & 4B & 62.1 & 56.1 & 56.6 & 47.1 & 35.2 & 61.7 & 48.2 & 52.4 & 49.5 & 35.9 \\
Qwen3-VL-8B~\cite{qwen2025qwen3vl} & 8B & 50.9 & 46.8 & 44.8 & 47.0 & 40.0 & 54.5 & 47.7 & 47.4 & 45.5 & 26.9 \\
Qwen3-VL-30B-A3B~\cite{qwen2025qwen3vl} & 30B & 55.8 & 54.4 & 52.4 & 44.8 & 41.0 & 62.3 & 51.0 & 51.7 & 48.8 & 32.0 \\
Qwen3-VL-32B~\cite{qwen2025qwen3vl} & 32B & 62.1 & 56.8 & 66.0 & 50.8 & 46.0 & 64.3 & 54.1 & 57.2 & 55.7 & 35.8 \\
\famsep
Qwen3.5-9B~\cite{qwen2026qwen35} & 9B & 77.9 & 64.3 & 71.9 & 55.3 & 51.2 & 77.9 & 61.3 & 65.7 & 62.1 & 43.3 \\
Qwen3.6-27B~\cite{qwen2026qwen36} & 27B & 78.7 & 64.1 & 72.6 & 58.4 & 49.6 & 77.3 & 62.9 & 66.2 & 63.0 & 40.0 \\
\specialrule{0.5pt}{1pt}{1.2pt}\specialrule{0.5pt}{0pt}{0pt}
\rowcolor{secband}\multicolumn{12}{l}{\textbf{Post-trained models}} \\
\specialrule{0.5pt}{0pt}{1pt}
\multicolumn{2}{l}{\color{black!55}\textit{from} Idefics2-8B~\cite{laurencon2024idefics2}} & {\color{black!55}\underline{25.1}} & {\color{black!55}\underline{25.1}} & {\color{black!55}\textbf{31.6}} & {\color{black!55}\underline{27.3}} & {\color{black!55}\underline{25.0}} & {\color{black!55}\underline{33.1}} & {\color{black!55}\underline{18.1}} & {\color{black!55}\underline{26.5}} & {\color{black!55}\textbf{26.5}} & {\color{black!55}\textbf{11.5}} \\
Mantis-8B-Idefics2~\cite{jiang2024mantis}\m{\S} & 8B & \textbf{35.0} & \textbf{31.1} & \underline{13.6} & \textbf{30.6} & \textbf{26.4} & \textbf{35.1} & \textbf{19.0} & \textbf{27.3} & \underline{24.4} & \underline{10.0} \\
\famsep
\multicolumn{2}{l}{\color{black!55}\textit{from} Qwen2-VL-7B~\cite{wang2024qwen2vl}} & {\color{black!55}\underline{40.6}} & {\color{black!55}\underline{33.6}} & {\color{black!55}\textbf{36.9}} & {\color{black!55}\underline{36.9}} & {\color{black!55}\underline{34.2}} & {\color{black!55}\underline{40.3}} & {\color{black!55}\underline{38.9}} & {\color{black!55}\underline{37.3}} & {\color{black!55}\underline{36.5}} & {\color{black!55}\underline{18.7}} \\
Migician-7B~\cite{li2025migician}\m{\S} & 7B & \textbf{45.2} & \textbf{39.1} & \underline{35.7} & \textbf{38.2} & \textbf{36.2} & \textbf{47.4} & \textbf{40.6} & \textbf{40.3} & \textbf{38.1} & \textbf{27.5} \\
UniVG-R1-7B~\cite{bai2025univgr1}\m{\S} & 7B & 38.5 & 32.2 & 35.5 & 34.2 & 33.4 & 35.7 & 36.3 & 35.1 & 34.7 & 18.1 \\
\famsep
\multicolumn{2}{l}{\color{black!55}\textit{from} Qwen2.5-VL-3B~\cite{bai2025qwen25vl}} & {\color{black!55}\textbf{32.8}} & {\color{black!55}24.2} & {\color{black!55}\textbf{33.2}} & {\color{black!55}\textbf{29.7}} & {\color{black!55}\textbf{34.6}} & {\color{black!55}\textbf{44.2}} & {\color{black!55}\textbf{33.0}} & {\color{black!55}\textbf{33.1}} & {\color{black!55}\textbf{31.5}} & {\color{black!55}\textbf{10.5}} \\
HATCH-3B~\cite{oi2026hatch}\m{\S} & 3B & 28.2 & \underline{25.5} & \underline{31.0} & 27.3 & \underline{33.1} & \underline{41.6} & 21.1 & \underline{29.7} & \underline{28.5} & \underline{10.1} \\
MindCube-3B~\cite{wang2026mindcube}\m{\S} & 3B & \underline{29.0} & \textbf{25.9} & 28.8 & \underline{28.6} & 27.9 & 34.4 & \underline{27.4} & 28.9 & 28.1 & 9.6 \\
\famsep
\multicolumn{2}{l}{\color{black!55}\textit{from} Qwen2.5-VL-7B~\cite{bai2025qwen25vl}} & {\color{black!55}45.8} & {\color{black!55}38.3} & {\color{black!55}40.0} & {\color{black!55}42.0} & {\color{black!55}36.1} & {\color{black!55}40.3} & {\color{black!55}42.2} & {\color{black!55}40.7} & {\color{black!55}40.2} & {\color{black!55}27.2} \\
Visual-Jigsaw-7B~\cite{wu2026visualjigsaw}\m{\S} & 7B & 44.0 & 40.6 & 40.2 & 40.7 & 36.1 & 40.3 & 40.4 & 40.3 & 39.9 & 23.6 \\
VLAA-Thinker-7B~\cite{chen2025vlaathinker} & 7B & 44.5 & 39.0 & 40.3 & 42.7 & 36.6 & 40.3 & 41.9 & 40.8 & 40.5 & 25.7 \\
NoisyRollout-7B~\cite{liu2025noisyrollout} & 7B & 46.4 & 40.3 & 40.4 & 42.8 & 36.2 & 39.6 & 42.0 & 41.1 & 40.7 & 27.3 \\
VPPO-7B~\cite{huang2026vppo} & 7B & 47.1 & 41.3 & 41.9 & 42.5 & 36.3 & 39.0 & \underline{42.4} & 41.5 & 41.3 & 25.3 \\
R1-ShareVL-7B~\cite{yao2025r1sharevl} & 7B & \underline{47.5} & 42.6 & 42.3 & 42.9 & \underline{36.8} & 39.6 & 41.4 & 41.9 & 41.6 & 24.2 \\
PAPO-G-H-7B~\cite{wang2026papo} & 7B & 47.1 & 41.5 & 42.0 & \underline{43.2} & 36.3 & 40.9 & \textbf{42.7} & 42.0 & 41.5 & 25.3 \\
PeRL-7B~\cite{zhang2025perl}\m{\S} & 7B & 46.6 & \underline{44.0} & 41.7 & 40.8 & 36.8 & \underline{44.2} & 40.7 & 42.1 & 41.0 & 27.0 \\
PAPO-D-7B~\cite{wang2026papo} & 7B & 45.8 & 42.3 & \underline{43.1} & 42.6 & 36.4 & 42.9 & 42.0 & \underline{42.2} & \underline{41.7} & \underline{27.7} \\
\ours \cicaseven (ours) & 7B & \textbf{52.1} & \textbf{47.1} & \textbf{44.3} & \textbf{47.2} & \textbf{44.0} & \textbf{50.0} & 42.4 & \textbf{46.7} & \textbf{45.5} & \textbf{29.9} \\
\famsep
\multicolumn{2}{l}{\color{black!55}\textit{from} GLM-4.1V-9B-Thinking~\cite{hong2025glm41v}} & {\color{black!55}\underline{64.3}} & {\color{black!55}\underline{58.2}} & {\color{black!55}\underline{66.4}} & {\color{black!55}\underline{51.8}} & {\color{black!55}\underline{44.7}} & {\color{black!55}\underline{70.1}} & {\color{black!55}\underline{55.3}} & {\color{black!55}\underline{58.7}} & {\color{black!55}\underline{56.3}} & {\color{black!55}\underline{37.7}} \\
\ours \cicaglm (ours) & 9B & \textbf{69.3} & \textbf{61.1} & \textbf{67.1} & \textbf{52.3} & \textbf{46.4} & \textbf{77.3} & \textbf{55.4} & \textbf{61.3} & \textbf{57.6} & \textbf{40.2} \\
\famsep
\multicolumn{2}{l}{\color{black!55}\textit{from} Qwen3.5-9B~\cite{qwen2026qwen35}} & {\color{black!55}\underline{77.9}} & {\color{black!55}\underline{64.3}} & {\color{black!55}\underline{71.9}} & {\color{black!55}\underline{55.3}} & {\color{black!55}\underline{51.2}} & {\color{black!55}\underline{77.9}} & {\color{black!55}\underline{61.3}} & {\color{black!55}\underline{65.7}} & {\color{black!55}\underline{62.1}} & {\color{black!55}\underline{43.3}} \\
\ours \cicanine (ours) & 9B & \textbf{80.0} & \textbf{68.3} & \textbf{80.7} & \textbf{60.8} & \textbf{58.4} & \textbf{81.8} & \textbf{64.0} & \textbf{70.6} & \textbf{68.0} & \textbf{47.0} \\
\bottomrule
\end{tabular}
\end{table*}

\begin{table*}[t]
\centering\small
\setlength{\tabcolsep}{6pt}
\caption{\textbf{Gain over the base model.} \bench average (Avg) of open-source MLLMs post-trained from public base models and of those base models, for models trained for multi-image tasks (top), models post-trained for reasoning, video, spatial or perception tasks, most with reinforcement learning (middle), and \cica (bottom). The two models of a pair share hardware and runner, but pairs run separately, so the score of a base model can differ slightly between pairs. $\Delta$ is the difference before rounding, and $^{\dagger}$ marks a $\Delta$ whose 95\% interval excludes zero.}
\label{tab:supp-basedelta}
\begin{tabular}{llccr}
\toprule
Model & Base model & Base & Model & $\Delta$ \\
\midrule
Migician-7B~\cite{li2025migician} & Qwen2-VL-7B~\cite{wang2024qwen2vl} & 37.3 & 40.3 & +3.1\rlap{$^{\dagger}$} \\
UniVG-R1-7B~\cite{bai2025univgr1} & Qwen2-VL-7B & 37.3 & 35.1 & $-$2.1\rlap{$^{\dagger}$} \\
Visual-Jigsaw-7B~\cite{wu2026visualjigsaw} & Qwen2.5-VL-7B~\cite{bai2025qwen25vl} & 40.5 & 40.3 & $-$0.2 \\
PeRL-7B~\cite{zhang2025perl} & Qwen2.5-VL-7B & 40.5 & 42.1 & +1.6\rlap{$^{\dagger}$} \\
HATCH-3B~\cite{oi2026hatch} & Qwen2.5-VL-3B~\cite{bai2025qwen25vl} & 33.1 & 29.7 & $-$3.4\rlap{$^{\dagger}$} \\
MindCube-3B~\cite{wang2026mindcube} & Qwen2.5-VL-3B & 33.1 & 28.9 & $-$4.2\rlap{$^{\dagger}$} \\
Mantis-8B-Idefics2~\cite{jiang2024mantis} & Idefics2-8B~\cite{laurencon2024idefics2} & 26.5 & 27.3 & +0.8 \\
\midrule
Video-R1-7B~\cite{feng2025videor1} & Qwen2.5-VL-7B & 40.6 & 37.4 & $-$3.1\rlap{$^{\dagger}$} \\
VideoRFT-7B~\cite{wang2025videorft} & Qwen2.5-VL-7B & 40.6 & 38.2 & $-$2.3\rlap{$^{\dagger}$} \\
Open-o3-Video-7B~\cite{meng2026openo3video} & Qwen2.5-VL-7B & 40.6 & 39.8 & $-$0.8 \\
Spatial-SSRL-7B~\cite{liu2026spatialssrl} & Qwen2.5-VL-7B & 40.6 & 40.8 & +0.3 \\
Time-R1-7B~\cite{wang2025timer1} & Qwen2.5-VL-7B & 40.6 & 41.0 & +0.4 \\
ArrowRL-7B~\cite{xue2025arrowrl} & Qwen2.5-VL-7B & 40.6 & 41.2 & +0.6 \\
ThinkLite-VL-7B~\cite{wang2025thinklitevl} & Qwen2.5-VL-7B & 40.6 & 40.7 & +0.2 \\
NoisyRollout-7B~\cite{liu2025noisyrollout} & Qwen2.5-VL-7B & 40.6 & 41.1 & +0.5\rlap{$^{\dagger}$} \\
Vision-SR1-7B~\cite{li2026visionsr1} & Qwen2.5-VL-7B & 40.6 & 38.4 & $-$2.2\rlap{$^{\dagger}$} \\
Vision-Zero-7B~\cite{wang2026visionzero} & Qwen2.5-VL-7B & 40.5 & 40.4 & $-$0.2 \\
R1-ShareVL-7B~\cite{yao2025r1sharevl} & Qwen2.5-VL-7B & 40.5 & 41.9 & +1.3\rlap{$^{\dagger}$} \\
SophiaVL-R1-7B~\cite{fan2026sophiavlr1} & Qwen2.5-VL-7B & 40.5 & 40.9 & +0.4 \\
Perception-R1-7B~\cite{xiao2026perceptionr1} & Qwen2.5-VL-7B & 40.5 & 40.7 & +0.2 \\
VisionReasoner-7B~\cite{liu2026visionreasoner} & Qwen2.5-VL-7B & 40.5 & 40.6 & +0.1 \\
TON-7B-Math~\cite{wang2025ton} & Qwen2.5-VL-7B & 40.5 & 38.1 & $-$2.4\rlap{$^{\dagger}$} \\
SpatialReasoner-7B~\cite{ma2025spatialreasoner} & Qwen2.5-VL-7B & 40.5 & 34.4 & $-$6.2\rlap{$^{\dagger}$} \\
VLAA-Thinker-7B~\cite{chen2025vlaathinker} & Qwen2.5-VL-7B & 40.5 & 40.8 & +0.2 \\
SpatialLadder-3B~\cite{li2026spatialladder} & Qwen2.5-VL-3B & 33.2 & 29.1 & $-$4.1\rlap{$^{\dagger}$} \\
CapRL-3B~\cite{xing2026caprl} & Qwen2.5-VL-3B & 33.2 & 32.8 & $-$0.4 \\
Embodied-R1-3B~\cite{yuan2026embodiedr1} & Qwen2.5-VL-3B & 33.2 & 33.4 & +0.2 \\
GRIT-3B~\cite{fan2025grit} & Qwen2.5-VL-3B & 33.2 & 31.7 & $-$1.5\rlap{$^{\dagger}$} \\
Mulberry-7B~\cite{yao2025mulberry} & Qwen2-VL-7B & 37.3 & 34.0 & $-$3.3\rlap{$^{\dagger}$} \\
TVC-7B~\cite{sun2025tvc} & Qwen2-VL-7B & 37.3 & 31.8 & $-$5.4\rlap{$^{\dagger}$} \\
\midrule
\oursrow \cicaseven (ours) & Qwen2.5-VL-7B & 40.7 & 46.7 & +6.0\rlap{$^{\dagger}$} \\
\oursrow \cicanine (ours) & Qwen3.5-9B~\cite{qwen2026qwen35} & 65.7 & 70.6 & +4.9\rlap{$^{\dagger}$} \\
\oursrow \cicaglm (ours) & GLM-4.1V-9B-Thinking~\cite{hong2025glm41v} & 58.7 & 61.3 & +2.6\rlap{$^{\dagger}$} \\
\bottomrule
\end{tabular}
\end{table*}

\subsection{Ablations}\label{sec:supp-ablations}
The runs of \cref{tab:ablation} train Qwen2.5-VL-7B as in \cref{tab:main}, from its public checkpoint on all 49,105 items of \bench-Train.
They use the rewards, prompts, group size, batch size and learning-rate schedule of \cica on this backbone and, except in the GRPO row, its $\beta$ and $\tau$ (\cref{tab:supp-cica-defaults}). The GRPO ($\beta=0$) and \cica rows are the 7B models of \cref{tab:main}. All rows are evaluated with the same runner on the same hardware, on all of \bench and on the 1,000 items of \cref{fig:gain} under gray masking. \cref{tab:supp-betatau} varies $\beta$ and $\tau$.%

\section{Additional results}\label{sec:supp-results}
\subsection{Evaluating MLLMs}\label{sec:supp-res-mllm}
\cref{tab:supp-leaderboard} gives the accuracy per task category of most models in \cref{tab:leaderboard} and of further base and post-trained models.
Over the 58 open-source MLLMs, the order by the macro average has a Kendall correlation of 0.95 with the order by the micro average (Avg$^{\dagger}$) and of 0.96 with the order by the macro average without text/web (Spearman 0.994 and 0.996). Under all three averages, Qwen3.6-27B, Qwen3.5-9B and MiMo-VL-7B are the three leading models of the 58, and \cicaseven ranks above every other model post-trained from Qwen2.5-VL-7B.
\cref{tab:supp-basedelta} compares open-source MLLMs post-trained from public backbones with those backbones, both evaluated on the same hardware with the same runner.
\begin{table}[t]
\centering\footnotesize
\setlength{\tabcolsep}{2.85pt}
\caption{\textbf{Items that combine images are among the hardest.} For the label categories below (each with at least 200 items), the accuracy difference in points from each model's own average, averaged over 58 open-source MLLMs ($\Delta$), and the share of models whose 95\% interval lies below (Harder) or above (Easier) zero. The labels are those of \cref{sec:supp-groups}.}
\label{tab:supp-axes}
\begin{tabular}{@{}llrrrr@{}}
\toprule
Axis & Category & Items & $\Delta$ & Harder & Easier \\
\midrule
Cross-image operation & merge or compose & 434 & $-$15.5 & 95\% & 0\% \\
 & align or register & 731 & $-$11.7 & 95\% & 0\% \\
 & transfer or apply & 939 & $-$12.3 & 84\% & 10\% \\
 & order a sequence & 1,015 & $-$12.8 & 97\% & 0\% \\
 & find differences & 1,888 & $-$8.5 & 86\% & 3\% \\
 & compare & 3,498 & +6.2 & 0\% & 100\% \\
\midrule
Relation & spatial & 3,104 & $-$7.3 & 97\% & 0\% \\
 & logical or formal & 3,565 & $-$6.1 & 84\% & 9\% \\
 & semantic & 7,800 & +7.2 & 0\% & 100\% \\
\midrule
Operations per item & three or more & 1,756 & $-$3.8 & 71\% & 3\% \\
 & two & 8,386 & $-$0.7 & 38\% & 7\% \\
 & one & 8,471 & +1.5 & 5\% & 66\% \\
\midrule
Reasoning hops & four or more & 7,351 & $-$4.1 & 98\% & 0\% \\
 & two & 1,779 & +8.6 & 2\% & 90\% \\
\midrule
Time scale & minutes to hours & 648 & $-$16.7 & 98\% & 0\% \\
\bottomrule
\end{tabular}
\end{table}

\begin{figure}[t]
  \centering
  \includegraphics[width=\linewidth]{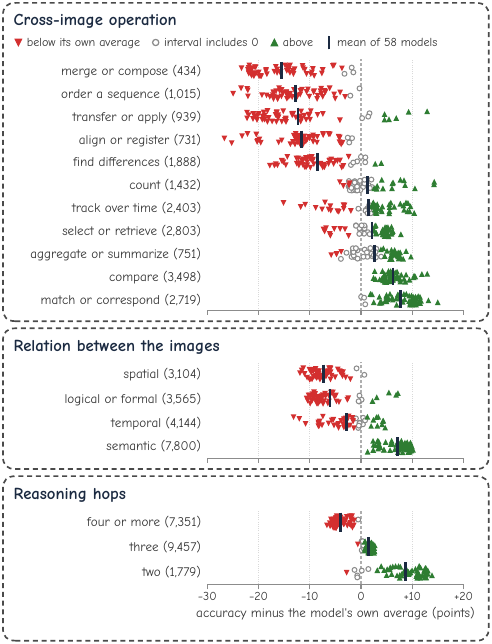}
  \caption{\textbf{Hardest item types for open-source MLLMs.} Accuracy on each label category minus the model's accuracy on all labeled \bench items, for the 58 open-source MLLMs of \cref{tab:supp-leaderboard,tab:supp-basedelta}, over the cross-image operations, image relations and reasoning hops with at least 200 items (item counts in parentheses). A downward red triangle marks a model whose 95\% interval lies below zero, an upward green triangle a model whose interval lies above zero, a gray circle a model whose interval includes zero, and a bar the mean over models.}
  \label{fig:supp-labels}
\end{figure}

\cref{tab:supp-axes,fig:supp-labels} break accuracy down by label category for 58 open-source MLLMs.

\subsection{One-image recovery and wrong answers}\label{sec:supp-res-rho}
\begin{figure*}[t]
  \centering
  \includegraphics[width=\linewidth]{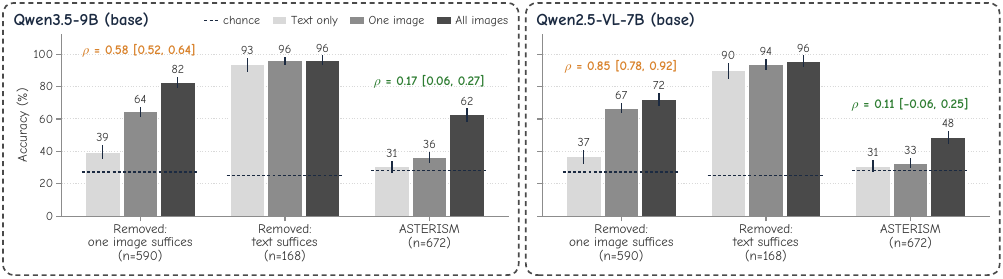}
  \caption{\textbf{Items removed by the audit are solvable from one image or from the text; \bench items are not.} Accuracy with all images replaced by gray (text only), with one image kept (mean over the kept image), and with all images, for Qwen3.5-9B and Qwen2.5-VL-7B on the multiple-choice items of \cref{fig:reliance}. Whiskers show 95\% intervals, and dashed lines the accuracy of a uniform guess. The recovery ratio $\rho$ is the share of the all-image gain that one image recovers on the items removed because one image suffices and on \bench items, with its 95\% interval, and \cref{tab:rhoK} gives $\rho$ over all removed items.}
  \label{fig:ladder}
\end{figure*}

\cref{tab:rhoK} shows that one image recovers more of the all-image gain on removed items than on \bench items in every image-count bin, after standardizing the image-count distribution, and without items whose answer is an image index.
\cref{fig:ladder} shows the text-only, one-image and all-image accuracies of Qwen3.5-9B and Qwen2.5-VL-7B on the items of \cref{fig:reliance}.
Among the 15 models whose all-image accuracy exceeds their text-only accuracy by at least 10 points on both item sets, so that the denominator of \cref{eq:rho} is large, the median $\rho$ is 0.77 on removed items and 0.18 on \bench items, and $\rho$ on \bench items stays between 0.09 and 0.31.
The 95\% interval of the gap between the two medians in \cref{sec:shortcut} comes from resampling the items and then the models, 1,000 times.

\paragraph{Single-image removals.}
Items that the audit removed as answerable from one image have a median $\rho$ of 0.80 across the 23 MLLMs of \cref{sec:shortcut}, against 0.18 for \bench items, and every model recovers more on the removed items.

\paragraph{The gap holds with images of other items.} \cref{tab:supp-swaprho} gives $\rho$ of each of the 23 MLLMs when images of other items replace the gray ones, next to its value under gray masking.
Every model recovers more on the removed items under both.

\begin{figure}[t]
  \centering
  \includegraphics[width=\linewidth]{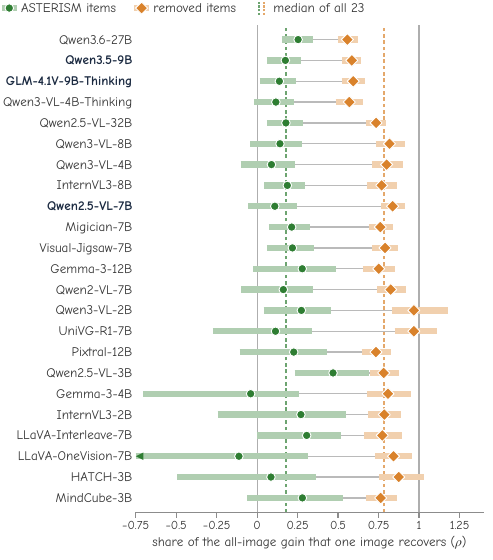}
  \caption{\textbf{One-image recovery $\rho$ of all 23 MLLMs.} Recovery ratio $\rho$ under gray masking on the items of \cref{fig:reliance} for each of the 23 open-source MLLMs of \cref{sec:shortcut}, ordered by \bench score (\cref{tab:supp-leaderboard}). Bands show 95\% intervals, dashed lines mark the median over the 23 models, and bold names mark the backbones that \cref{sec:cica-results} trains. An arrow marks an interval that continues past the axis.}
  \label{fig:supp-reliance-all}
\end{figure}

\paragraph{Wrong answers follow one image.} \cref{tab:supp-match34} gives the matched comparison of \cref{sec:shortcut} for each of the 31 open-source MLLMs, namely the 23 models of \cref{fig:supp-reliance-all} except MindCube-3B~\cite{wang2026mindcube}, plus nine further models.
On the \bench items of \cref{tab:match}, Match and Chance coincide on two-option items, and on the items with three or more options Match exceeds Chance by 30.0, 41.0 and 30.0 points for Qwen3.5-9B, Qwen2.5-VL-7B and GLM-4.1V-9B-Thinking.
\begin{table*}[t]
\centering\footnotesize
\setlength{\tabcolsep}{4pt}
\caption{\textbf{One-image recovery ratio $\rho$ by image count.} Removed items vs \bench items of \cref{fig:reliance} under gray masking, with 95\% intervals and, in the image-count rows, $n$ in parentheses (five- and six-image items are merged). The last rows cover all image counts (all), removed items reweighted to the \bench image-count distribution with five and six images as separate counts (K-standardized), and items whose answer is not an image index (no index answers). Negative values arise when one image with gray companions scores below the text-only input, and values above 1 when it scores above all images.}
\label{tab:rhoK}
\begin{tabular}{@{}lllll@{}}
\toprule
& \multicolumn{2}{c}{Qwen3.5-9B~\cite{qwen2026qwen35}} & \multicolumn{2}{c}{Qwen2.5-VL-7B~\cite{bai2025qwen25vl}} \\
\cmidrule(lr){2-3}\cmidrule(lr){4-5}
Images & removed & \bench & removed & \bench \\
\midrule
2 & 0.55 [0.46, 0.63] (298) & 0.38 [0.22, 0.52] (297) & 0.75 [0.64, 0.88] (298) & $-$0.08 [$-$0.70, 0.22] (297) \\
3 & 0.17 [$-$0.39, 0.45] (94) & $-$0.13 [$-$1.11, 0.26] (58) & 0.91 [0.73, 1.18] (94) & 0.22 [$-$0.36, 0.55] (58) \\
4 & 0.58 [0.25, 0.93] (83) & $-$0.04 [$-$1.22, 0.31] (107) & 0.87 [0.52, 1.34] (83) & 0.14 [$-$0.29, 0.40] (107) \\
5--6 & 0.69 [0.60, 0.79] (283) & 0.04 [$-$0.12, 0.16] (210) & 0.88 [0.78, 1.00] (283) & 0.21 [$-$0.06, 0.44] (210) \\
\midrule
all & 0.58 [0.52, 0.64] & 0.17 [0.06, 0.27] & 0.84 [0.77, 0.91] & 0.11 [$-$0.06, 0.25] \\
K-standardized & 0.64 & 0.17 & 0.88 & 0.11 \\
no index answers & 0.62 & 0.20 & 0.85 & 0.15 \\
\bottomrule
\end{tabular}
\end{table*}

\begin{table}[t]
\centering\footnotesize
\setlength{\tabcolsep}{4pt}
\caption{\textbf{The gap in one-image recovery holds for every MLLM when images of other items replace the gray ones.} One-image recovery ratio $\rho$ of the 23 open-source MLLMs of \cref{fig:supp-reliance-all} on the multiple-choice items of \cref{fig:reliance}, with the other images gray (Gray) or replaced by images of other items (Swapped).}
\label{tab:supp-swaprho}
\resizebox{\linewidth}{!}{%
\begin{tabular}{@{}lcccc@{}}
\toprule
 & \multicolumn{2}{c}{Gray} & \multicolumn{2}{c}{Swapped} \\
\cmidrule(lr){2-3}\cmidrule(lr){4-5}
Model & Removed & \bench & Removed & \bench \\
\midrule
Gemma-3-4B~\cite{gemma2025gemma3} & 0.81 & $-$0.04 & 0.73 & 0.07 \\
Gemma-3-12B~\cite{gemma2025gemma3} & 0.75 & 0.28 & 0.58 & 0.14 \\
GLM-4.1V-9B-Thinking~\cite{hong2025glm41v} & 0.59 & 0.14 & 0.53 & 0.12 \\
HATCH-3B~\cite{oi2026hatch} & 0.87 & 0.09 & 0.82 & $-$0.06 \\
InternVL3-2B~\cite{zhu2025internvl3} & 0.79 & 0.27 & 0.67 & 0.36 \\
InternVL3-8B~\cite{zhu2025internvl3} & 0.77 & 0.19 & 0.56 & 0.17 \\
LLaVA-Interleave-7B~\cite{li2025llavainterleave} & 0.77 & 0.31 & 0.62 & 0.05 \\
LLaVA-OneVision-7B~\cite{li2025llavaonevision} & 0.84 & $-$0.11 & 0.64 & 0.15 \\
Migician-7B~\cite{li2025migician} & 0.76 & 0.21 & 0.65 & 0.19 \\
MindCube-3B~\cite{wang2026mindcube} & 0.76 & 0.28 & 0.67 & 0.30 \\
Pixtral-12B~\cite{agrawal2024pixtral} & 0.73 & 0.23 & 0.63 & 0.24 \\
Qwen2-VL-7B~\cite{wang2024qwen2vl} & 0.82 & 0.16 & 0.63 & 0.11 \\
Qwen2.5-VL-3B~\cite{bai2025qwen25vl} & 0.78 & 0.47 & 0.68 & $-$0.06 \\
Qwen2.5-VL-7B~\cite{bai2025qwen25vl} & 0.84 & 0.11 & 0.68 & 0.10 \\
Qwen2.5-VL-32B~\cite{bai2025qwen25vl} & 0.73 & 0.18 & 0.64 & 0.11 \\
Qwen3-VL-2B~\cite{qwen2025qwen3vl} & 0.97 & 0.27 & 0.77 & 0.06 \\
Qwen3-VL-4B~\cite{qwen2025qwen3vl} & 0.80 & 0.09 & 0.70 & 0.15 \\
Qwen3-VL-4B-Thinking~\cite{qwen2025qwen3vl} & 0.57 & 0.12 & 0.55 & 0.11 \\
Qwen3-VL-8B~\cite{qwen2025qwen3vl} & 0.82 & 0.14 & 0.74 & 0.16 \\
Qwen3.5-9B~\cite{qwen2026qwen35} & 0.58 & 0.17 & 0.53 & 0.09 \\
Qwen3.6-27B~\cite{qwen2026qwen36} & 0.56 & 0.25 & 0.48 & 0.19 \\
UniVG-R1-7B~\cite{bai2025univgr1} & 0.97 & 0.11 & 0.72 & 0.16 \\
Visual-Jigsaw-7B~\cite{wu2026visualjigsaw} & 0.79 & 0.22 & 0.71 & 0.17 \\
\midrule
Median & 0.78 & 0.18 & 0.65 & 0.14 \\
\bottomrule
\end{tabular}}
\end{table}

\begin{table}[t]
\centering\footnotesize
\setlength{\tabcolsep}{3pt}
\caption{\textbf{Wrong answers repeat the model's own one-image answers.} For 31 open-source MLLMs on multiple-choice \bench items with two to six images, the wrong answers that pick an option and have a matched item ($n$), the share that repeat one of the model's one-image answers (Match), the share that repeat the one-image answers of another item with the same number of options, the same correct letter and the same number of images (Matched chance), their difference with its 95\% interval (Diff.), and, averaged over the other models, the share that repeat those models' one-image answers (Others). Models with $n<100$ are not listed.}
\label{tab:supp-match34}
\resizebox{\linewidth}{!}{%
\begin{tabular}{lrrrlr}
\toprule
Model & $n$ & Match & Matched chance & Diff. & Others \\
\midrule
Gemma-3-4B~\cite{gemma2025gemma3} & 412 & 90.5 & 47.6 & +42.9 [39.2, 46.5] & 54.2 \\
Gemma-3-12B~\cite{gemma2025gemma3} & 399 & 83.5 & 48.1 & +35.4 [31.3, 39.2] & 50.3 \\
Gemma-4-12B~\cite{gemma2026gemma4} & 336 & 74.4 & 46.5 & +27.9 [23.3, 32.7] & 56.8 \\
GLM-4.1V-9B-Thinking~\cite{hong2025glm41v} & 262 & 80.9 & 53.3 & +27.6 [22.6, 32.3] & 53.2 \\
HATCH-3B~\cite{oi2026hatch} & 386 & 86.3 & 41.0 & +45.3 [41.3, 48.9] & 54.5 \\
InternVL3-2B~\cite{zhu2025internvl3} & 160 & 83.8 & 27.9 & +55.9 [49.3, 62.1] & 51.0 \\
InternVL3-8B~\cite{zhu2025internvl3} & 320 & 84.1 & 47.2 & +36.9 [32.8, 40.9] & 52.5 \\
LLaVA-Interleave-7B~\cite{li2025llavainterleave} & 361 & 82.5 & 42.0 & +40.5 [36.2, 44.6] & 47.6 \\
LLaVA-OneVision-7B~\cite{li2025llavaonevision} & 393 & 87.8 & 43.7 & +44.1 [40.4, 47.5] & 56.8 \\
LLaVA-OneVision-2-8B~\cite{an2026llavaonevision2} & 342 & 79.2 & 42.4 & +36.8 [31.9, 41.6] & 55.2 \\
Mantis-8B-Idefics2~\cite{jiang2024mantis} & 320 & 80.6 & 34.8 & +45.8 [41.3, 50.4] & 54.3 \\
Migician-7B~\cite{li2025migician} & 259 & 86.5 & 39.6 & +46.9 [42.2, 51.6] & 59.1 \\
MiMo-VL-7B~\cite{xiaomi2025mimovl} & 269 & 74.0 & 47.9 & +26.1 [20.8, 31.5] & 51.3 \\
mPLUG-Owl3-7B~\cite{ye2025mplugowl3} & 316 & 89.9 & 33.4 & +56.5 [52.4, 60.2] & 56.9 \\
NoisyRollout-7B~\cite{liu2025noisyrollout} & 316 & 82.6 & 50.4 & +32.2 [27.8, 36.6] & 57.1 \\
Pixtral-12B~\cite{agrawal2024pixtral} & 396 & 88.6 & 46.2 & +42.4 [38.8, 45.8] & 56.1 \\
Qwen2-VL-7B~\cite{wang2024qwen2vl} & 187 & 81.8 & 28.8 & +53.0 [48.0, 58.0] & 60.6 \\
Qwen2.5-VL-3B~\cite{bai2025qwen25vl} & 303 & 80.5 & 36.3 & +44.2 [39.7, 48.8] & 57.2 \\
Qwen2.5-VL-7B~\cite{bai2025qwen25vl} & 321 & 84.4 & 49.5 & +34.9 [30.8, 39.1] & 57.9 \\
Qwen2.5-VL-32B~\cite{bai2025qwen25vl} & 267 & 75.7 & 47.0 & +28.7 [22.7, 34.4] & 50.3 \\
Qwen3-VL-2B~\cite{qwen2025qwen3vl} & 330 & 84.5 & 49.1 & +35.4 [31.0, 39.3] & 55.3 \\
Qwen3-VL-4B~\cite{qwen2025qwen3vl} & 291 & 84.5 & 50.5 & +34.0 [29.5, 38.5] & 56.6 \\
Qwen3-VL-4B-Thinking~\cite{qwen2025qwen3vl} & 199 & 67.8 & 43.1 & +24.7 [18.2, 31.4] & 54.5 \\
Qwen3-VL-8B~\cite{qwen2025qwen3vl} & 311 & 83.6 & 50.2 & +33.4 [28.8, 37.6] & 57.4 \\
Qwen3-VL-30B-A3B~\cite{qwen2025qwen3vl} & 307 & 79.8 & 46.7 & +33.1 [27.9, 38.2] & 53.0 \\
Qwen3-VL-32B~\cite{qwen2025qwen3vl} & 274 & 75.5 & 47.8 & +27.7 [22.4, 32.9] & 53.8 \\
Qwen3.5-9B~\cite{qwen2026qwen35} & 205 & 75.1 & 47.4 & +27.7 [21.5, 33.6] & 47.9 \\
Qwen3.6-27B~\cite{qwen2026qwen36} & 229 & 69.4 & 47.0 & +22.4 [16.4, 28.6] & 55.2 \\
R1-ShareVL-7B~\cite{yao2025r1sharevl} & 306 & 81.0 & 48.2 & +32.8 [28.4, 37.2] & 54.4 \\
UniVG-R1-7B~\cite{bai2025univgr1} & 100 & 79.0 & 19.1 & +59.9 [51.2, 68.7] & 55.6 \\
Visual-Jigsaw-7B~\cite{wu2026visualjigsaw} & 316 & 81.6 & 54.4 & +27.2 [22.7, 31.7] & 55.6 \\
\bottomrule
\end{tabular}}
\end{table}

\begin{table}[t]
\centering\footnotesize
\setlength{\tabcolsep}{2.6pt}
\caption{\textbf{The single-image inputs of \cica also show the gap.} One-image recovery ratio $\rho$ when the other images are hidden inside the model as \cica does (\cref{sec:cica-divergence}). Accuracy is the exact probability of the correct option among the valid answers of the items with a fixed set of options (775 \bench and 796 removed items, which add the yes/no and true/false items to the multiple-choice items of \cref{fig:reliance}).}
\label{tab:supp-mask}
\begin{tabular}{@{}lccc@{}}
\toprule
Model & \bench items & Removed items & Difference \\
\midrule
Qwen2.5-VL-7B~\cite{bai2025qwen25vl} & 0.22 & 0.74 & +0.52 \\
Qwen3.5-9B~\cite{qwen2026qwen35} & 0.11 & 0.66 & +0.55 \\
\bottomrule
\end{tabular}
\end{table}

\subsection{Training with \cica}\label{sec:supp-res-cica}
\cref{tab:supp-leaderboard} gives the accuracy per task category of each backbone and of the same backbone trained with \cica.
On \bench, the paired bootstrap~\cite{efron1979bootstrap} 95\% interval over items excludes zero for the difference between \cica and each other row of \cref{tab:main}.

\cref{tab:rungs} gives the differences per masking condition behind \cref{fig:gain} for the three backbones against the base model and the five RL baselines, and the error bars of \cref{fig:gain} show their paired bootstrap 95\% intervals over items.
On Qwen2.5-VL-7B against the base model, GRPO and DAPO, \cref{tab:supp-maskswap} gives these differences with the other images hidden inside the model as \cica does (a) or replaced by images from other items (b).

\begin{table*}[t]
\centering\footnotesize
\setlength{\tabcolsep}{5pt}
\caption{\textbf{\cica's gains by masking condition.} Accuracy difference (\cica minus baseline, points) on the \bench items of \cref{fig:gain}, and the difference between the all-image and one-image gains before rounding.}
\label{tab:rungs}
\begin{tabular}{@{}llrrrr@{}}
\toprule
Backbone & vs. & text only & one image & all images & all $-$ one \\
\midrule
Qwen2.5-VL-7B~\cite{bai2025qwen25vl} & base & +0.5 & +1.1 & +6.5 & +5.5 \\
 & GRPO & $-$0.9 & +0.0 & +5.0 & +5.0 \\
 & DAPO & +0.1 & +1.0 & +4.2 & +3.2 \\
 & APPO & +0.3 & +0.9 & +3.5 & +2.6 \\
 & VGPO & $-$0.1 & +1.0 & +4.4 & +3.4 \\
 & CFPO & $-$1.3 & $-$0.9 & +3.5 & +4.4 \\
\midrule
GLM-4.1V-9B-Thinking~\cite{hong2025glm41v} & base & +1.1 & +1.6 & +2.9 & +1.3 \\
 & GRPO & +2.0 & +1.6 & +3.7 & +2.1 \\
 & DAPO & +0.9 & +1.8 & +3.7 & +1.9 \\
 & APPO & +2.0 & +0.3 & +2.9 & +2.6 \\
 & VGPO & +1.1 & $-$0.5 & +2.9 & +3.4 \\
 & CFPO & +1.1 & +0.6 & +2.5 & +1.9 \\
\midrule
Qwen3.5-9B~\cite{qwen2026qwen35} & base & $-$1.0 & +0.9 & +5.7 & +4.8 \\
 & GRPO & $-$0.7 & +0.6 & +3.7 & +3.1 \\
 & DAPO & +1.6 & +2.3 & +4.7 & +2.4 \\
 & APPO & $-$0.5 & +2.4 & +4.6 & +2.2 \\
 & VGPO & +0.5 & +1.1 & +4.4 & +3.3 \\
 & CFPO & +1.4 & +4.3 & +8.1 & +3.8 \\
\bottomrule
\end{tabular}
\end{table*}

\begin{table*}[t]
\centering\footnotesize
\setlength{\tabcolsep}{5pt}
\caption{\textbf{\cica's gains by condition when the other images are hidden inside the model (a) or replaced with images from other items (b).} (a) As \cref{tab:rungs}, but the text-only and one-image conditions hide images inside the model as \cica does (\cref{sec:cica-divergence}) instead of graying them out, and accuracy is the exact probability of the correct option among the valid answers of the 775 \bench items of \cref{tab:supp-mask}. (b) As \cref{tab:rungs}, but the first condition (no true image) replaces every image with an image from another item and the second (one true image) keeps one true image and replaces the others (average over the kept image), on another 1,000 \bench items. $^{\dagger}$ marks a difference whose 95\% interval excludes zero.}
\label{tab:supp-maskswap}%
\makeatletter
\let\@currentcounter\@empty\protected@edef\@currentlabel{\thetable a}\protected@edef\cref@currentlabel{[table][\arabic{table}][]\thetable a}\label{tab:supp-maskgain}%
\let\@currentcounter\@empty\protected@edef\@currentlabel{\thetable b}\protected@edef\cref@currentlabel{[table][\arabic{table}][]\thetable b}\label{tab:supp-swapgain}%
\makeatother
\begin{tabular}{@{}llrrrr@{\phantom{$^{\dagger}$}\hspace{2\tabcolsep}}rrrr@{\phantom{$^{\dagger}$}}}
\toprule
 & & \multicolumn{4}{c}{(a) Hidden inside the model} & \multicolumn{4}{c}{(b) Images from other items} \\
\cmidrule(lr){3-6}\cmidrule(l){7-10}
Backbone & vs. & text only & one image & all images & all $-$ one & no true image & one true image & all images & all $-$ one \\
\midrule
Qwen2.5-VL-7B~\cite{bai2025qwen25vl} & base & +0.6 & +0.6 & +8.3\rlap{$^{\dagger}$} & +7.7\rlap{$^{\dagger}$} & $-$1.2 & +1.6 & +6.7\rlap{$^{\dagger}$} & +5.1\rlap{$^{\dagger}$} \\
 & GRPO & +0.5 & +0.3 & +5.7\rlap{$^{\dagger}$} & +5.4\rlap{$^{\dagger}$} & $-$0.5 & 0.0 & +5.9\rlap{$^{\dagger}$} & +5.9\rlap{$^{\dagger}$} \\
 & DAPO & +0.2 & $-$0.2 & +4.0\rlap{$^{\dagger}$} & +4.3\rlap{$^{\dagger}$} & $-$1.5 & +0.6 & +3.5\rlap{$^{\dagger}$} & +2.9\rlap{$^{\dagger}$} \\
\bottomrule
\end{tabular}
\end{table*}

\subsection{Single-image benchmarks}\label{sec:supp-res-single}
On ten single-image benchmarks for hallucination, real-world images, high-resolution images, and OCR and diagrams, \cref{tab:supp-single} compares each backbone with the same backbone trained with \cica.
\cica raises the average score over the ten benchmarks on each backbone (Avg).

\begin{table*}[t]
\centering\footnotesize
\setlength{\tabcolsep}{4pt}
\caption{\textbf{Single-image benchmarks.} Scores on ten single-image benchmarks of each backbone and of the same backbone trained with \cica. The benchmarks are grouped into hallucination (HallusionBench~\cite{guan2024hallusionbench}, POPE~\cite{li2023pope}), real-world images (RealWorldQA~\cite{xai2024grok15v}, MME-RealWorld-Lite~\cite{zhang2025mmerealworld}, R-Bench-Dis~\cite{li2025rbench}), high-resolution images (V$^{*}$Bench~\cite{wu2024vstar}, HR-Bench 4K and 8K~\cite{wang2025hrbench}) and OCR and diagrams (OCRBench~\cite{liu2024ocrbench}, AI2D~\cite{kembhavi2016ai2d}) and are evaluated with VLMEvalKit~\cite{duan2024vlmevalkit} and its default answer extraction. POPE reports F1. R-Bench-Dis applies real-world corruptions to the images. Avg is the mean of the ten columns, and the higher value of each pair is in bold.}
\label{tab:supp-single}
\begin{tabular}{lrrrrrrrrrr>{\hspace{4pt}}r}
\toprule
 & \multicolumn{2}{c}{Hallucination} & \multicolumn{3}{c}{Real-world} & \multicolumn{3}{c}{High-resolution} & \multicolumn{2}{c}{OCR/diagram} & \\
\cmidrule(lr){2-3}\cmidrule(lr){4-6}\cmidrule(lr){7-9}\cmidrule(lr){10-11}
Model & \begin{tabular}[b]{@{}r@{}}Hallusion\\Bench\end{tabular} & POPE & \begin{tabular}[b]{@{}r@{}}RealWorld\\QA\end{tabular} & \begin{tabular}[b]{@{}r@{}}MME-RW\\Lite\end{tabular} & \begin{tabular}[b]{@{}r@{}}R-Bench\\Dis\end{tabular} & \begin{tabular}[b]{@{}r@{}}V$^{*}$\\Bench\end{tabular} & \begin{tabular}[b]{@{}r@{}}HR-Bench\\4K\end{tabular} & \begin{tabular}[b]{@{}r@{}}HR-Bench\\8K\end{tabular} & \begin{tabular}[b]{@{}r@{}}OCR\\Bench\end{tabular} & AI2D & Avg \\
\midrule
Qwen2.5-VL-7B~\cite{bai2025qwen25vl} & 50.1 & 86.5 & 69.4 & 44.5 & 65.3 & 75.9 & 68.8 & 65.0 & 88.1 & 83.7 & 69.7 \\
\oursrow\sub \cica (ours) & \textbf{51.0} & \textbf{89.2} & \textbf{71.1} & \textbf{48.7} & \textbf{68.5} & \textbf{77.0} & \textbf{72.6} & \textbf{69.5} & \textbf{88.3} & \textbf{83.9} & \textbf{72.0} \\
\midrule
GLM-4.1V-9B-Thinking~\cite{hong2025glm41v} & \textbf{63.1} & 87.2 & 72.5 & 46.1 & 69.9 & \textbf{78.0} & 74.1 & 66.4 & 84.9 & 87.5 & 73.0 \\
\oursrow\sub \cica (ours) & \textbf{63.1} & \textbf{88.4} & \textbf{75.2} & \textbf{47.6} & \textbf{72.7} & \textbf{78.0} & \textbf{75.0} & \textbf{66.5} & \textbf{85.7} & \textbf{87.8} & \textbf{74.0} \\
\midrule
Qwen3.5-9B~\cite{qwen2026qwen35} & 65.7 & 88.3 & 81.3 & 59.5 & 75.4 & 88.5 & 88.8 & 83.4 & 89.4 & 87.7 & 80.8 \\
\oursrow\sub \cica (ours) & \textbf{67.3} & \textbf{88.5} & \textbf{81.7} & \textbf{60.8} & \textbf{76.6} & \textbf{89.5} & \textbf{90.0} & \textbf{85.7} & \textbf{90.1} & \textbf{90.5} & \textbf{82.1} \\
\bottomrule
\end{tabular}
\end{table*}

\subsection{MMIU}\label{sec:supp-res-multi}
\cref{tab:supp-mmiu} gives the MMIU~\cite{meng2025mmiu} scores of Qwen2.5-VL-7B and Qwen3.5-9B, the five RL baselines of \cref{tab:main} and both backbones trained with \cica.
MMIU builds 52 multi-image tasks from existing datasets and releases them as 60 task splits.
On both backbones, \cica scores above the base model and all five RL baselines.

\begin{table}[t]
\centering\footnotesize
\setlength{\tabcolsep}{3pt}
\caption{\textbf{MMIU.} MMIU~\cite{meng2025mmiu} accuracy of Qwen2.5-VL-7B and Qwen3.5-9B, the five RL baselines of \cref{tab:main} and each backbone trained with \cica. Evaluation uses the official prompt and answer extraction of MMIU and the settings of \cref{sec:supp-eval}, and the score averages the 60 task splits of the official release as its evaluation script does.}
\label{tab:supp-mmiu}
\begin{adjustbox}{max width=\columnwidth}
\begin{tabular}{lrrrrrr>{\columncolor{oursgreen}}c}
\toprule
Backbone & Base & GRPO & DAPO & APPO & VGPO & CFPO & \cica (ours) \\
\midrule
Qwen2.5-VL-7B~\cite{bai2025qwen25vl} & 51.9 & 53.0 & 52.9 & 53.0 & 52.8 & 52.7 & \textbf{54.4} \\
Qwen3.5-9B~\cite{qwen2026qwen35} & 64.5 & 65.2 & 63.0 & 64.5 & 66.1 & 60.8 & \textbf{68.5} \\
\bottomrule
\end{tabular}
\end{adjustbox}
\end{table}

\subsection{Shortcut after training}\label{sec:supp-res-shortcut}
\begin{figure}[t]
  \centering
  \includegraphics[width=\linewidth]{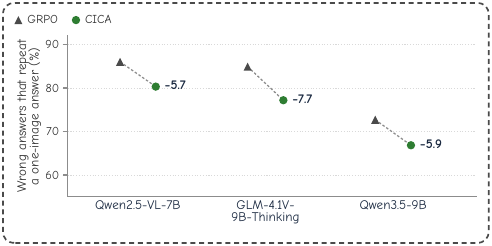}
  \caption{\textbf{After training with \cica, fewer wrong answers repeat a one-image answer than after GRPO.} Share of the wrong all-image answers that equal an answer the same model gives with one image visible and the other images gray, for each backbone trained with GRPO or with \cica. Items are the multiple-choice items among 1,000 \bench items on which the base model and both trained models give an answer in every masking condition. Numbers give the change of \cica from GRPO in points.}
  \label{fig:supp-shortcut}
\end{figure}

\cref{fig:supp-shortcut} repeats the wrong-answer test of \cref{sec:shortcut} for each backbone trained with GRPO or with \cica, under the gray masking of \cref{fig:gain}. After training with \cica, the share of wrong answers that repeat a one-image answer lies below the share after GRPO on each backbone, by up to 7.7 points on GLM-4.1V-9B-Thinking~\cite{hong2025glm41v}.

\subsection{Reliability of the gains}\label{sec:supp-res-reliability}
With 18,696 items, the standard error of a model's \bench score is at most 0.7 points.
Pairing the answers of two models on the same items lowers the standard error of their difference to 0.4--0.6 points for \cica against the base model and against the strongest RL baseline of \cref{tab:main}.
On each backbone, the paired bootstrap 95\% intervals of both differences, with items resampled within each task category (2,000 resamples), lie at least 1.3 points above zero.
The micro average (Avg$^{\dagger}$ in \cref{tab:supp-leaderboard}), which weights all items equally, and the macro average without text/web, the smallest category, also put \cica above the base model and the strongest RL baseline on each backbone.
Without the 1,406 rewritten items, \cica scores 2.8 to 5.8 points above the base model and 2.3 to 3.6 points above the strongest RL baseline on each of Qwen2.5-VL-7B, GLM-4.1V-9B-Thinking and Qwen3.5-9B, again with paired intervals above zero.
Averaged over the six task categories other than retrieval, \cica also scores above the base model and all five RL baselines on each backbone.

\section{Limitations and future work}\label{sec:supp-limits}
\bench takes its images and most of its questions from the 19 sources, so its domains and task mix follow theirs.
Its 1,406 rewrites, new human-checked questions on those images, would alone make a benchmark larger than most of the sources (\cref{tab:supp-ledger}).
Later versions can add sources from other domains.

The \bench score macro-averages the seven task categories so that the largest do not dominate it (\cref{sec:scorer-bench}), which gives text/web, the smallest, equal weight.
We also report the micro average, which weights all items equally (\cref{tab:supp-leaderboard}).
Under that average, or without text/web, the open-source MLLMs keep nearly the same order, and \cica stays above the base model and the strongest RL baseline on each backbone (\cref{sec:supp-res-mllm,sec:supp-res-reliability}).
Source identifiers also let users drop or reweight any source (\cref{sec:supp-data-use}).

For the audit of over 32,000 source items (\cref{tab:supp-ledger}), Claude Fable~5 and GPT-5.6-Sol, two frontier MLLMs from different model families and among the strongest at the time, give the verdicts, and a human adjudicator settles their disagreements and escalated verdicts (\cref{sec:supp-audit}).
The verdicts on items that no human reviewed thus rest on these models, and on a stratified sample a reviewer agrees with the audit as often as with a second reviewer, while humans check every corrected label and rewrite (\cref{sec:supp-human}).
The single-image filter likewise relies on a fixed committee of four open-source MLLMs, and for 23 MLLMs outside it one image recovers far less of the all-image gain on \bench items than on removed items (\cref{sec:shortcut}).

We train \cica on three backbones, with one training run per model in \cref{tab:main,tab:ablation}.
On each backbone, \cica scores above every RL baseline by more than the spread of the RL baselines, and its gain over each is larger with all images visible than with one (\cref{tab:main,tab:rungs}).
On Qwen2.5-VL-7B, all six \cica runs of \cref{tab:ablation,tab:supp-betatau} that keep $\sigma$ tied to the images score above every RL baseline.
\cica also scores above every RL baseline on MMIU and keeps or raises every single-image score (\cref{tab:supp-mmiu,tab:supp-single}).
The five RL baselines train on the same \bench-Train items as \cica, cleared of near-duplicates of \bench images (\cref{sec:supp-train-data,sec:supp-baselines}).
The three backbones differ in architecture and pre-training (\cref{sec:supp-backbones}), and two of them were already post-trained with RL~\cite{qwen2026qwen35,hong2025glm41v}.
In particular, GLM-4.1V-9B-Thinking went through large-scale RL from STEM problems to GUI agents~\cite{hong2025glm41v}, and the RL baselines move its score less than on the other two backbones (\cref{tab:main}).
With limited compute, we leave testing \cica on larger MLLMs to future work.

\cica moves credit to the tokens that need more than one image and does not target retrieval items, in which finding the one image or page that holds the answer is the task.
It raises retrieval accuracy on all three backbones, by 2.7 points on Qwen3.5-9B and slightly on the other two (\cref{tab:supp-leaderboard}), and crediting the search itself remains open.

\section{Data card}\label{sec:supp-data}
\subsection{Licenses and release}\label{sec:supp-data-license}
\bench keeps the license of each source benchmark (\cref{tab:supp-license}).
For the sources whose license allows redistribution and commercial use, we release the items with their images, together with our corrected labels and answer formats, the rewritten items and the per-item labels.
For the sources with non-commercial terms or without a stated license, we release the item identifiers, our corrections and rewritten questions and options, and build scripts that assemble the items from the public release of each source.
Where the dataset card and the repository or paper of a source state different licenses, we follow the more restrictive one. Together with \bench, we release the weights of \cicaseven, \cicanine and \cicaglm with their configuration, tokenizer and processor files.
\begin{table}[t]
\centering\footnotesize
\setlength{\tabcolsep}{3pt}
\caption{\textbf{Licenses of the source benchmarks.} The license that each source states on its dataset card or in its repository, and what \bench releases for it, either the items with their images (items) or the item identifiers with a build script that assembles the items from the public release (identifiers). $^{a}$The dataset card and the repository or paper state different licenses, and the more restrictive one is listed. $^{b}$Named on the dataset card, whose license file is empty. Images keep the licenses of the datasets they come from.}
\label{tab:supp-license}
\begin{tabular}{@{}l>{\raggedright\arraybackslash}p{0.46\linewidth}l@{}}
\toprule
Source & License & Release \\
\midrule
MileBench~\cite{song2024milebench} & CC BY-NC 4.0$^{a}$ & identifiers \\
MMRB~\cite{cheng2025mmrb} & CC BY 4.0 & items \\
VLM2-Bench~\cite{zhang2025vlm2bench} & CC BY-NC 4.0$^{a}$ & identifiers \\
MMR-Life~\cite{li2026mmrlife} & CC BY 4.0 & items \\
ReMI~\cite{kazemi2024remi} & CC BY 4.0$^{b}$ & items \\
MuirBench~\cite{wang2025muirbench} & CC BY 4.0 & items \\
MRAG-Bench~\cite{hu2025mragbench} & CC BY 4.0 & items \\
M4Bench~\cite{ye2025m4bench} & not stated & identifiers \\
MV-MATH~\cite{wang2025mvmath} & MIT & items \\
BLINK~\cite{fu2024blink} & Apache 2.0 & items \\
MMLongBench-Doc~\cite{ma2024mmlongbenchdoc} & CC BY-NC 4.0 & identifiers \\
DUDE~\cite{vanlandeghem2023dude} & CC BY 4.0 & items \\
MaRVL~\cite{liu2021marvl} & CC BY 4.0 (text), CC BY-NC-SA 4.0 (images) & identifiers \\
MMSI-Bench~\cite{yang2026mmsibench} & CC BY 4.0 & items \\
Q-Bench2~\cite{zhang2024qbenchplus} & S-Lab License 1.0 (non-commercial)$^{a}$ & identifiers \\
MIRB~\cite{zhao2024mirb} & MIT & items \\
DiCoBench~\cite{li2026dicobench} & MIT & items \\
EMMA~\cite{hao2025emma} & not stated & identifiers \\
Mantis-Eval~\cite{jiang2024mantis} & Apache 2.0 & items \\
\bottomrule
\end{tabular}
\end{table}

\subsection{Composition and intended use}\label{sec:supp-data-use}
\bench holds 18,696 items from 19 public benchmarks, grouped into seven task categories (\cref{tab:stats}), and every item has two or more images.
\begin{table}[t]
\centering\footnotesize
\setlength{\tabcolsep}{2.6pt}
\caption{\textbf{\bench statistics.} Items, source benchmarks (Src), mean and maximum images per item, the percentage of items whose answer is an option letter (Opt), yes/no or true/false (Y/N), a number (Num) or a short text (Str), and source items whose label, answer format, question or options differ from the source release beyond option order (Fix).}
\label{tab:stats}
\begin{tabular}{@{}lrrrrrrrrr@{}}
\toprule
Category & Items & Src & \multicolumn{2}{c}{Images} & Opt & Y/N & Num & Str & Fix \\
\cmidrule(lr){4-5}
& & & mean & max & & & & & \\
\midrule
Aggregate & 1,055 & 13 & 3.4 & 15 & 56 & 0 & 37 & 7 & 19 \\
Compare/change & 2,262 & 14 & 2.9 & 35 & 99 & 1 & 0 & 0 & 106 \\
Reasoning/verif. & 4,793 & 13 & 3.0 & 35 & 35 & 36 & 22 & 7 & 86 \\
Spatial corresp. & 4,618 & 13 & 3.7 & 35 & 96 & 0 & 0 & 4 & 91 \\
Temporal/action & 3,463 & 9 & 12.6 & 109 & 99 & 0 & 0 & 1 & 172 \\
Text/web & 154 & 7 & 3.4 & 10 & 100 & 0 & 0 & 0 & 8 \\
\midrule
Retrieval & 2,351 & 8 & 19.1 & 64 & 39 & 2 & 17 & 42 & 87 \\
\midrule
\bench & 18,696 & 19 & 7.0 & 109 & 72 & 10 & 10 & 9 & 569 \\
\bottomrule
\end{tabular}
\end{table}

Source items enter only if they pass a two-stage single-image filter and an audit by two MLLMs with a human adjudicator.
The audit also corrects labels, answer formats and text.
Rewrites keep the images of their source items and enter only if no single image answers them, and a human reviewer checks every corrected label and every rewrite (\cref{sec:supp-audit,sec:supp-ledger,sec:supp-human}).
Each item, as released or as assembled by the build scripts, records its source benchmark and source item identifier, the images in order, the prompt and any options, the answer with its answer format, its task category, whether it is a rewrite, and the per-item labels where they exist (\cref{sec:supp-groups}).
\bench is intended for evaluating MLLMs, and \bench-Train, from which near-duplicates of \bench images are removed, is the set intended for training (\cref{sec:train-set}).
Training on \bench items and identifying or profiling the people shown in its images are out of scope, as is commercial use of the items whose sources carry non-commercial terms (\cref{tab:supp-license}).
\bench collects no new images and inherits the images and, apart from the 1,406 rewrites and the 569 items whose label, answer format or text we correct, the items of its sources, which can include identifiable people and sensitive content.
The source identifiers and per-item labels let users exclude any source, category or item.
We maintain \bench in versioned releases and track reported errors and content concerns in a public issue list.
\begin{table}[t]
\centering\small
\setlength{\tabcolsep}{4pt}
\caption{\textbf{Lowering $\beta$ to 0.5 or doubling $\tau$ keeps the \bench score of \cica above every RL baseline of \cref{tab:main} on Qwen2.5-VL-7B.} \bench score of each training objective. Both changes bring the weights closer to 1 and the score closer to that of GRPO ($\beta=0$).}
\label{tab:supp-betatau}
\begin{tabular}{lr}
\toprule
Training objective & \bench \\
\midrule
GRPO ($\beta=0$) & 41.7 \\
\cica, $\beta=0.5$ & 44.2 \\
\cica, $\tau=1.0$ & 44.5 \\
\oursrow \cica ($\beta=0.95$, $\tau=0.5$) & 46.7 \\
\bottomrule
\end{tabular}
\end{table}

\end{document}